\documentclass{article}
\PassOptionsToPackage{numbers, compress}{natbib}

 \usepackage[preprint]{neurips_2025}

\usepackage{bbding}
\makeatletter
  \newcommand\figcaption{\def\@captype{figure}\caption}
  \newcommand\tabcaption{\def\@captype{table}\caption}
\makeatother

\usepackage[utf8]{inputenc} % allow utf-8 input
\usepackage[T1]{fontenc}    % use 8-bit T1 fonts
\usepackage{url}            % simple URL typesetting
\usepackage{booktabs}       % professional-quality tables
\usepackage{amsfonts}       % blackboard math symbols
\usepackage{nicefrac}       % compact symbols for 1/2, etc.
\usepackage{microtype}      % microtypography
\usepackage{xcolor}         % colors
\usepackage{colortbl}
\usepackage{color}
\usepackage{amsmath}
\usepackage{graphicx}
\usepackage{adjustbox}
\usepackage{caption}
\usepackage{subcaption}
\usepackage{float}

\usepackage{multirow}

\definecolor{rule}{HTML}{F27970}
\definecolor{model}{HTML}{BB9727}
\definecolor{answer}{HTML}{54B345}
\definecolor{process}{HTML}{05B9E2}
\definecolor{backred}{RGB}{255, 190, 190}
\definecolor{backblue}{RGB}{210, 230, 250}

\newcommand{\red}[1]{\textcolor{red}{#1}}
\newcommand{\blue}[1]{\textcolor{blue}{#1}}
\newcommand{\best}{\cellcolor{backred}}

\newcommand{\high}{\cellcolor{backblue}}
\newcommand{\header}[1]{\text{#1}}

\usepackage{xspace}
\usepackage[shortlabels]{enumitem}
\newcommand{\dataset}{\textsc{PhyX}\xspace}

\usepackage{tocloft}
\usepackage{etoc}
\usepackage{titletoc}

\newcommand\DoToC{%
  \startcontents
  \printcontents{}{1}{\noindent \textbf{\Large{Table of Contents in Appendix}}\vskip3pt\vskip5pt}
  \vskip3pt\vskip5pt
}
\newcommand{\listappendixname}{List of Appendices}
\newlistof{appendices}{apc}{\listappendixname}
\newlistof{appfigures}{afg}{List of Case Study Figures}

\usepackage[normalem]{ulem}
\useunder{\uline}{\ul}{}

\usepackage{makecell}
\usepackage{array}

\definecolor{citecolor}{HTML}{2980b9}
\definecolor{linkcolor}{HTML}{c0392b}

\usepackage[hidelinks,breaklinks=true,colorlinks,bookmarks=false,citecolor=citecolor,linkcolor=linkcolor]{hyperref}

\makeatletter
\newenvironment{chapquote}[2][2em]
  {\setlength{\@tempdima}{#1}%
   \def\chapquote@author{#2}%
   \parshape 1 \@tempdima \dimexpr\textwidth-2\@tempdima\relax%
   \itshape}
  {\par\normalfont\hfill--\ \chapquote@author\hspace*{\@tempdima}\par\bigskip}
\makeatother

\title{\begin{minipage}{.08\textwidth}
\centering
\vspace{-4pt}
\includegraphics[width=\linewidth]{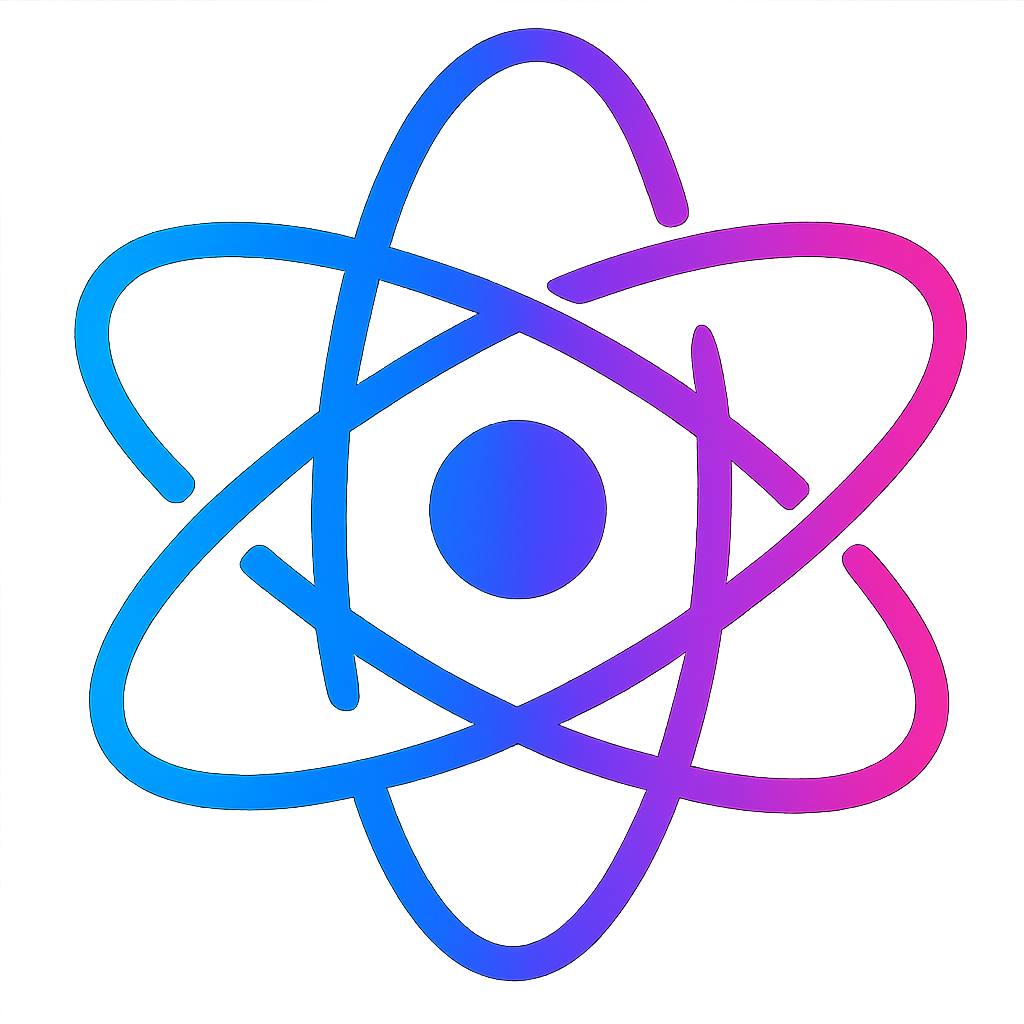} 
\end{minipage}
\dataset: Does Your Model Have the ``Wits'' \\for Physical Reasoning?}

\author{Hui Shen$^{1,2}$, Taiqiang Wu$^{1}$, Qi Han$^{3}$, Yunta Hsieh$^{2}$,\vspace{0.1cm}\\ \textbf{Jizhou Wang$^4$, Yuyue Zhang$^3$, Yuxin Cheng$^1$, Zijian Hao$^{3}$, Yuansheng Ni$^{5}$,}\vspace{0.1cm}\\ \textbf{Xin Wang$^{6}$, Zhongwei Wan$^{6}$, Kai Zhang$^{6}$, Wendong Xu$^{1}$, Jing Xiong$^{1}$,}\vspace{0.1cm}\\ \textbf{Ping Luo$^{1}$, Wenhu Chen$^{5}$, Chaofan Tao$^{1}$, Z. Morley Mao$^{2}$, Ngai Wong$^{1}$}\vspace{0.3cm}\\
  $^1$The University of Hong Kong, 
  $^2$University of Michigan, 
  $^3$Independent,\\
  $^4$University of Toronto,
  $^5$University of Waterloo,
  $^6$The Ohio State University
}

\begin{document}

\maketitle

\begin{abstract}
    Existing benchmarks fail to capture a crucial aspect of intelligence: \textit{physical reasoning}, the integrated ability to combine domain knowledge, symbolic reasoning, and understanding of real-world constraints. To address this gap, we introduce \dataset: the first large-scale benchmark designed to assess models' capacity for physics-grounded reasoning in visual scenarios. 
    \dataset includes 3K meticulously curated multimodal questions spanning \textbf{6} reasoning types across \textbf{25} sub-domains and \textbf{6} core physics domains: thermodynamics, electromagnetism, mechanics, modern physics, optics, and wave \& acoustics.
    In our comprehensive evaluation, even state-of-the-art models struggle significantly with physical reasoning. \textbf{GPT-4o}, \textbf{Claude3.7-Sonnet}, and \textbf{GPT-o4-mini} achieve only \textbf{32.5\%}, \textbf{42.2\%}, and \textbf{45.8\%} accuracy respectively—performance gaps exceeding \textbf{29\%} compared to human experts. 
    Our analysis exposes critical limitations in current models: \textit{over-reliance on memorized disciplinary knowledge}, \textit{excessive dependence on mathematical formulations}, and \textit{surface-level visual pattern matching} rather than genuine physical understanding. We provide in-depth analysis through fine-grained statistics, detailed case studies, and multiple evaluation paradigms to thoroughly examine physical reasoning capabilities.
    To ensure reproducibility, we implement an evaluation protocol based on widely-used toolkits such as \href{https://github.com/open-compass/VLMEvalKit}{VLMEvalKit}, enabling one-click evaluation. More details are available on our project page: \url{https://phyx-bench.github.io/}.
\end{abstract}
    
\section{Introduction}
\label{sec:introduction}

\vspace{-3mm}
\begin{chapquote}{Richard Feynman}
Physics is the most fundamental and all-inclusive of the sciences.
\end{chapquote}
\vspace{-3mm}

% Research progression: Establish foundational advances and emerging challenges
% Research progression: Establish foundational advances and emerging challenges
% The evolution of large language models (LLMs) has achieved landmark successes in formal reasoning domains via chain-of-thought (CoT)~\citep{wei2022chain} and reinforcement learning~\citep{ouyang2022training, guo2025deepseek}. 
State-of-the-art models~\citep{guo2025deepseek, openaio1, gemini25} now can basically solve Olympiad-level mathematical problems with human-competitive accuracy on benchmarks including AIME~\citep{AIME2024}, GPQA~\citep{rein2024gpqa}, MATH-500~\citep{hendrycks2021measuring} and OlympiadBench~\citep{he2024olympiadbench}, \textit{etc.}. Emerging multimodal large language models (MLLMs) like GPT-4o~\citep{openai2024gpt4ocard} and Claude-3.7-Sonnet~\citep{claude37} further offer promising pathways by combining visual understanding into reasoning capabilities. 
Recent advances in multimodal foundation models have spurred the development of benchmarks assessing disciplinary knowledge \citep{yue2024mmmu} and mathematical problems\citep{wang2024measuring, zhang2024mathverse, lumathvista}.
However, these evaluations overlook a critical dimension of machine intelligence: \textit{physical reasoning}, the ability to integrate disciplinary knowledge, symbolic operations, and understanding of real-world constraints. 

Physical problem-solving fundamentally differs from pure mathematical reasoning or science knowledge question answering  by requiring models to: (1) decode implicit conditions in the questions (e.g., interpreting "smooth surface" in the question as the coefficient of friction equals to zero), (2) maintain physical consistency across the reasoning chains since the laws of physics do not change with different reasoning trajectories. These differences arise from the fundamental distinction between physical reasoning and the more textual or abstract forms of reasoning emphasized in prior science-related and math-related benchmarks. 
More importantly, the capacity of physical reasoning assesses the model to ground the abstract physical formulas in the real-world visionary scenarios. It typically demands tight integration of visual perception ("Is the surface rough or smooth?"), material properties ("Will the wooden block float?"), and dynamic simulations ("How will dominoes cascade?"). 
% Crucially, humans solve such problems through multimodal mental modeling that current single-modality LLMs cannot replicate. 
% Concretely speacking, the proposed benchmark aims to answering the following questions:
% Technological frontier: Position MLLMs as potential solution candidates

\begin{figure*}[t!]
\centering
\includegraphics[width=0.9\textwidth]{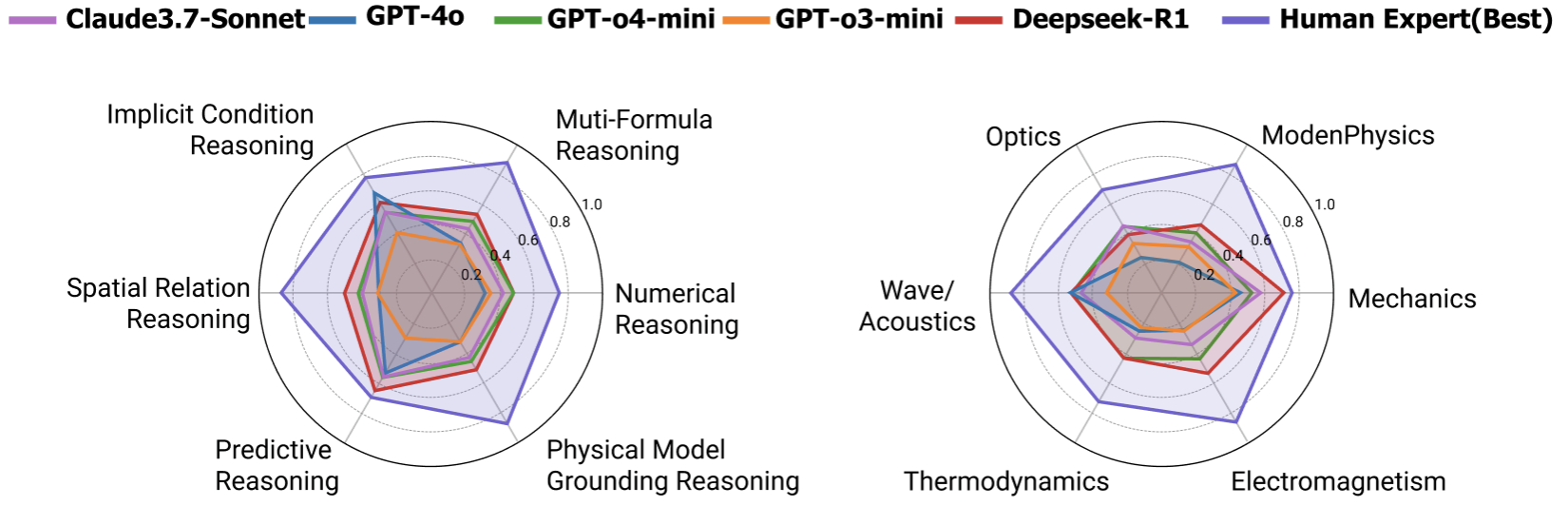} % 缩小到全宽的80%
% \includegraphics[width=\textwidth]{figures/tmp.PNG}

% \vspace{-2pt} % 调整图片与caption的间距
\caption{Accuracies of three leading MLLMs, two leading LLM and human performance on our proposed \dataset across 6 physical reasoning types and 6 domains.}
\label{fig:intro_performance}
% \vspace{-3pt} % 调整底部间距
\end{figure*}

%%% [这下面的分析需要结合表来讲]
% \begin{enumerate}[label=\roman*.]
%     \item \textbf{Whether disciplinary knowledge and  mathematical problems are sufficient for evaluation?} The advancements of LLMs in text-based reasoning, particularly in formal systems like mathematics and logic, have built the foundation for more complex reasoning. However, when visual information is integrated into MLLMs, the nature of reasoning fundamentally changes. Multi-modal reasoning usually involves interpreting complex scenes within intuitive physical and spatial frameworks. Effectively reasoning about such scenes requires an understanding of implicit physical rules (e.g., gravity, collisions), object properties (e.g., material, rigidity), and spatial relationships.

%     \item \textbf{Do MLLMs truly reason the real-world physical phenomena in evaluation?} In Figure~\ref{}, we showcase three examples from current benchmarks. Given this, we demonstrate that current visual-reasoning benchmarks might not be comprehensive enough to assess the genuine multi-modal reasoning capabilities of MLLMs.

\begin{figure*}[t!]
\centering
% \vspace{0.2cm}
\includegraphics[width=\textwidth]{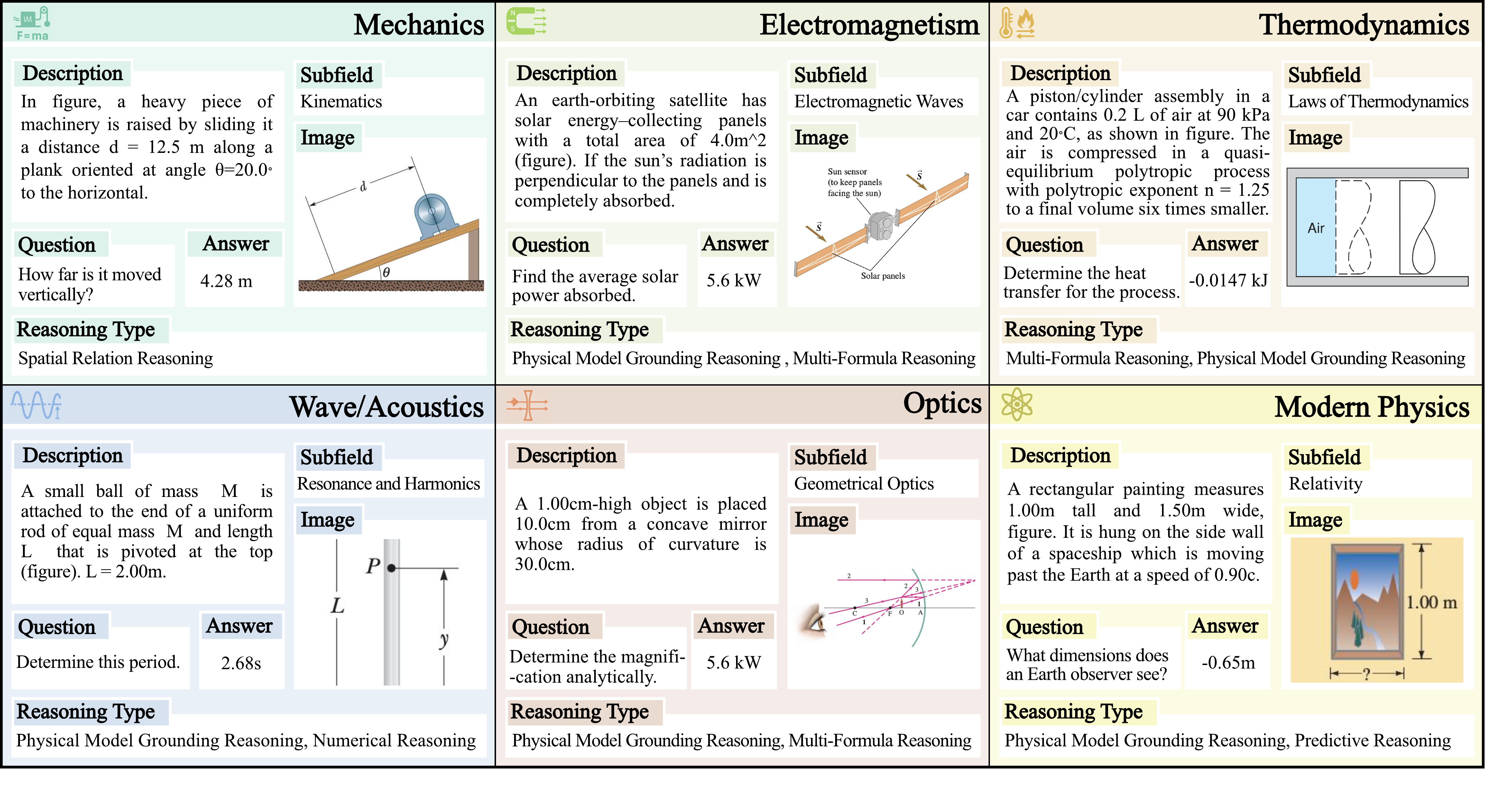}
% \vspace{0.02cm}
   \caption{Sampled \dataset examples from each domain.}
\label{fig:domains}
\vspace{-0.2cm}
\end{figure*}

To address these gaps, we present \textbf{\dataset}, the first large-scale benchmark designed for evaluating physics-based reasoning via multimodal problem-solving with three core innovations: (1) 3,000 newly collected questions with realistic physical scenarios requiring integrated visual analysis and causal reasoning, (2) Expert-validated data design covering six fundamental physics domains with representative examples illustrated in Figure~\ref{fig:domains}, and six distinct physical reasoning types, {Physical Model Grounding Reasoning}, {Spatial Relation Reasoning}, {Multi-Formula Reasoning}, {Implicit Condition Reasoning}, {Numerical Reasoning}, and {Predictive Reasoning} and (3) Strict unified three-step evaluation protocols, account for varying instruction-following capabilities across models and enables accurate assessment of reasoning. Each scenario undergoes rigorous validation by physics Ph.D. students to ensure scientific accuracy while eliminating dataset bias.

In addition to MLLMs, our benchmark supports to evaluate LLMs by translating the images into text descriptions, thereby enabling an assessment of LLMs on these visually-grounded tasks.

Our evaluation of 16 foundation models reveals an unprecedented capability gap: While the worst performance group of physics undergraduates and graduates achieve 75.6\% accuracy, the best-performing MLLM (GPT-o4-mini) scores only 45.8\%. This 30-point performance chasm persists across all physics domains, most notably in Modern Physics (human 86.7\% vs. model 40.6\%) and Wave/Acoustics (human 86.7\% vs. model 52.7\%), as shown in Figure~\ref{fig:intro_performance}. 

These results expose three critical shortcomings in current multimodal reasoning frameworks: (1) Visual reasoning errors (39.6\%) indicate that models frequently misinterpret visual context, underscoring their limited capability in accurately extracting and reasoning from realistic physical scenarios.
(2) The inconsistent performance across input variations—specifically, Full-Text, Text-DeRedundancy, and Text-Minimal—demonstrates that MLLMs remain overly dependent on textual descriptions, failing to effectively leverage visual input for reasoning. 
(3) Comparing physical reasoning performance to mathematical reasoning benchmarks such as MathVerse~\citep{lumathvista} and MATH-V~\citep{wang2024measuring} reveals that physical reasoning poses significantly greater challenges, highlighting a critical need for improved integration of abstract concepts and real-world knowledge.
\textbf{\dataset} thus provides both a diagnostic toolkit for model improvement and a roadmap for developing physically-grounded AI systems.

Our contributions can be summarized as follows:
\noindent\textbf{Novel Benchmark Design:} We introduce \dataset, the first large-scale benchmark for evaluating the reasoning capabilities in the physical world for both multi-modal models and language models. Curated by experts, it spans 25 fine-grained domains and 6 reasoning types with realistic scenarios.
\noindent\textbf{Versatile Evaluation Framework:} \dataset supports versatile evaluation frameworks, including \textit{assessment formats} (multiple-choice vs. open-ended) and \textit{hierarchical answer judge} (rule-based and model-based). 
%\textit{reasoning styles} (CoT vs. non-CoT), and 
% \textit{learning paradigms} (zero-shot vs. few-shot)
It also seamlessly integrates with mainstream toolkits (e.g., VLMEvalKit) for reproducible benchmarking.
\noindent\textbf{Critical Insights on Reasoning:} We provide granular performance analysis and reveal some interesting observations, which sheds light on the design of the future models that jointly consider the disciplinary knowledge, symbolic operations, and real-world constraints for high-level physical reasoning.
    
% \end{enumerate}

\begin{figure*}[t!]
\centering
% \vspace{0.2cm}
\includegraphics[width=\textwidth]{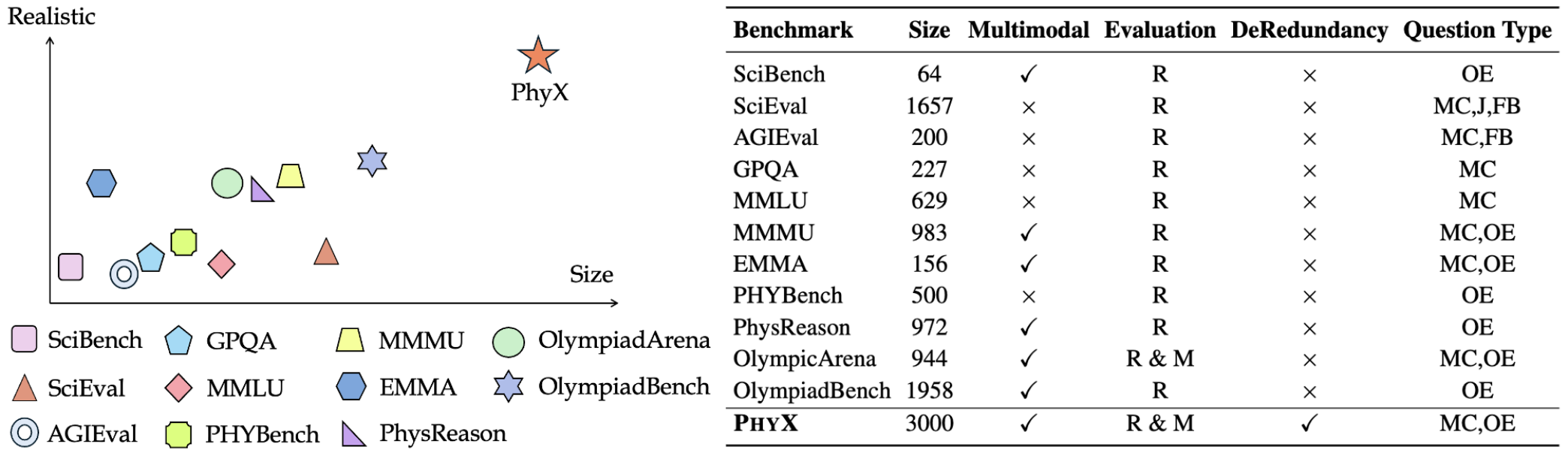}
% \vspace{0.02cm}
   \caption{Comparison with existing physics benchmarks. Realistic refers to the extent to which the dataset contains visually realistic physical scenarios. Size indicates the number of physics questions with images in multimodal benchmarks or total physics questions in text-only benchmarks. For evaluation methods, R: rule-based, M: model-based. For question type, OE: Open-ended, MC: Multiple-choice, FB: Fill-in-the-blank, J: Judgement. Upon comparison, \dataset leads in all aspects.}
\label{fig:comparison_with_benchmarks}
% \vspace{-0.2cm}
\end{figure*}

\begin{figure*}[t!]
\centering
% \vspace{0.2cm}
\includegraphics[width=\textwidth]{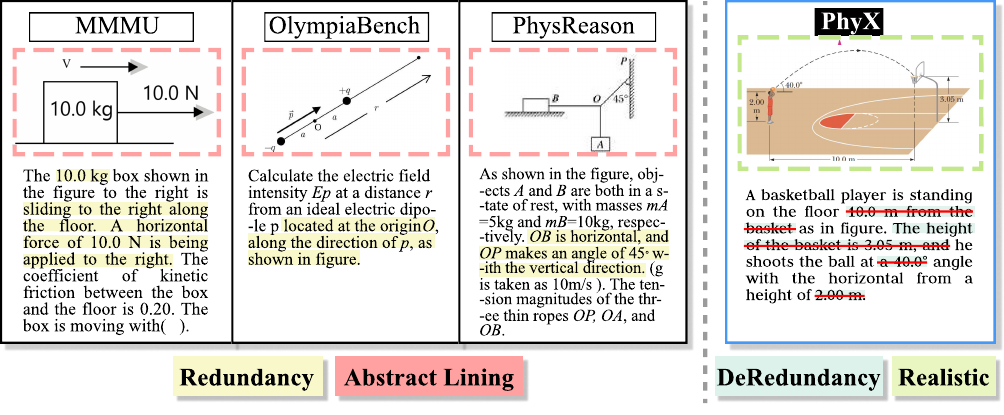}
% \vspace{0.02cm}
   \caption{Existing benchmarks that contain physics-related questions suffer from information redundancy and abstract representation. In contrast,  de-redundancy in  \dataset benchmark increases the difficulty, as models can perceive concepts from  ONE modality only. Additionally, realistic visuals provide authentic context that challenges models to accurately  apply physical laws. 
}
\label{fig:existing_physics}
% \vspace{-0.2cm}
\end{figure*}

\section{The PhyX Benchmark}
\label{sec:phyx}

\begin{figure*}[t]
\centering
\begin{minipage}[c]{0.38\textwidth}
\small
\centering

  \tabcaption{\textbf{Key Statistics of \dataset.}}
  \label{tab:phyx_stats}
  \centering
  \begin{adjustbox}{width=\linewidth}
   \begin{tabular}{lr}
 \toprule
 \textbf{Statistic} & \textbf{Number} \\
 \midrule
  \textbf{Total new questions} & \textbf{6,000} \\
  ~- Multiple-choice questions & 3,000 (50.0\%) \\
  ~- Open-ended questions & 3,000 (50.0\%) \\
 \midrule
 Unique number of images & 3,000 \\
Unique number of questions & 3,000 \\
  \midrule
 Maximum description length & 288\\
 Maximum question length & 119 \\
 Maximum option length & 46 \\
 Average description length & 48.3\\
 Average question length & 14.6 \\
 Average option length & 11.2 \\
 \bottomrule
 \end{tabular}
 \end{adjustbox}
\end{minipage}
\qquad
\begin{minipage}[c]{0.50\textwidth}
\centering
% \vspace{-0.2cm}
\label{fig3.5}
% \vspace{0.15cm}
\includegraphics[width=0.95\textwidth]{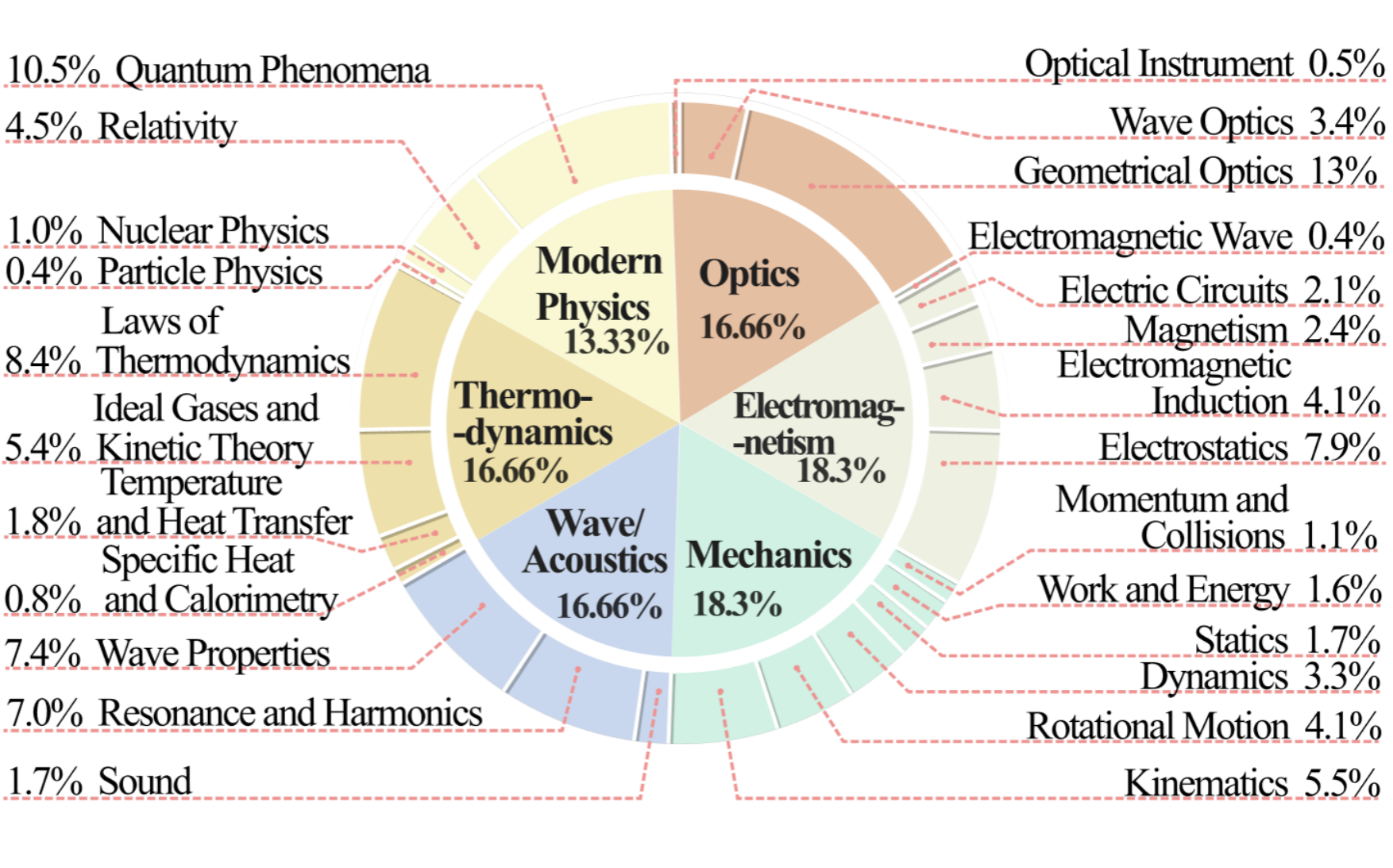}
\caption{\textbf{Fine-grained Distribution of \dataset.}}
\end{minipage}
\label{tab:}
\end{figure*}

% In Sec.~\ref{sec:dataset_overview}, we first present an overview of the curated visual physical dataset in \dataset. Then, in Sec.~\ref{sec:comparison}, we compare \dataset with existing benchmarks. 

\subsection{Overview of \dataset}
\label{sec:dataset_overview}

We introduce \dataset, a novel benchmark meticulously curated to assess the physical reasoning capabilities of foundation models.
\dataset consists of 3,000 visually-grounded physics questions, meticulously curated to cover six distinct physics domains including \textit{Mechanics} (550), \textit{Electromagnetism} (550), \textit{Thermodynamics} (500), \textit{Wave/Acoustics} (500), \textit{Optics} (500), and \textit{Modern Physics} (400).
Each problem in \dataset is centered around realistic physical scenarios to robustly assess the model's ability to reason the physical world.
Detailed data statistics are summarized in Table~\ref{tab:phyx_stats}, with representative question examples from each domains illustrated in Figure~\ref{fig:domains}. 
%
% All the questions in our benchmark were manually collected by graduate students in STEM fields, drawing from online sources and textbooks. 
%
To enable comprehensive assessment, each question within \dataset has been categorized into six well-defined physical reasoning types: \textit{Physical Model Grounding Reasoning}, \textit{Spatial Relation Reasoning}, \textit{Multi-Formula Reasoning}, \textit{Implicit Condition Reasoning}, \textit{Numerical Reasoning}, and \textit{Predictive Reasoning}. Detailed definitions and illustrative examples of these reasoning types are provided in Appendix~\ref{sec:a_physical_reasoning_definition}.

Through its carefully curated structure and extensive coverage of diverse reasoning dimensions, \dataset represents a robust resource for systematically benchmarking and advancing the capabilities of foundation models in realistic physical reasoning tasks.

\subsection{Data Curation Process}
\textbf{Data Collection.} To ensure high-quality data, we design a four-stage data collection process. 
Firstly, we conducted an in-depth survey of core physics disciplines to determine the coverage of our benchmark. We selected diverse physics domains and subfields, and defined a set of reasoning types.
Secondly, we recruited a team of graduate students in STEM fields to serve as expert annotators. Annotators are instructed to comply with copyright and licensing rules by avoiding content from sources that restrict copying or redistribution. To mitigate potential data contamination in foundation models, they are also advised to select questions for which answers are not immediately available alongside the problem, such as those found in separate materials or at the end of textbooks.
Then, each open-ended question is required to be converted into a multiple-choice version, and vice versa. We also constructed three parallel versions of each question: (1) the original version as it appears in the textbook; (2) a concise version where redundant textual information—those duplicated by the corresponding image—was removed; and (3) a question-only version that retains only the core question.
Lastly, to support evaluation of LLMs and facilitate multi-modal understanding, we used GPT-4o to generate descriptive captions for each image, aim to summarize the visual content in a self-contained textual form. This data curation process results in a diverse collection of 3,300 questions from various sources. The detailed annotation protocol is in Appendix~\ref{appendix:data_annotation}.

\textbf{Data Quality Control.} To further control the quality of our data, we perform a three-stage data cleaning process. First, we detect potentially duplicated questions by analyzing lexical overlap, followed by manual review from physics Ph.D. students to confirm and remove duplicates. Then, we filter out the shortest 10\% of questions based on their textual length. This rigorous process plays a crucial role in maintaining the quality and difficulty of \dataset.

\begin{figure*}[t!]
\centering
% \vspace{0.2cm}
\includegraphics[width=\textwidth]{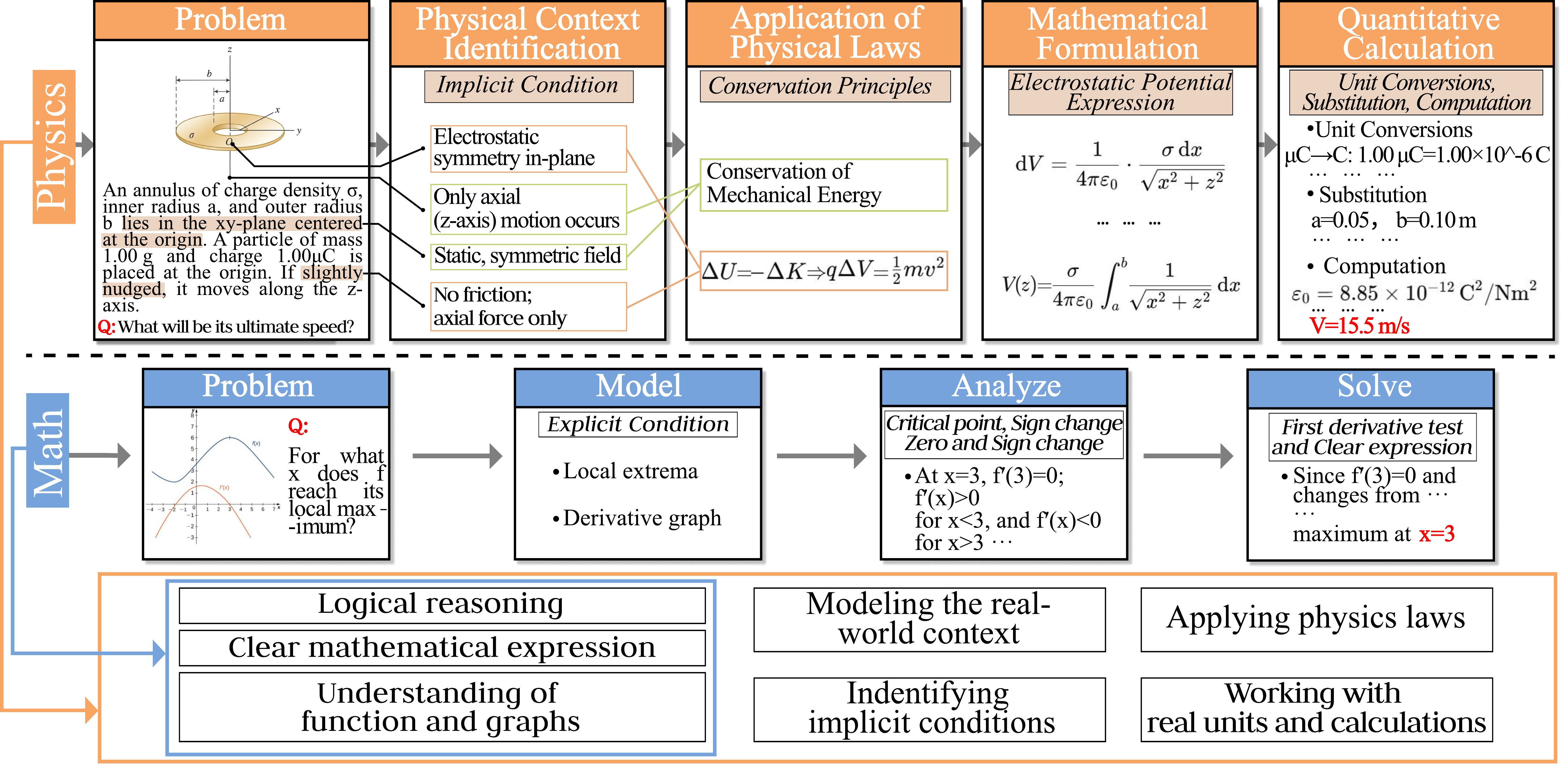}
% \vspace{0.02cm}
   \caption{An real example of reasoning trajectory based on GPT-4o and the comparison of required capabilities when solving physical and mathematical problems.}
\label{fig:physics_math}
\vspace{-0.2cm}
\end{figure*}

\subsection{Key Difference Compared to Existing Benchmarks}
\label{sec:comparison}

\textbf{Compared with Scientific Knowledge Benchmarks.}
From Figure~\ref{fig:comparison_with_benchmarks},  science benchmarks like MMMU \citep{yue2024mmmu} cover broad disciplinary reasoning but lack focus on deep reasoning capability. These benchmarks often rely on memorization and basic understanding of disciplinary knowledge, with tasks that prioritize factual recall or simple cross-modal association.  In contrast, \dataset specializes in university-level hard questions through high-fidelity visual scenarios. Unlike generalist benchmarks, our tasks demand integration of visual cues with implicit physical laws, requiring models to surpass mere knowledge recall and perform nuanced, context-driven inference. This targeted design evaluates true multimodal reasoning about the physical world, exposing gaps in models’ ability to handle professional-level scientific challenges.

\textbf{Compared with Mathematical Reasoning Benchmarks.}
Mathematical reasoning benchmarks, such as MathVista~\citep{lumathvista}, MathVerse~\citep{zhang2024mathverse}, and MATH-V~\citep{wang2024measuring}, focus on logical deduction with clear expressions and explicit conditions, representing a subset of the challenges in physical reasoning. Physical reasoning, as evaluated by \dataset, extends beyond these by requiring models to model real-world contexts (e.g., dynamic physical systems), identify implicit conditions from visual cues (e.g., Figure~\ref{fig:physics_math}), and integrate the application of physical laws with symbolic logic, which are key capabilities absent in purely mathematical tasks. This makes \dataset a more comprehensive test of multimodal reasoning, capturing the complexity of real-world physics problems.

\textbf{Compared with Physics-related Benchmarks}
Existing benchmarks (e.g., PHYBench \citep{qiu2025phybench}, UGPhysics \citep{xu2025ugphysics}, OlympiadBench \citep{he2024olympiadbench}) prioritize text-based problems or schematic visuals, limiting their assessment of multimodal reasoning. In details, PHYBench’s  problems and UGPhysics’s  questions rely heavily on textual descriptions, while OlympiadBench’s problems use simplified diagrams, as shown in Figure~\ref{fig:existing_physics}. These benchmarks mainly test disciplinary knowledge but overlook the integration of visual perception with implicit physical constraints. \dataset bridges these gaps by embedding high-fidelity visual scenarios that require models to decode complex visual cues,  infer context-specific physical laws and then reasoning problems. Unlike existing datasets, \dataset mandates equal reliance on both modalities with information de-redundancy, providing a rigorous evaluation of professional-level physical reasoning in multimodal large language models.
\section{Experiments}
\label{sec:experiments}

\begin{table*}[t!]
\vspace{-1mm}
    \caption{Accuracy scores on the \textit{testmini} subset of \dataset. The highest scores of models in each section and the overall highest score are respectively highlighted in \blue{blue} and \red{red}.}
\centering
 \renewcommand\tabcolsep{6pt}
 \renewcommand\arraystretch{1.05}
 \resizebox{1.0\linewidth}{!}{
    \begin{tabular}{lcccccc}
    \toprule
    \multirow{2}{*}{\header{\textbf{Models}}} & \multicolumn{2}{c}{\textbf{Full-Text}} & \multicolumn{2}{c}{\textbf{Text-DeRedundancy}} & \multicolumn{2}{c}{\textbf{Text-Minimal}} \\
    \cmidrule(lr){2-3} \cmidrule(lr){4-5} \cmidrule(lr){6-7}
    & \header{\textit{Open-Ended}} & \header{\textit{Multi-Choice}} & \header{\textit{Open-Ended}} & \textit{Multi-Choice} & \header{\textit{Open-Ended}} & \textit{Multi-Choice} \\
    \midrule
    Random Choice & - & 25 & - & 25 & - & 25 \\
    Human Expert (Worst) & - & - & 75.6 & - & - & - \\
    Human Expert (Medium) & - & - & 77.8 & - & - & - \\
    Human Expert (Best) & - & - & 78.9 & - & - & - \\
    \midrule
    \multicolumn{7}{l}{\hfill \textit{Multimodal Large Language Models} } \\
    \midrule
    Claude3.7-Sonnet & 44.4 & 65.8 & 42.2 & 64.5 & 17.2 & 41.6 \\
    Claude3.5-Sonnet & 40.2 & 62.6 & 39.0 & 63.5 & 17.0 & 43.5 \\
    Claude3.5-Haiku & 7.9 & 37.0 & 13.6 & 37.5 & 5.5 & 31.7 \\
    GPT-o4-mini & \high{49.0} & \best{87.9} & \high{45.8} & \best{86.9} & \best{24.1} & \best{62.6} \\
    GPT-4o & 33.9 & 61.0 & 32.5 & 57.6 & 14.3 & 43.8 \\
    InternVL3-78B & 35.9 & 45.6 & 33.1 & 46.9 & 14.8 & 40.5 \\
    Yi-VL-34B & 3.5 & 34.8 & 3.4 & 34.1 & 1.9 & 34.1 \\
    InternVL3-14B & 9.0 & 46.9 & 7.9 & 47.5 & 5.1 & 45.9 \\
    InternVL3-8B & 6.3 & 45.5 & 6.5 & 44.9 & 4.6 & 44.0 \\
    MiniCPM-o-8B & 7.1 & 31.8 & 7.2 & 31.6 & 3.2 & 34.2 \\
    LLaVA-OneVision-7B & 7.2 & 37.7 & 5.7 & 37.3 & 2.7 & 38.0 \\
    DeepSeek-VL2-4.5B & 11.4 & 28.2 & 10.2 & 27.8 & 4.7 & 27.3 \\
    Kimi-VL-A3B-Instruct-2.8B & 15.6 & 37.1 & 15.4 & 38.7 & 8.1 & 39.3 \\
    \midrule
    \multicolumn{7}{l}{\hfill \textit{Large Language Models} } \\
    \midrule
    DeepSeek-R1 & \best{51.8} & 63.1 & \best{51.2} & 62.9 & \high{22.2} & 43.6 \\
    DeepSeek-V3 & 40.7 & 70.8 & 36.3 & 67.5 & 16.2 & 49.9 \\
    % Qwen3-8B &  &  & 27.5 & 48.4 & 12.1 & 41.8 \\
    GPT-o3-mini & 36.9 & \high{78.5} & 31.5 & \high{76.9} & 14.3 & \high{56.2} \\
    \bottomrule
    \end{tabular}
}
% \vspace{-3mm}
\label{tab:main_exp}
\end{table*}

% We conduct a systematic evaluation of existing LLMs and MLLMs on \dataset. We first introduce the experimental setup in Sec.~\ref{sec:exp_setup} followed by our evaluation protocols in Sec.~\ref{sec:eval_protocols}. Then, we present the main results in Sec.~\ref{sec:main_exp_results}, followed by a error analysis in Sec.~\ref{sec:error_analysis}.

\subsection{Experimental Setup}
\label{sec:exp_setup}

\textbf{The \textit{testmini} Subset} \dataset comprises 3,000 high-quality visual physics problems and 18,000 corresponding test instances. To streamline evaluation and accelerate model development validation, we extract a smaller representative subset named \textit{testmini} including 1,000 problems and 6,000 instances. The construction of \textit{testmini} involved a proportional random sampling strategy across different physics domains of \dataset. The quantitative evaluations in all subsequent experiments were assessed on this \textit{testmini} subset.

\textbf{Baselines.} We include random chance as naive baselines. Additionally, we recruiting 15 undergraduate and graduate physics students to represent the expert performance baseline, each student was tasked with completing 18 questions. The students were divided into three groups of five, and the results of each group are reported separately. Then, we conduct experiments on (a) Reasoning MLLMs: 
% Gemini-2.5-Pro~\citep{gemini25}, 
GPT-o4-mini~\citep{gpto4mini}, Claude-3.7-Sonnet~\citep{claude37}, LLaVA-OneVision-7B~\citep{li2024llava} MiniCPM-o~\citep{yao2024minicpm}, (b) General MLLMs: GPT-4o~\citep{openai2024gpt4ocard}, Claude-3.5-Sonnet~\citep{claude35sonnet}, Claude-3.5-Haiku~\citep{claude35haiku}, InternVL3~\citep{zhu2025internvl3}, Yi-VL-34B~\citep{young2024yi},
% , Qwen-VL2.5~\citep{}
(c) LLMs: o3-mini~\citep{gpto3mini}, DeepSeek-R1~\citep{guo2025deepseek}, DeepSeek-V3~\citep{deepseekai2025deepseekv3technicalreport},
% LLaMA-3~\citep{}, 
Qwen-3-4B~\citep{yang2025qwen3technicalreport}, augmented with image captions generated by GPT-4o. 

\subsection{Evaluation Protocols}
\label{sec:eval_protocols}

Our evaluation is conducted with Chain-of-Thought (CoT) prompting to assess the reasoning capability of models. For both open-ended (OE) and multiple-choice (MC) questions, the instruction-following capabilities of models can vary significantly. To this end, we design a universal evaluation pipeline for all recent LLMs and MLLMs with different instruction-following capabilities: 

\textbf{Step 1. Prediction Generation.} Initially, the models generate prediction given the input query, which incorporates different problem description according to the specific settings, the question, and the image, using the template defined in Appendix~\ref{appendix:cot_prompt}.

\textbf{Step 2. Answer Extraction.} The raw predictions often contain reasoning steps, explanations, or irrelevant conversational filler. To precisely extract the definitive answer from these raw outputs, we separately employ rule-based answer extraction strategies, which are detailed in Appendix~\ref{appendix:rule_extraction}.

\textbf{Step 3. LLM Judge.} For OE questions, the next step is comparing the extracted answer against the ground truth to determine its correctness. Given that answers in OE physics questions can be expressed in myriad ways, we proposed an evaluation mechanism using a LLM, such as DeepSeek-V3~\citep{deepseekai2025deepseekv3technicalreport}, as a judge, using the template defined in Appendix~\ref{appendix:prompt_answer_judge}. We feeds the answer extracted and the ground truth to a LLM multiple times and checks if a LLM succeed in all attempts. A preliminary study of 200 examples shows that DeepSeek-V3 can judge the answer with more than 99\% accuracy with affordable costs. For MC questions, we first attempt to directly match the option letter. If this direct matching fails, we then use a LLM as a judge, using the template for OE questions.

\begin{table*}[t!]
\vspace{-1mm}
\caption{Average scores by model across different domains of physics with open-ended text de-redundancy questions. The highest scores of models in each section and the overall highest score are respectively highlighted in \blue{blue} and \red{red}.}
\centering
\renewcommand\tabcolsep{6pt} % 适当减小列间距
\renewcommand\arraystretch{1.1}
\resizebox{1.0\linewidth}{!}{
    \begin{tabular}{l *{6}{>{\centering\arraybackslash}p{1.8cm}} c}
    \toprule
    \textbf{Models} & \textbf{Overall} & \textbf{\makecell{Mechanics}} & \textbf{\makecell{Electro-\\magnetism}} & \textbf{\makecell{Thermo-\\dynamics}} & \textbf{\makecell{Waves \&\\ Acoustics}} & \textbf{\makecell{Optics}} & \textbf{\makecell{Modern\\ Physics}} \\
    \midrule
    Human Expert (Worst)& 75.6 & 76.5 & 60.0 & 66.7 & 86.7 & 69.2 & 86.7 \\
        Human Expert (Medium)& 77.8 & 94.1 & 53.3 & 60.0 & 93.3 & 76.9 & 86.7 \\
    Human Expert (Best)& 78.9 & 76.5 & 86.7 & 73.3 & 86.7 & 69.2 & 86.7 \\
    \midrule
    \multicolumn{8}{l}{\hfill \textit{Multimodal Large Language Models}} \\
    \midrule
    Claude3.7-Sonnet & 42.2 & \high{58.2} & 36.7 & 31.5 & 46.7 & \best{44.6} & 35.2\\
    Claude3.5-Sonnet & 39.0 & 53.5 &  27.8 & 33.3 & 49.7 & 35.5 & 3.9 \\
    Claude3.5-Haiku & 13.6 & 18.8 & 8.9 & 11.5 & 18.8 &12.0  &11.5 \\
    GPT-o4-mini & \high{45.8} & 52.3 & \high{43.2} & \best{41.8} & 52.7 & 44.0  & \high{40.6} \\
    GPT-4o & 32.5 & 45.9 & 24.3 & 26.1 & \best{53.9} & 23.5 & 21.2\\
    InternVL3-78B & 33.1 & 48.8 & 27.2 & 25.5 & 43.0 & 28.9 & 24.8\\
    Yi-VL-34B & 3.4 & 1.8 & 3.5 & 4.8 & 2.4 & 4.2 & 3.6\\
    InternVL3-14B & 7.9 & 12.4 & 8.88 & 4.2 & 8.5 & 4.8 & 8.5\\
    InternVL3-8B & 6.5 & 10.6 & 6.5 & 3.6 & 4.9 & 6.6 & 6.7\\
    MiniCPM-o-8B & 7.2 & 11.8 & 6.5 & 6.1 & 7.3 & 6.0 & 5.5\\
    LLaVA-OneVision-7B & 5.7 & 10.6 & 4.1 & 6.1 & 7.3 & 3.0 & 3.0\\
    DeepSeek-VL2-4.5B & 10.2 & 16.5 & 7.1 & 10.3 & 13.3 &9.0&4.8 \\
    Kimi-VL-A3B-Instruct-2.8B & 15.4 & 20.6 & 10.1 & 13.3 & 20.0 & 16.2 & 12.1\\
    \midrule
    \multicolumn{8}{l}{\hfill \textit{Large Language Models}} \\
    \midrule
    DeepSeek-R1 & \best{51.2} & \best{71.8} & \best{53.2} & \best{41.8} & \best{53.9} & \high{39.8} & \best{46.1} \\
    DeepSeek-V3 & 36.3 & 52.9 & 39.6 & 28.5 & 36.4 & 28.9 & 30.9\\
    GPT-o3-mini & 31.5 & 41.8 & 24.9 & 23.6 & 32.1 & 33.7 & 32.7\\
    Qwen3-8B & 27.5 & 42.9 & 23.7 & 21.2 & 35.8 & 21.1 & 20.0 \\
    \bottomrule
    \end{tabular}
}
% \vspace{-3mm}
\label{tab:main_exp_domains}
\end{table*}

\subsection{Main Results}
\label{sec:main_exp_results}

In this section, we present a comprehensive comparison of LLMs and MLLMs on \dataset benchmark, detailed in Table~\ref{tab:main_exp} and Table~\ref{tab:main_exp_domains}. Our key findings can be summarized as follows:

\textbf{Challenging Nature of \dataset.} \dataset presents significant challenges for current models. Notably, even worst human experts achieve accuracy of 75.6\%, significantly outperforming all the models included in our comparative analysis. This disparity demonstrates an existing gap between human expertise and current model capabilities, reflecting the demanding standards inherent in \dataset.

\textbf{Multiple-Choice Format Narrows the Performance Gap.} The result reveals that multiple-choice questions reduce the performance gap across models, enabling weaker models to rely on surface-level cues. In contrast, open-ended questions demand genuine reasoning and precise answer generation, leading to greater differentiation between models. This suggests that the open-ended format provides higher discriminative power when evaluating multimodal reasoning capabilities.

\textbf{Model Performance across Different Domains.} As shown in Table~\ref{tab:main_exp_domains}, in domains such as Waves/Acoustics and Mechanics, which typically include natural images and questions requiring relatively less reasoning, models tend to achieve higher performance. Conversely, in domains such as Thermodynamics and Modern Physics, where tasks frequently demand intricate visual perception and multi-step reasoning, models performance is generally lower.

\subsection{Discussion}
\label{sec:fine_grained_analysis}

\textbf{Reasoning-oriented Models Perform Better.} Leading reasoning-oriented models such as GPT-o4-mini and DeepSeek-R1 achieve accuracies of 45.8\% and 51.2\%, respectively, significantly outperforming general-purpose models like GPT-4o and Claude3.7-Sonnet. The results highlight the advantage of models specifically optimized for reasoning tasks, suggesting that architectural and training differences play a key role in bridging the multimodal reasoning gap.

\textbf{LLMs Achieve Competitive Results.} Despite lacking direct visual input, LLMs such as DeepSeek-R1 and GPT-o3-mini perform competitively with most multimodal models. The strong performance of LLMs suggests that, in many cases, the caption provides sufficient visual context for reasoning. This highlights both the impressive generalization capabilities of LLMs and the current limitations of MLLMs in leveraging raw visual signals for physical reasoning.

\textbf{MLLMs' Physical Reasoning Relies More on Text.} Our experiments show a clear performance gradient across the three input variations: Full-Text, Text-DeRedundancy, and Text-Minimal, with decreasing accuracy in that order. This indicates that MLLMs rely heavily on detailed textual descriptions, highlighting their limited ability to reason purely from visual context.

\textbf{Physical Reasoning Poses Greater Challenges than Mathematical Reasoning.} Comparing GPT-4o’s performance on our physical reasoning dataset to its previously reported results on MathVista (63.8\%) and MATH-V (63.8\%), we observe notably lower accuracy in physical reasoning tasks. This finding emphasizes that physical reasoning inherently requires a deeper integration of abstract concepts and real-world knowledge, presenting a more substantial challenge for current models compared to purely mathematical contexts.

% \textbf{Zero-shot Evaluation Results in Lower Accuracy.} Our evaluation demonstrates that zero-shot model-based judgment of answer correctness consistently underperforms relative to five-shot prompting. This reveals the necessity of incorporating few-shot examples when leveraging model-based evaluators to reliably assess complex multimodal reasoning tasks.

% \textbf{Visual Noise Significantly Impedes Physical Reasoning Performance.} Introducing noise into input images notably reduces model accuracy, underscoring the sensitivity of MLLMs to visual clarity. This decline indicates that even minor visual perturbations disrupt the models' ability to extract accurate physical information, suggesting the current multimodal reasoning mechanisms lack robustness against degraded visual inputs.

\begin{figure*}[t!]
\centering
\includegraphics[width=0.95\textwidth]{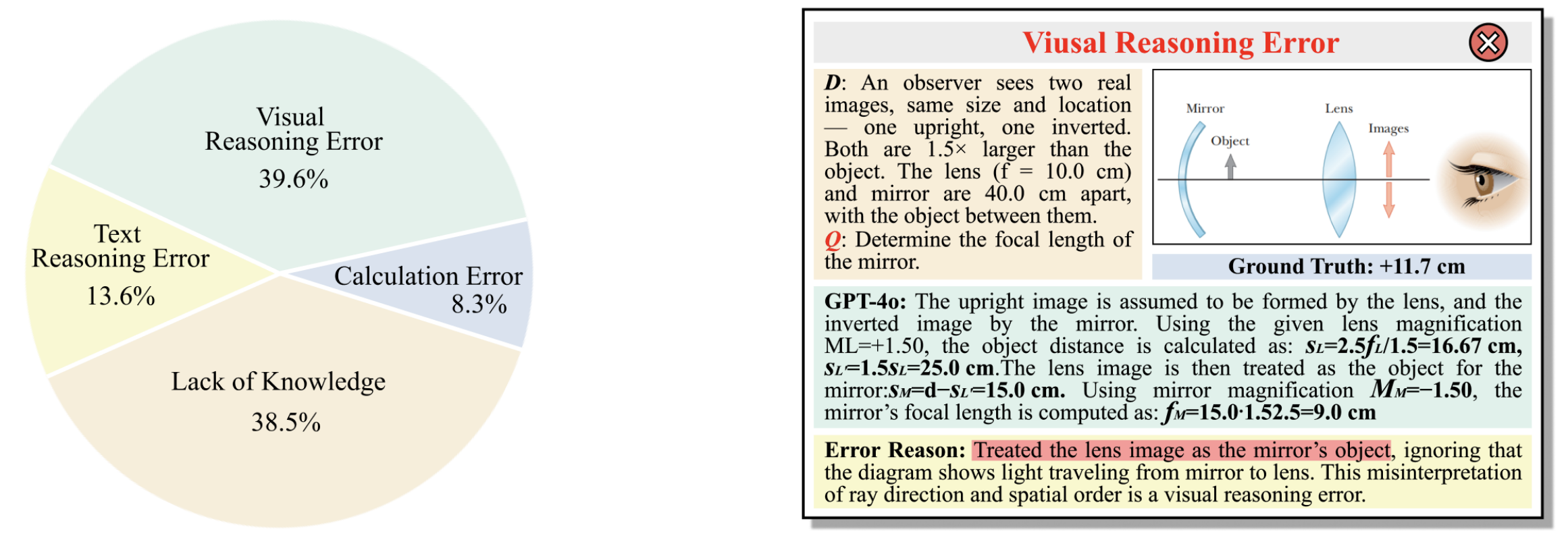}
\caption{The error distribution over 90 annotated errors based on GPT-4o with a typical visual reasoning error , which is easy for humans but challenging for GPT-4o. More examples can be found in the Appendix.}
\vspace{-0.4cm}
\label{fig:error_analysis}
\end{figure*}

\subsection{Error Analysis} 
\label{sec:error_analysis}

To dive into the reasoning capabilities and limitations of models, we meticulously inspected 96 randomly sampled incorrect predictions and performed an in-depth analysis based on GPT-4o. The objectives of this analysis were twofold: to identify current model weaknesses and to guide future enhancements in model design and training. The distribution of these errors is illustrated in Figure~\ref{fig:error_analysis}, and a comprehensive case study of 30 notable cases is included in Appendix~\ref{appendix:case_study}. 

\textbf{Visual Reasoning Errors (39.6\%)} arise from the model's incorrect extraction, spatial relationships, or reasoning based on visual information from realistic physical questions included in \dataset. A notable instance of this can be observed in Appendix~\ref{fig:electromagnetism_3}, where the model misread the voltage value shown in the image, leading to a numerical error in its calculation. Given the realistic nature of our images, visual reasoning errors constitute a larger proportion of mistakes, posing a significant new challenge to MLLMs compared to existing benchmarks.

\textbf{Text Reasoning Errors (13.5\%)} are characterized by incorrect processing or interpretation of textual content. The model occasionally struggles with implicit conditions, or incorrectly handles logical relationships presented in text form. An example of this can be illustrated in Appendix~\ref{fig:mechanics_4}, where the model overlooked the explicit instruction to ignore friction and instead reasoned that the coefficient of friction was required to solve the problem. This highlights areas for improved textual inference and contextual reasoning are critical to address these shortcomings.

\textbf{Lack of Knowledge (38.5\%)} reflects GPT-4o’s incomplete understanding of specific domain knowledge. As demonstrated in the example in Appendix~\ref{fig:optics_5}, the model lacks the fundamental knowledge that a difference in wave speeds across media invalidates direct geometric reasoning based on symmetric travel paths. Specifically, it ignores that the slower speed in the liver requires a correction when estimating depth from the reflection geometry, leading to an overestimated result.

\textbf{Calculation Error (8.3\%)} refer to mistakes in arithmetic operations, formula application, or unit conversions. These errors indicate that the model has grasped the physical context and relevant concepts but fails in the final step of numerical computation.

\section{Related Work}
\label{sec:related_work}

\textbf{Multi-modal Large Language Models.} 
Multi-modal large language models (MLLMs)~\citep{claude37, gpto4mini, gemini25} have shown great potential in a wide achieved excellent visual understanding ability by integrating both visual and textual data in a wide range of multimodal tasks. Recent advances in LLMs have motivated efforts~\citep{wei2022chain, ouyang2022training} to explore MLLM reasoning. Despite such achievements, it remains unclear whether these models truly possess advanced reasoning abilities when solving the visual tasks, especially in the physical area that is closer to the real world. To bridge this gap and comprehensively evaluate the physical reasoning capabilities of MLLMs, we introduce \dataset, a multimodal benchmark to evaluate the real reasoning ability of recent advanced MLLMs in physics.

\textbf{LLM Benchmarks.}  Several benchmarks~\citep{hendrycksmeasuring, sun2024scieval, rein2024gpqa, austin2021program, zhou2023instructionfollowingevaluationlargelanguage} have been proposed to evaluate LLM's ability on various aspects. Among these works, the most related one is PHYBench~\citep{qiu2025phybench}, which also focuses in the physic reasoning area. Although evaluating the same discipline, their scope remains narrow since it includes only a small number of questions, making it insufficient to fully assess a model's reasoning capabilities. Furthermore, PHYBench concentrates exclusively on evaluating the understanding of physics concepts by language models through text. However, in real-world scenarios, solving physics problems also requires visual perception and interpretation. 

\textbf{MLLM Benchmarks.} Recently, several MLLM scientific benchmark~\citep{yue2024mmmu, wang2024scibench, he2024olympiadbench, huang2024olympicarena, zhang2025physreason, hao2025can} have also been proposed. For example, PhysReason~\citep{zhang2025physreason} includes a multimodal subset of 972 physics problems
with figures to evaluate the MLLMs. EMMA~\citep{hao2025can} composes 2,788 problems covering various scientific area such as mathematics, physics, and coding. However, all of these benchmarks only contain a small subset of data in physics area, which still could not fully evaluate the MLLM's ability on reasoning and solving the advanced physics problems.

\section{Conclusion and Limitations}
\label{sec:conclusion}
Existing benchmarks have overlooked the critical task of physical reasoning, which requires integrating domain knowledge, symbolic reasoning, and real-world constraints. To address this, we present \dataset, the first large-scale benchmark for evaluating physical reasoning in multimodal, visually grounded scenarios. Through rigorous evaluation, we reveal that state-of-the-art models exhibit significant limitations in physical reasoning, relying predominantly on memorized knowledge, mathematical formulas, and superficial visual patterns, rather than genuine understanding of physical principles. Our findings highlight the urgent need for future models to improve deep physical reasoning over surface-level associations, guiding the development of more intelligent models.

On the other hand, our benchmark focuses exclusively on English-language prompts and annotations. While this aligns with the dominant language used in most foundation models, it is not suitable for assessing a model's reasoning ability in other languages. Also, the images in our dataset depict physically realistic scenarios but are often schematic or textbook-style rather than real-world photographs. While suitable for evaluating conceptual reasoning, this may not fully capture the complexity of perception in natural environments.

\bibliographystyle{unsrtnat}
\bibliography{main}

\appendix
\newpage

\setcounter{section}{0}
\renewcommand{\thesection}{\Alph{section}} 

\DoToC

\clearpage

\section{Ethics Statement}

\textbf{Legal Compliance.} All questions included in \dataset are sourced from publicly accessible materials. During data collection, annotators are instructed to strictly follow the copyright and licensing terms of the original platforms. Any content from sources that prohibit reuse or redistribution MUST be explicitly excluded. \dataset is a non-commercial project, and its usage aligns with the principles outlined in Fair Use §107: "the fair use of a copyrighted work, including such use by ...... scholarship, or research, is not an infringement of copyright", where fair use is determined by "the purpose and character of the use, including whether such use is of a commercial nature or is for nonprofit educational purposes" and "the effect of the use upon the potential market for or value of the copyrighted work."

\textbf{Dataset Intended Usage and License.} The full details of the \dataset dataset are presented in this paper, and both the \dataset and code for reproducing results will be made publicly available. The \dataset dataset is not supposed to be used to train models for cheating. The primary goal is to support the research community in benchmarking and advancing physical reasoning in LLMs and MLLMs. We take full responsibility for any rights violation that may arise. Both the \dataset data and our open-source code are released under the MIT license.

\section{Broader Impacts}
\label{sec:broader_impacts}

Our benchmark aims to advance the evaluation of MLLMs in the domain of physical reasoning. By focusing on realistic visual scenarios grounded in physics, we hope to contribute toward the development of AI systems with stronger scientific reasoning capabilities, which is an essential step for applications in education, science tutoring, and automated scientific discovery. In particular, this benchmark may support the design of models that assist learners in understanding complex physical concepts through both text and visuals.

Potential negative impacts are limited but worth noting. First, as our dataset is curated entirely in English, it may not generalize well to non-English-speaking contexts, inadvertently reinforcing language bias. Then, the scenarios in our dataset are schematic rather than real-world images, which may limit generalization to real-world physical perception tasks.

We believe these concerns are manageable and do not diminish the broader positive potential of the benchmark in promoting robust, multimodal physical reasoning in foundation models.

% \clearpage
\section{More Dataset Details}

\subsection{Question Distribution}

\begin{figure*}[t!]
\centering
% \vspace{0.2cm}
\includegraphics[width=0.9\textwidth]{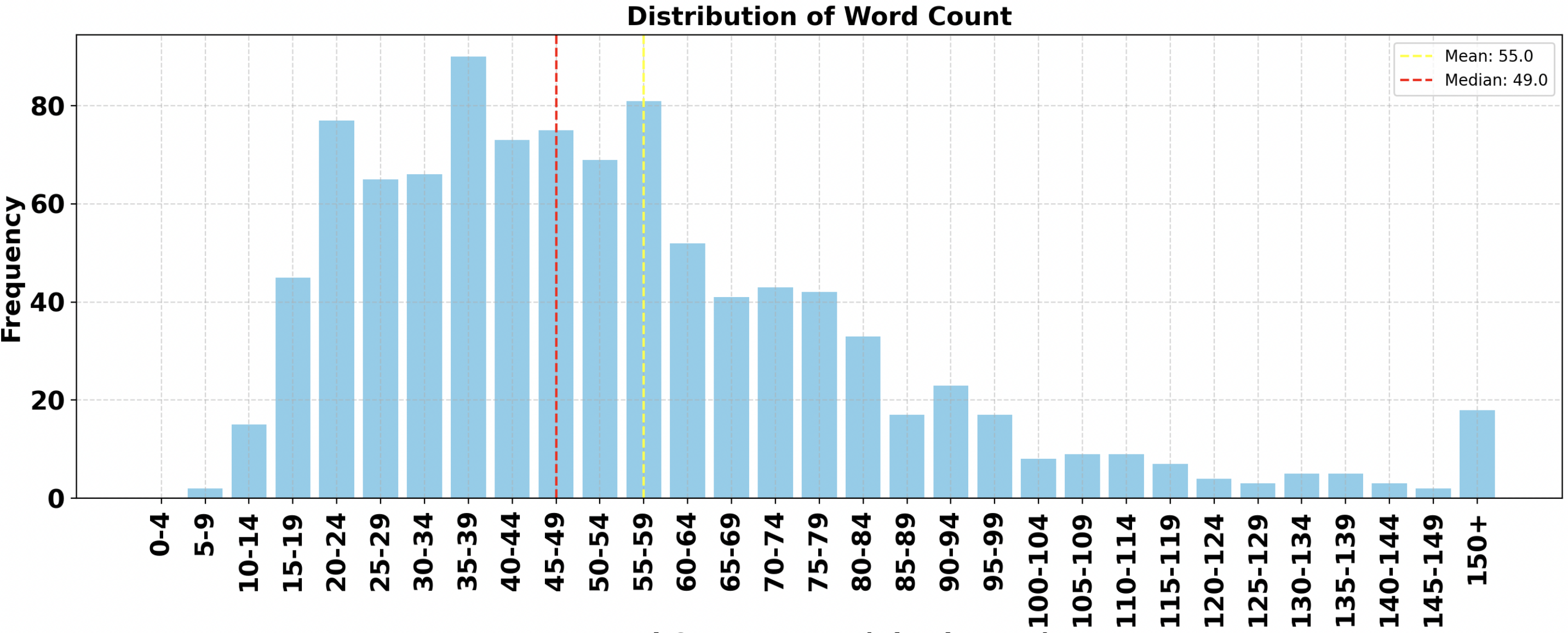}
% \vspace{0.02cm}
   \caption{The distribution of the number of words per question in \dataset.}
\label{fig:word_count_distribution}
% \vspace{-0.2cm}
\end{figure*}

All questions in \dataset are written in English. Figure~\ref{fig:word_count_distribution} presents the distribution of word counts of questions in Text-DeRedundancy setting, demonstrating the variation in question lengths. The similarity between the median and average word counts suggests a roughly symmetrical distribution. 

\subsection{Introduction of Domain and Subfield}
\label{appendix:intro_domains}

As shown in Table~\ref{tab: subfields}, \dataset covers 6 core domains and 25 subdomains.

\textbf{Mechanics.} Mechanics is the branch of physics concerned with the motion of objects and the forces that cause or change this motion. It encompasses both classical mechanics and key subfields such as \textit{Kinematics} (e.g., velocity, acceleration, free fall), \textit{Dynamics} (e.g., Newton’s laws, force analysis, friction), \textit{Work and Energy} (e.g., work-energy theorem, mechanical energy conservation), \textit{Momentum and Collisions} (e.g., conservation of momentum, elastic and inelastic collisions), \textit{Rotational Motion} (e.g., torque, angular acceleration, moment of inertia), and \textit{Statics} (e.g., torque balance, structural analysis). Mechanics lays the groundwork for much of physics, enabling the understanding of how and why objects move or remain at rest in various physical systems.

\textbf{Electromagnetism.} Electromagnetism explores the interactions between electric charges and magnetic fields. It includes the subfields of \textit{Electrostatics} (e.g., Coulomb’s law, electric fields and potential), \textit{Electric Circuits} (e.g., Ohm’s law, circuit analysis, RC circuits), \textit{Magnetism} (e.g., magnetic fields, Lorentz force, Ampère’s law), \textit{Electromagnetic Induction} (e.g., Faraday’s law, Lenz’s law, inductance), and optionally, \textit{Maxwell’s Equations and Electromagnetic Waves} for advanced topics. This domain underpins much of modern technology, including electric circuits, motors, and wireless transmission.

\textbf{Thermodynamics.} Thermodynamics is the study of heat, energy, and their transformations. Its subtopics include \textit{Temperature and Heat Transfer} (e.g., conduction, convection, radiation), \textit{Specific Heat and Calorimetry} (e.g., phase changes, heat calculations), \textit{Laws of Thermodynamics} (e.g., energy conservation, entropy), and \textit{Ideal Gases and Kinetic Theory} (e.g., gas laws, internal energy, pressure). This domain is central to engines, thermal systems, and understanding natural processes.

\textbf{Wave/Acoustics.} This domain investigates wave behavior and sound phenomena. Core subfields include \textit{Wave Properties} (e.g., speed, frequency, wavelength, interference), \textit{Sound} (e.g., pitch, loudness, Doppler effect, standing waves), and \textit{Resonance and Harmonics} (e.g., resonant frequencies, vibrations in strings and air columns). These concepts are crucial in fields ranging from acoustics to telecommunications.

\textbf{Optics.} Optics studies the behavior and properties of light. It includes \textit{Geometrical Optics} (e.g., reflection, refraction, lens imaging, total internal reflection), \textit{Wave Optics} (e.g., interference, diffraction, polarization), and \textit{Optical Instruments} (e.g., microscopes, telescopes, image formation). Optics has broad applications in imaging, vision science, and photonics.

\textbf{Modern Physics.} Modern Physics addresses phenomena beyond the scope of classical mechanics. Its key subfields include \textit{Relativity} (e.g., time dilation, mass-energy equivalence), \textit{Quantum Phenomena} (e.g., photoelectric effect, atomic models), \textit{Nuclear Physics} (e.g., radioactivity, nuclear reactions, mass defect), and optionally \textit{Particle Physics} (e.g., elementary particles, the Standard Model). These topics form the theoretical basis of contemporary physics and technology.

\subsection{Images by Domains}
\begin{figure*}[!htbp]
    \centering
\includegraphics[width=1\linewidth]{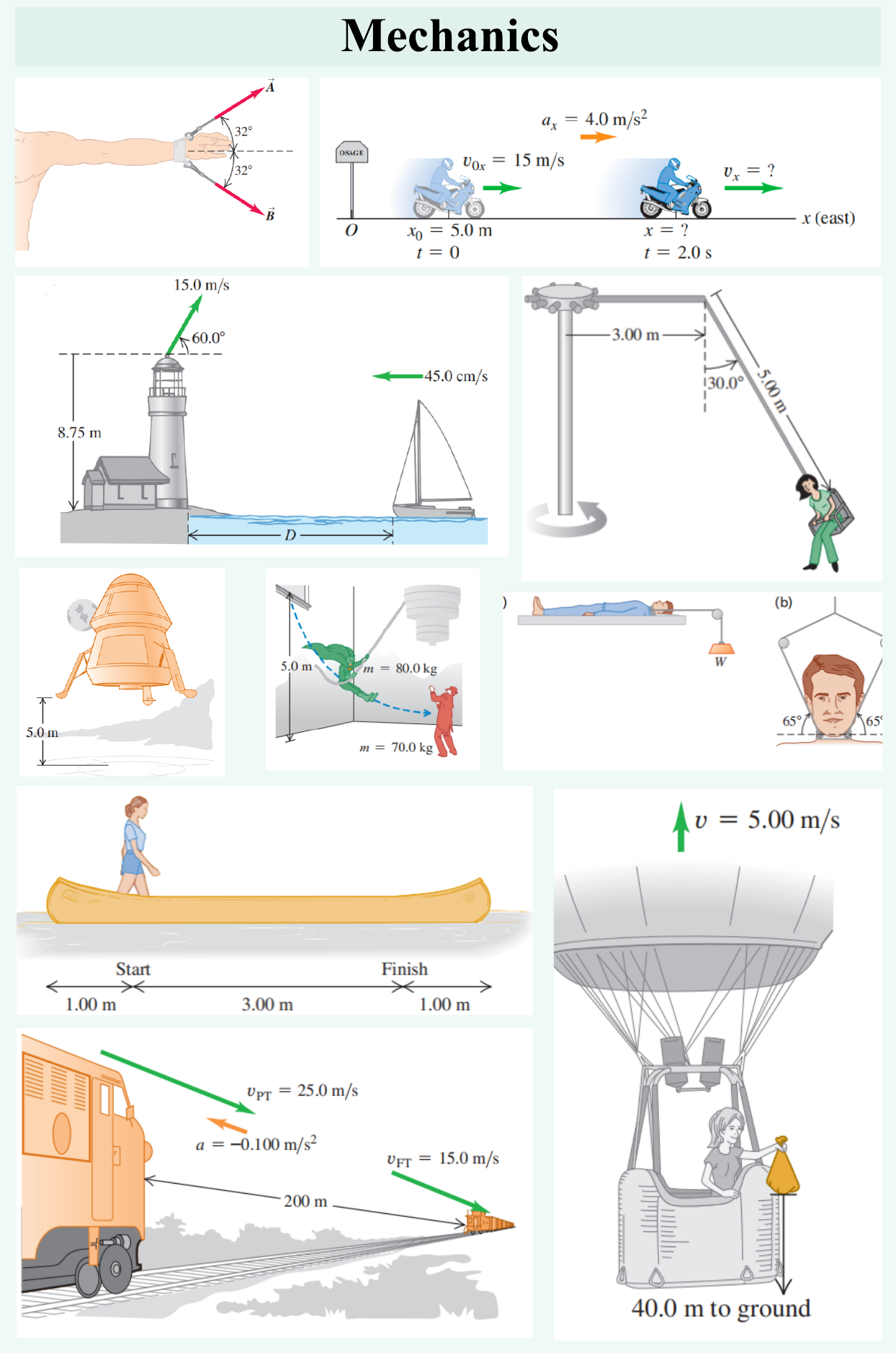}
    \caption{Examples of the visual context for the \textit{Mechanics} domain.}
\label{fig:examples_mechanics}
\end{figure*}

\begin{figure*}[!htbp]
    \centering
\includegraphics[width=1\linewidth]{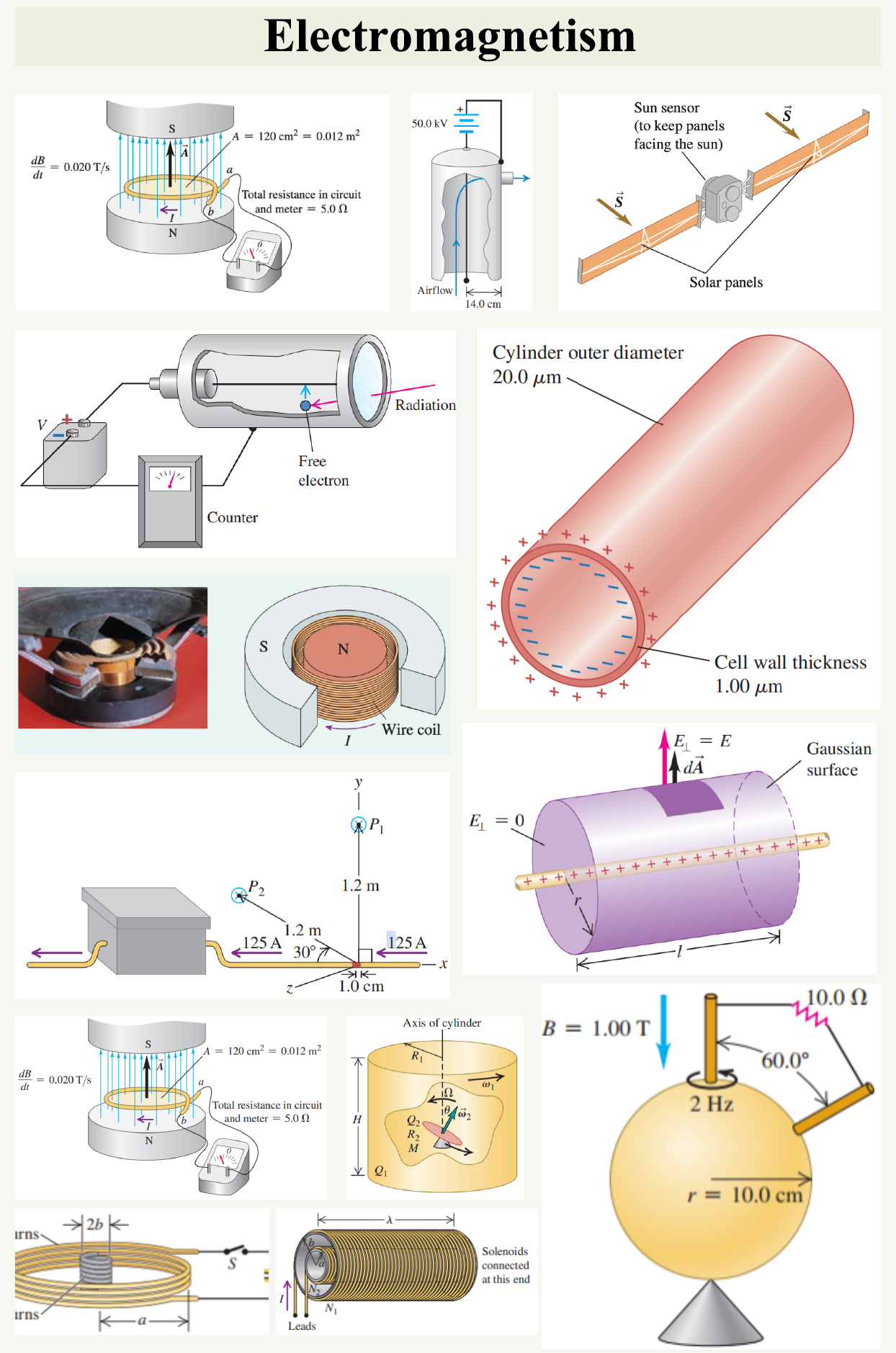}
    \caption{Examples of the visual context for the \textit{Electromagnetism} domain.}
\label{fig:examples_electromagnetism}
\end{figure*}

\begin{figure*}[!htbp]
    \centering
\includegraphics[width=1\linewidth]{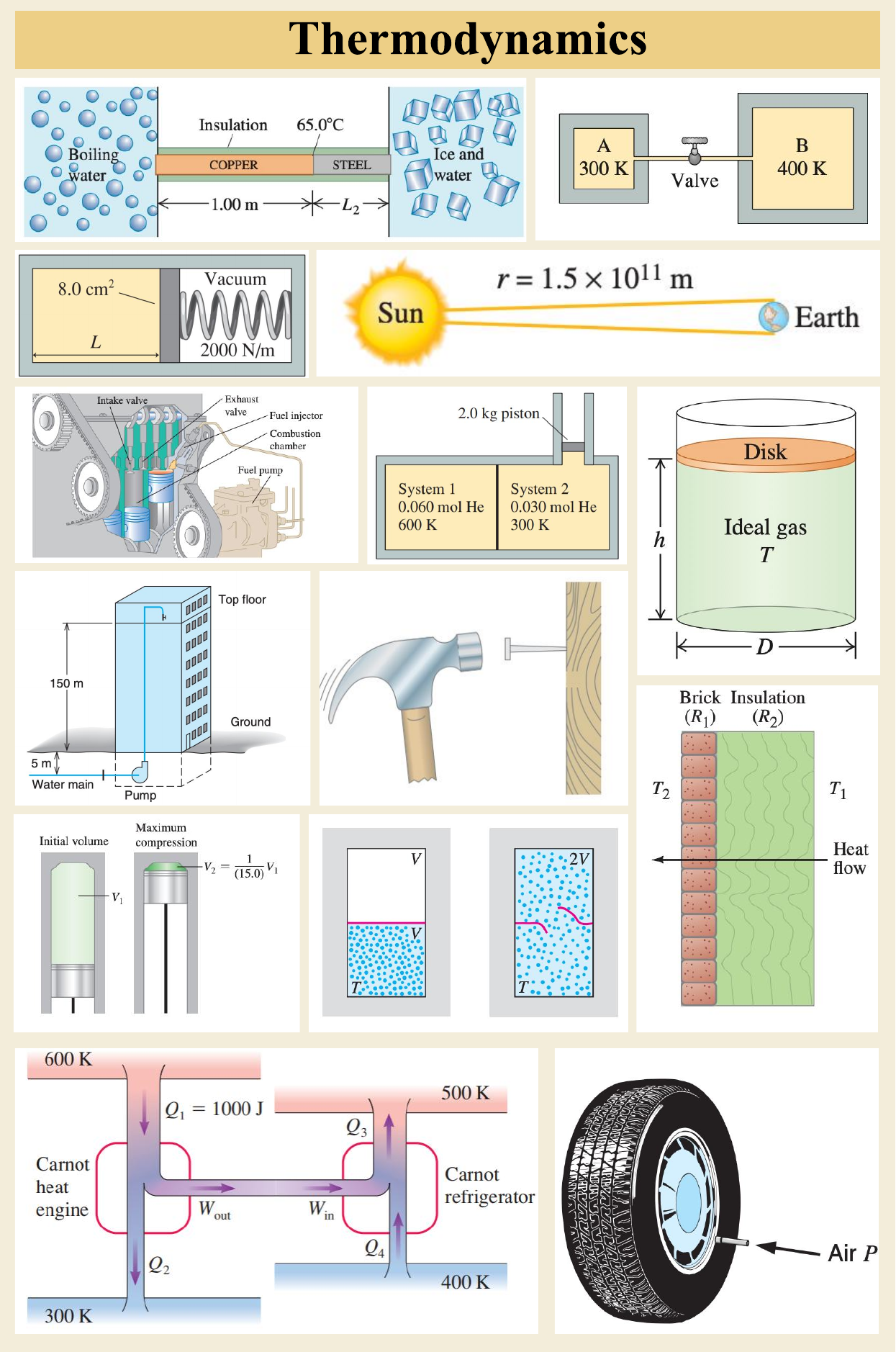}
    \caption{Examples of the visual context for the \textit{Thermodynamics} domain.}
\label{fig:examples_thermodynamics}
\end{figure*}

\begin{figure*}[!htbp]
    \centering
\includegraphics[width=1\linewidth]{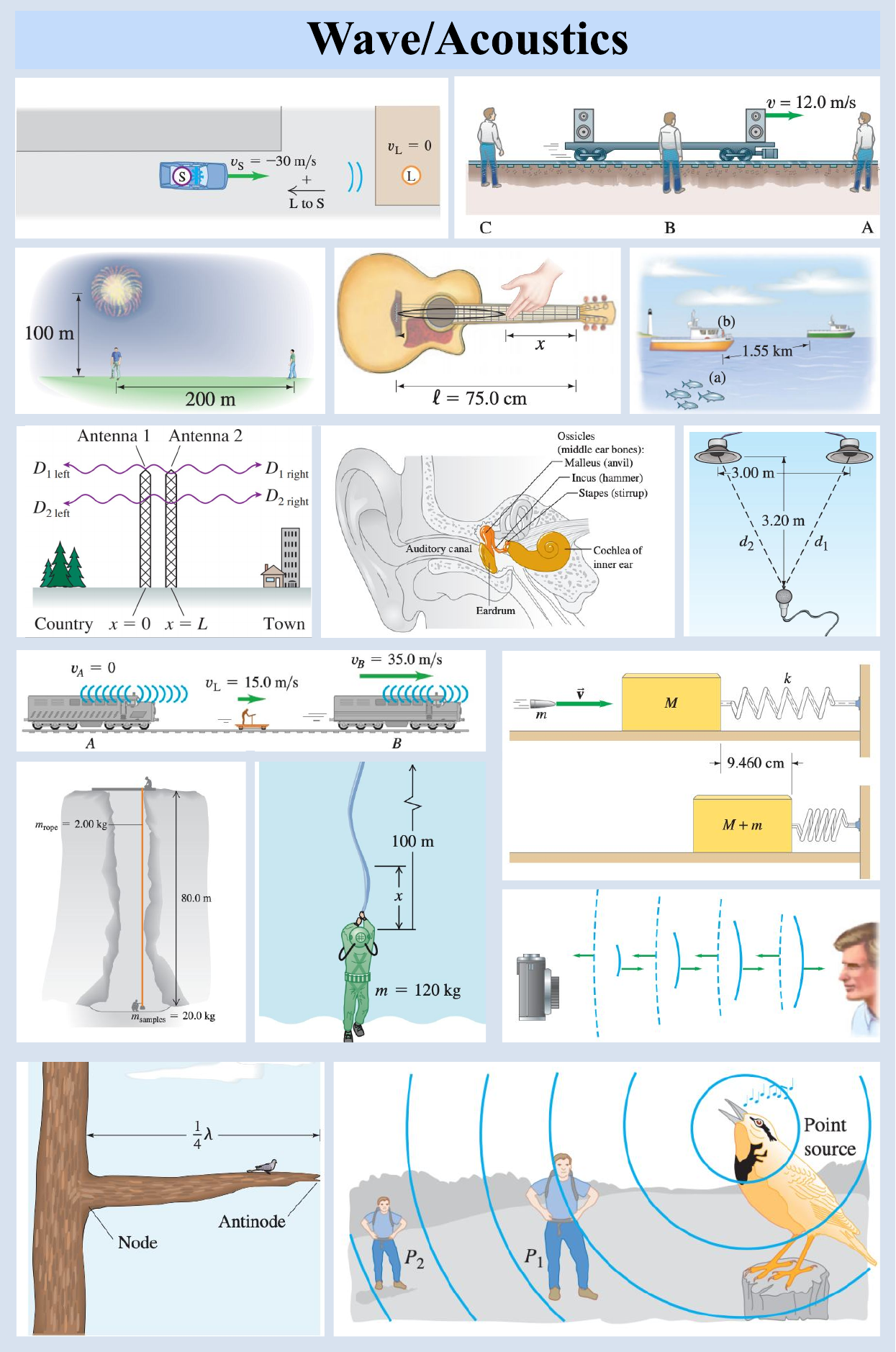}
    \caption{Examples of the visual context for the \textit{Wave/Acoustics} domain.}
\label{fig:examples_wave}
\end{figure*}

\begin{figure*}[!htbp]
    \centering
\includegraphics[width=1\linewidth]{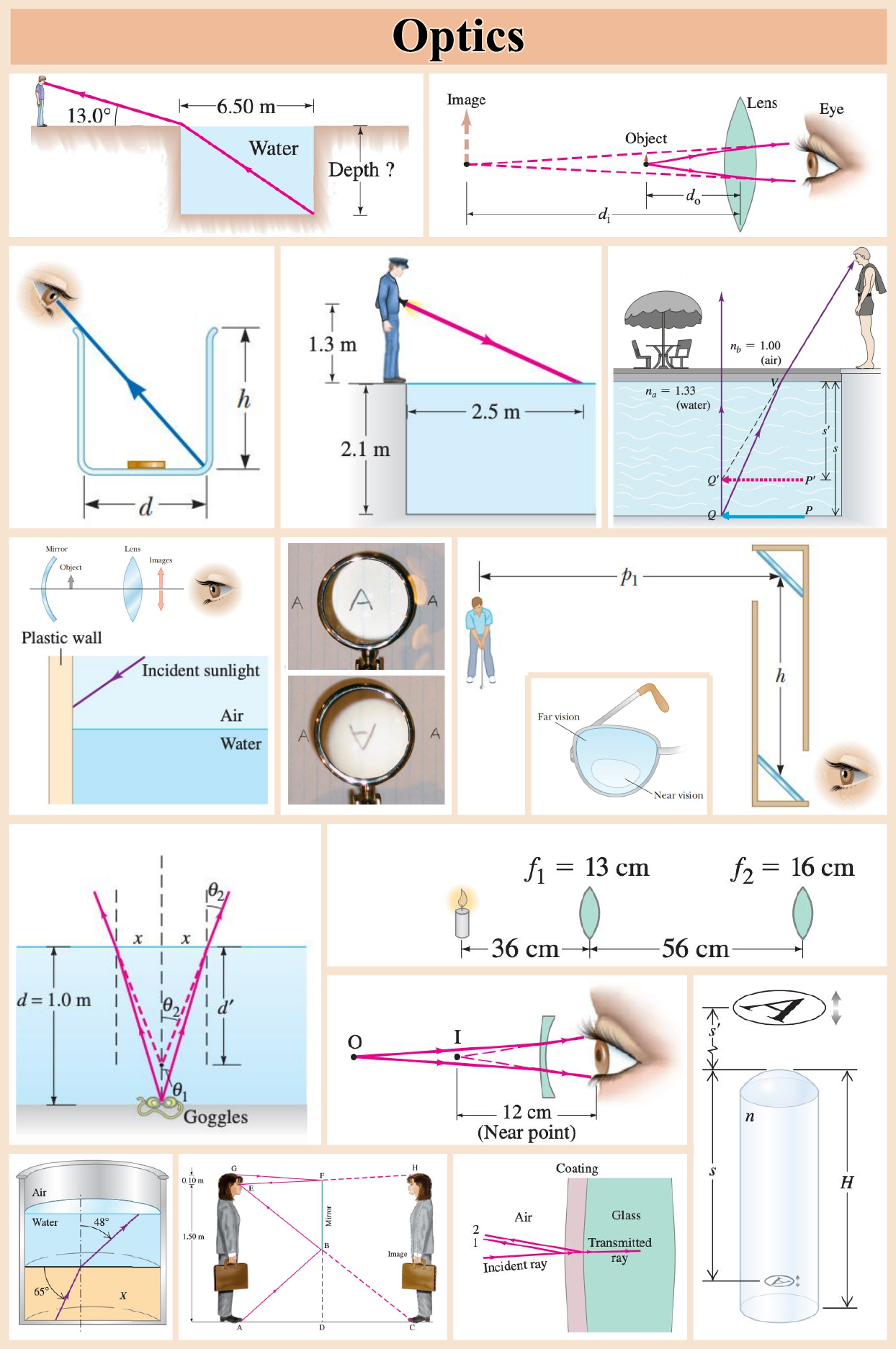}
    \caption{Examples of the visual context for the \textit{Optics} domain.}
\label{fig:examples_optics}
\end{figure*}

\begin{figure*}[!htbp]
    \centering
\includegraphics[width=1\linewidth]{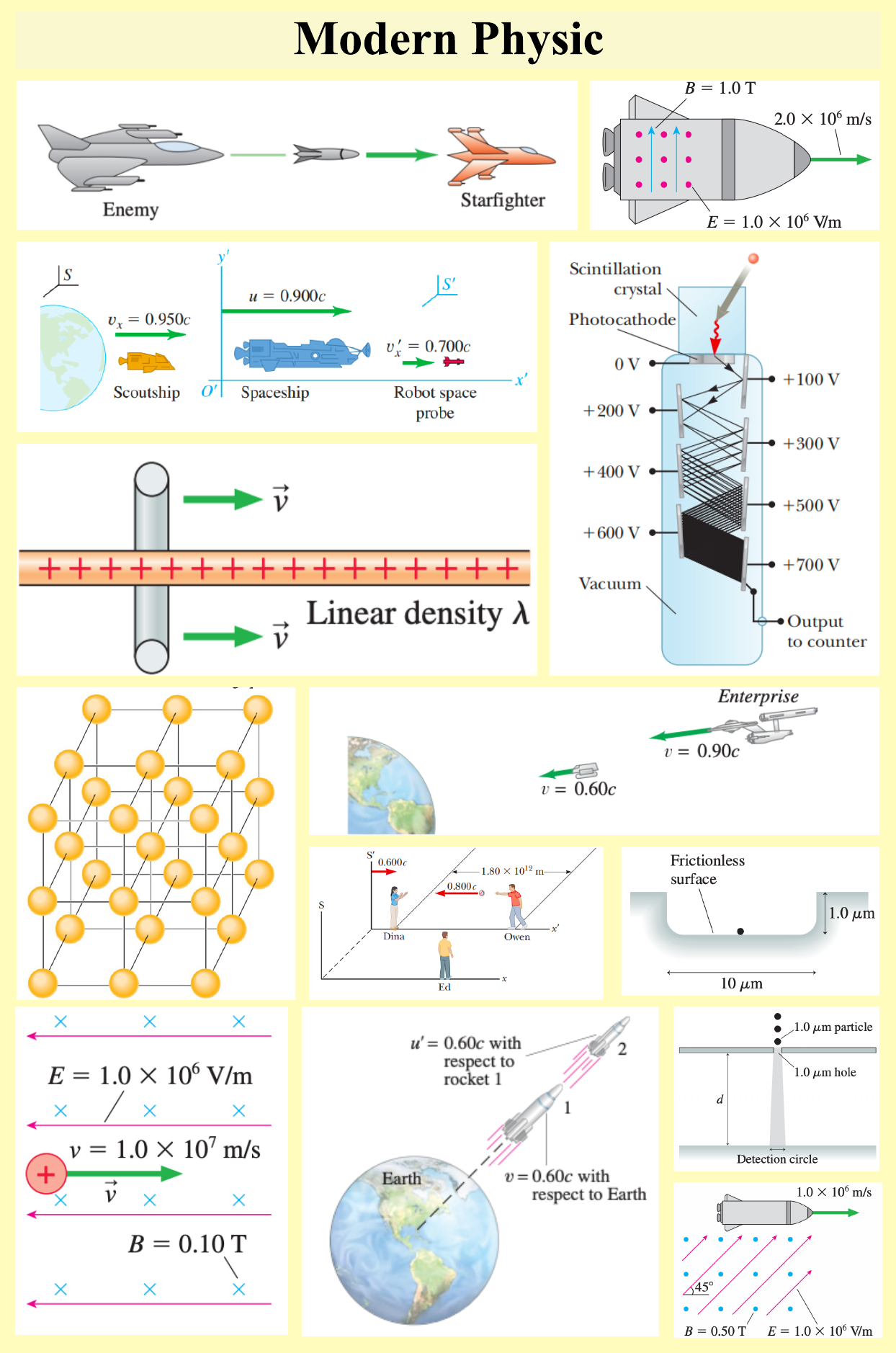}
    \caption{Examples of the visual context for the \textit{Modern Physics} domain.}
\label{fig:examples_modern}
\end{figure*}

In this section, we present images example from the physics problems in \dataset. 
Figure~\ref{fig:examples_mechanics}, Figure~\ref{fig:examples_electromagnetism}, Figure~\ref{fig:examples_thermodynamics}, Figure~\ref{fig:examples_wave}, Figure~\ref{fig:examples_optics} and Figure~\ref{fig:examples_modern} show images from the problems under the category of Mechanics, Electromagnetism, Thermodynamics, Wave/Acoustics, Optics, Modern Physics, respectively.

We observe that the images in our dataset are highly realistic, often depicting concrete physical scenarios rather than stylized or abstract illustrations. While they are not real-world photographs, these visuals are grounded in plausible physical settings. This realism provides essential context for physical reasoning and helps bridge the gap between abstract physics principles and their real-world manifestations.

Across domains, the visual characteristics vary in alignment with the nature of the physical concepts. Despite their domain-specific variations, a unifying theme across all categories is the consistent use of realistic and context-rich imagery, which provides essential grounding for physical interpretation and distinguishes our benchmark from other datasets with overly synthetic or schematic visual content.

\begin{table*}
    \centering
    \begin{tabular}{>{\centering\arraybackslash}p{0.21\textwidth} p{0.70\textwidth}}
    \toprule
    \textbf{Domain} & \multicolumn{1}{c}{\textbf{Subfields}} \\
    \midrule
    Optics & Optical Instrument, Wave Optics, and Geometrical Optics \\
    \midrule
    Electromagnetism & Electromagnetic Wave, Electric Circuits, Magnetism, Electromagnetic Induction, and Electrostatics  \\
    \midrule
    Mechanics & Momentum and Collisions, Work and Energy, Statics, Dynamics, Relational Motion, and Kinematics. \\
    \midrule
    Wave/Acoustics & Sound, Resonance and Harmonics, and Wave Properties \\
    \midrule
    Thermodynamics & Specific Heat and Calorimetry, Temperature and Heat Transfer, Ideal Gases and Kinetic Theory, and Laws of Thermodynamics \\
    \midrule
    Modern Physics & Particle Physics, Nuclear Physics, Relativity, and Quantum Phenomena  \\
    \bottomrule
    \end{tabular}
    \caption{Subfields included in each domain in \dataset.}
    \label{tab: subfields}
\end{table*}

\subsection{Physical Reasoning Definition}
\label{sec:a_physical_reasoning_definition}

Six physical reasoning types are defined in Table~\ref{tab:a_reasoning_type}.

% \clearpage
\section{More Evaluation Details}
\label{appendix:evaluation_details}
We conduct all experiments on NVIDIA A100 80G GPUs.

\subsection{CoT Prompting for Generating Answer}
\label{appendix:cot_prompt}

The CoT prompting for generating answer is shown in Figure~\ref{appendix:cot_prompt}.

\begin{figure}
    \centering
    \includegraphics[width=0.8\linewidth]{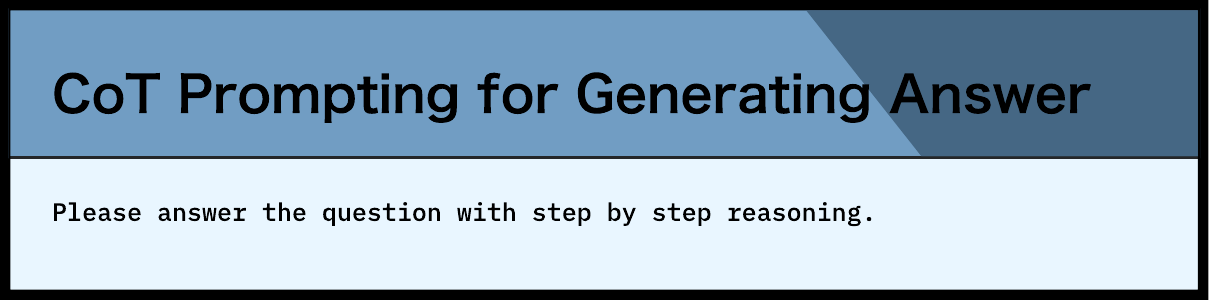}
    \caption{CoT prompting for generating answer.}
    \label{fig:a_CoT_Prompting_for_Generating_Answer}
\end{figure}

\subsection{Rule-based Answer Extraction}
\label{appendix:rule_extraction}

\begin{figure}
    \centering
    \includegraphics[width=0.8\linewidth]{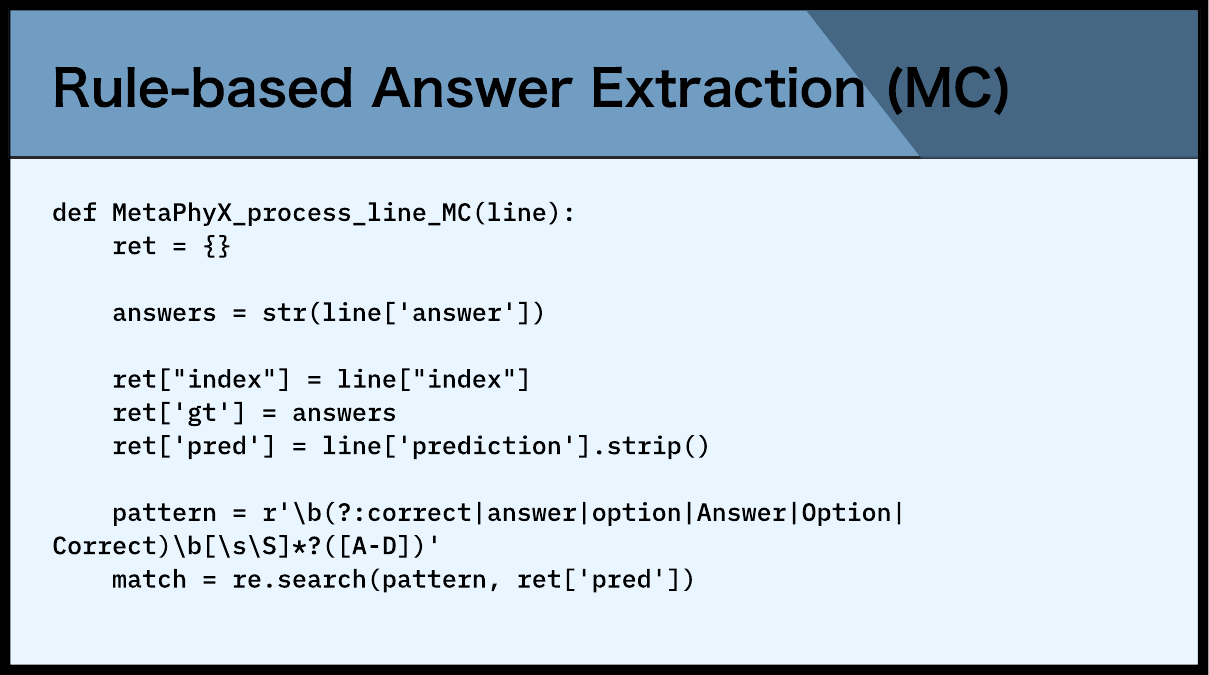}
    \caption{Rule-based answer extraction strategy for MC questions.}
    \label{fig:a_Rule_based_Answer_Extraction_MC}
\end{figure}

\begin{figure}
    \centering
    \includegraphics[width=0.8\linewidth]{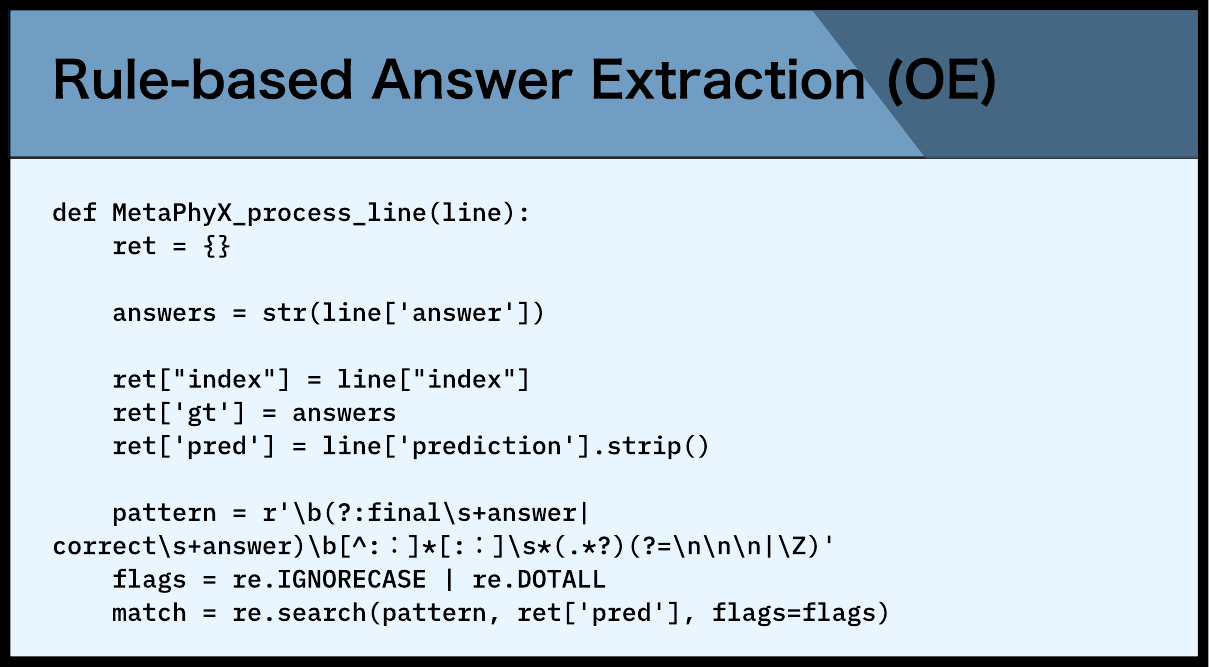}
    \caption{Rule-based answer extraction strategy for OE questions.}
    \label{fig:a_Rule_based_Answer_Extraction_OE}
\end{figure}

The rule-based answer extraction strategies for MC and OE questions are shown in Figure~\ref{fig:a_Rule_based_Answer_Extraction_MC} and Figure~\ref{fig:a_Rule_based_Answer_Extraction_OE}, respectively.

\subsection{Prompt for Answer Judge}
\label{appendix:prompt_answer_judge}

\begin{figure}
    \centering
    \includegraphics[width=0.8\linewidth]{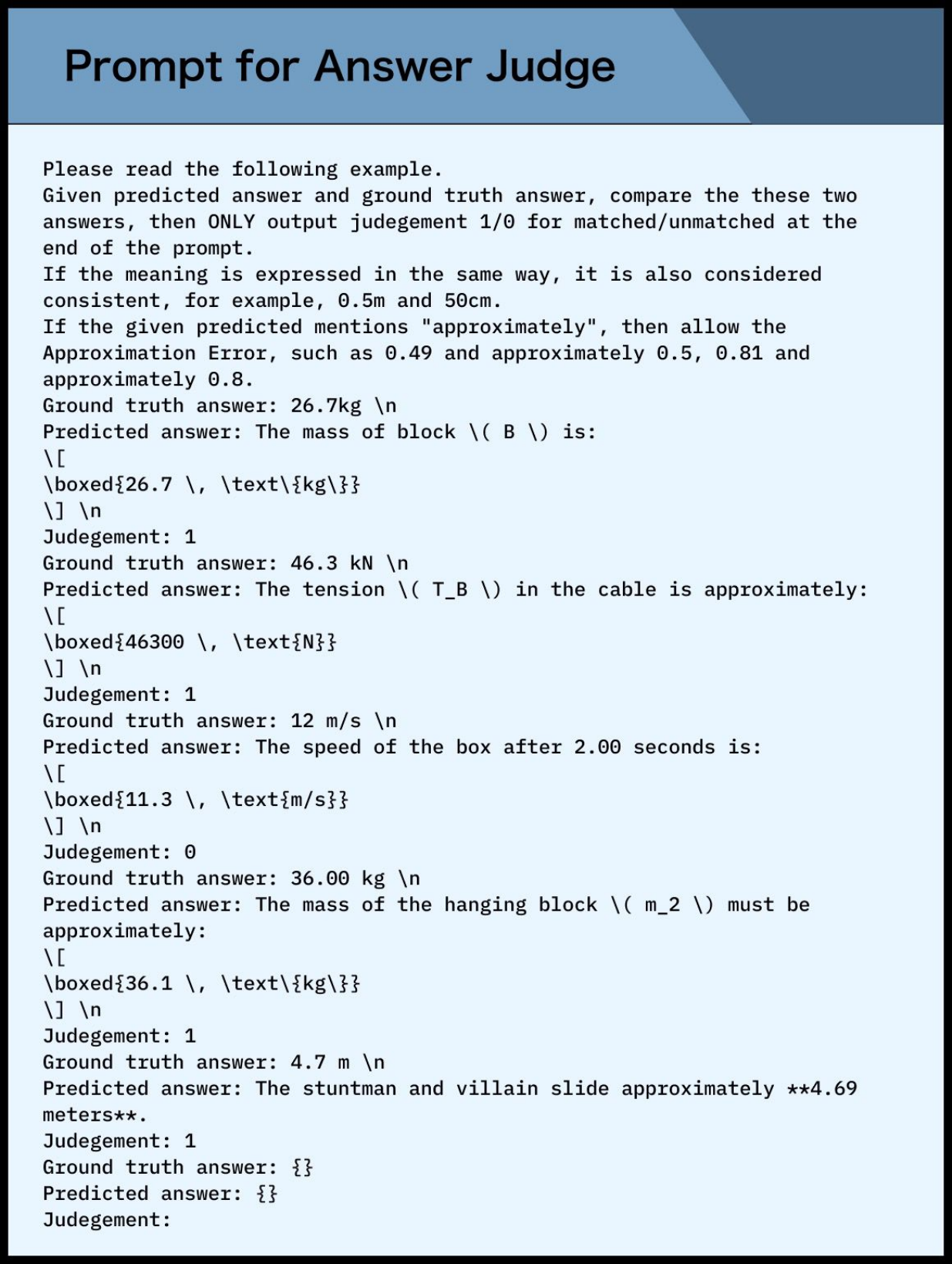}
    \caption{Rule-based answer extraction strategy for OE questions.}
    \label{fig:a_fig_prompt_for_answer_judge}
\end{figure}

The prompt for answer judge is shown in Figure~\ref{fig:a_fig_prompt_for_answer_judge}.

% \subsection{Zero-shot Results}
% \label{appendix:zero_shot_results}

% As shown in Table~\ref{appendix:tab_zero_shot}, LLM judges benefit from few-Shot prompts. When evaluating identical predictions from GPT-4o on \dataset, the judgment accuracy of the LLM judge noticeably drops from the 5-shot to the 0-shot setting. This highlights that the LLM judge relies on in-context examples to accurately assess complex multimodal reasoning, suggesting that few-shot prompting is essential for reliable evaluation.

% \input{table/a_zero_shot}

\subsection{Prompt for Caption Generation}
\label{appendix:prompt_caption_generation}

\begin{figure}
    \centering
    \includegraphics[width=0.5\linewidth]{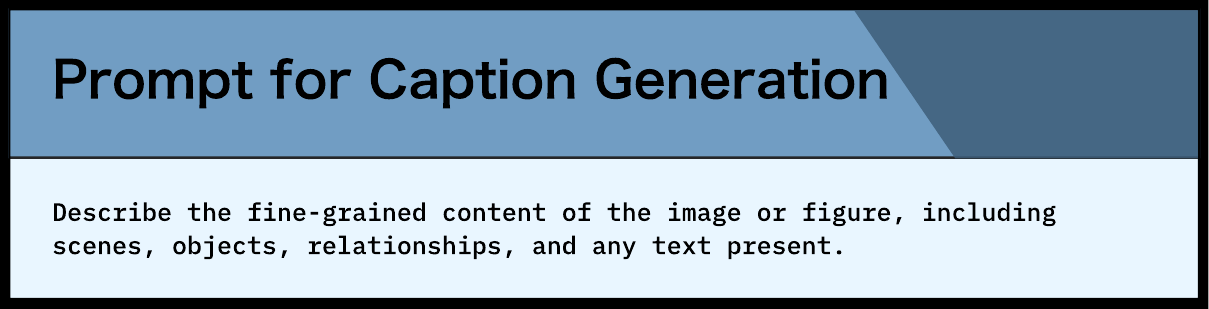}
    \caption{Prompt template for caption generation.}
    \label{fig:a_Prompt_for_Caption_Generation}
\end{figure}

The prompt for caption generation is shown in Figure~\ref{fig:a_Prompt_for_Caption_Generation}

\begin{table*}
    \centering
    \begin{tabular}{>{\centering\arraybackslash}p{0.21\textwidth} p{0.75\textwidth}}
    \toprule
    \textbf{Physical Reasoning} & \multicolumn{1}{c}{\textbf{Description}} \\
    \midrule
    Physical Model Grounding Reasoning & This reasoning involves connecting the specific details of a problem description to fundamental physical concepts, laws, and idealized models. It's the process of identifying which area of physics is relevant and selecting the appropriate simplified representations that allow the problem to be analyzed using established physical principles and equations. Essentially, it translates a real-world or described scenario into a solvable physics framework. \\
    \midrule
    Spatial Relation Reasoning & This focuses on understanding and manipulating the geometric and directional aspects of a physics problem. It involves visualizing the setup, determining the positions, orientations, distances, angles, and relative movements of objects. This often requires using coordinate systems, vectors (including resolving them into components), and geometric principles. \\
    \midrule
    Multi-Formula Reasoning & This reasoning type is required when a problem cannot be solved using a single physics equation. It involves identifying multiple relevant formulas or principles and understanding how they interrelate. The process typically involves using the output of one formula as the input for another, or setting up and solving a system of simultaneous equations derived from different physical laws. \\
    \midrule
    Implicit Condition Reasoning & This involves recognizing and utilizing information or constraints that are not explicitly stated in the problem text but are implied by the context, standard physics assumptions, or specific keywords. Examples include understanding that "starts from rest" means the initial velocity is zero, a "smooth" surface implies zero friction, a "light string" or "light pulley" means its mass is negligible, or that an object reaching its maximum height has a momentary vertical velocity of zero. \\
    \midrule
    Numerical Reasoning & This reasoning refers to problems where solving requires the application of advanced mathematical methods beyond basic algebra and trigonometry. This includes techniques such as calculus, solving differential equations that model the system, vector calculus, Fourier analysis, linear algebra for complex systems, or other higher-level mathematical procedures necessary to manipulate the physical formulas and arrive at a solution. This applies when the mathematical technique itself is a core part of solving the physics, regardless of whether the final answer is purely numerical or symbolic. \\
    \midrule
    Predictive Reasoning & This involves using established physical laws and the initial conditions of a system to forecast its future state or behavior. Based on the principles governing the situation, you calculate or deduce what will happen after a certain time or interaction. Examples include predicting the trajectory of a projectile, the final temperature of a mixture after thermal equilibrium is reached, or the velocity of objects after a collision. \\
    \bottomrule
    \end{tabular}
    \caption{Definitions of six physical reasoning categories in \dataset.}
    \label{tab:a_reasoning_type}
\end{table*}

\subsection{Prompt for Reasoning Type Labeling}
\label{appendix:prompt_reasoning_type}
\begin{figure}
    \centering
    \includegraphics[width=0.5\linewidth]{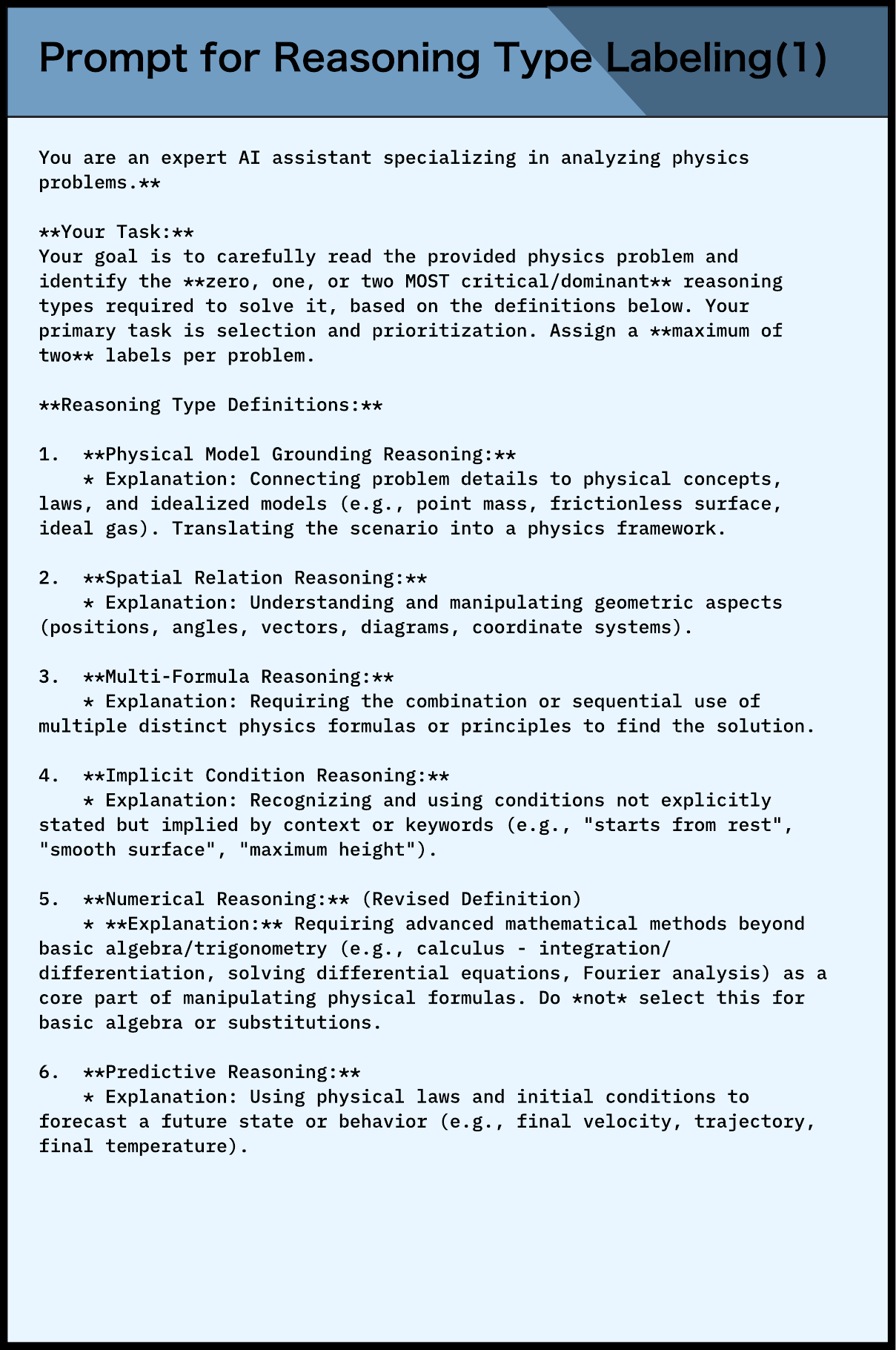}
    \caption{Prompt for reasoning type labeling (1).}
    \label{fig:a_Prompt_for_Reasoning_Type_Labeling_1}
\end{figure}

\begin{figure}
    \centering
    \includegraphics[width=0.5\linewidth]{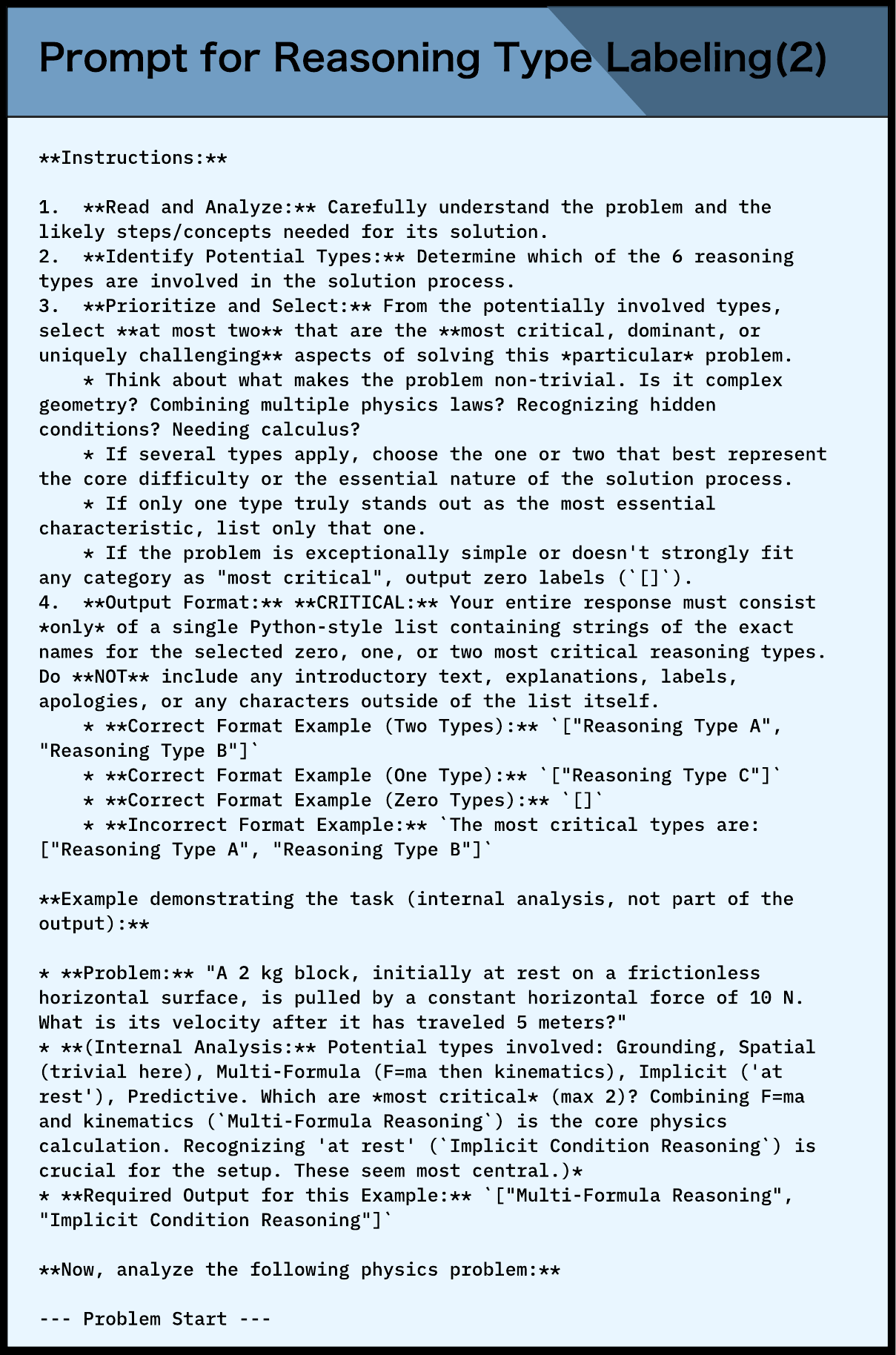}
    \caption{Prompt for reasoning type labeling (2).}
    \label{fig:a_Prompt_for_Reasoning_Type_Labeling_2}
\end{figure}

The prompt for reasoning type labeling is shown in Figure~\ref{fig:a_Prompt_for_Reasoning_Type_Labeling_1} and Figure~\ref{fig:a_Prompt_for_Reasoning_Type_Labeling_2}

\clearpage
\section{Case Study}
\label{appendix:case_study}
\phantomsection
\label{list:list_of_figures}
\listofappfigures

% \clearpage
\renewcommand{\thefigure}{\arabic{figure}}
\setcounter{figure}{0}  % 

\begin{table*}[!htbp]
    \centering
    \begin{tabular}{l l >{\centering\arraybackslash}m{2.8cm} >{\centering\arraybackslash}m{2.8cm} >{\centering\arraybackslash}m{2.8cm}}

        \toprule
        Domain &
        \makecell{Correct} & 
        \makecell{Visual\\Reasoning Error} & 
        \makecell{Text\\Reasoning Error} & 
        \makecell{Lack of\\Knowledge} \\
        \toprule
        Mechanics & \ref{fig:mechanics_1}, \ref{fig:mechanics_2} &
        \ref{fig:mechanics_3} & \ref{fig:mechanics_4} & 
        \ref{fig:mechanics_5} \\
        
        Electromagnetism & \ref{fig:electromagnetism_1}, \ref{fig:electromagnetism_2} & 
        \ref{fig:electromagnetism_3} & 
        \ref{fig:electromagnetism_4} & 
        \ref{fig:electromagnetism_5} \\
        
        Thermodynamics & \ref{fig:thermodynamics_1}, \ref{fig:thermodynamics_2} & 
        \ref{fig:thermodynamics_3} & 
        \ref{fig:thermodynamics_4} & 
        \ref{fig:thermodynamics_5} \\
        
        Wave/Acoustics & \ref{fig:wave_1}, \ref{fig:wave_2} & 
        \ref{fig:wave_3} & \ref{fig:wave_4} & \ref{fig:wave_5} \\
         
        Optics & \ref{fig:optics_1}, \ref{fig:optics_2} & 
        \ref{fig:optics_3} & \ref{fig:optics_4} & \ref{fig:optics_5} \\
        
        Modern Physics & \ref{fig:modern_1}, \ref{fig:modern_2} & 
        \ref{fig:modern_3} & \ref{fig:modern_4} & \ref{fig:modern_5} \\
        \bottomrule
    \end{tabular}
    \caption{Table index of case study figures by domains with associated error categories.}
    \label{tab:list_of_case_study_figures}
\end{table*}

\clearpage

\begin{figure*}[!htbp]
    \centering
\includegraphics[width=1\linewidth]{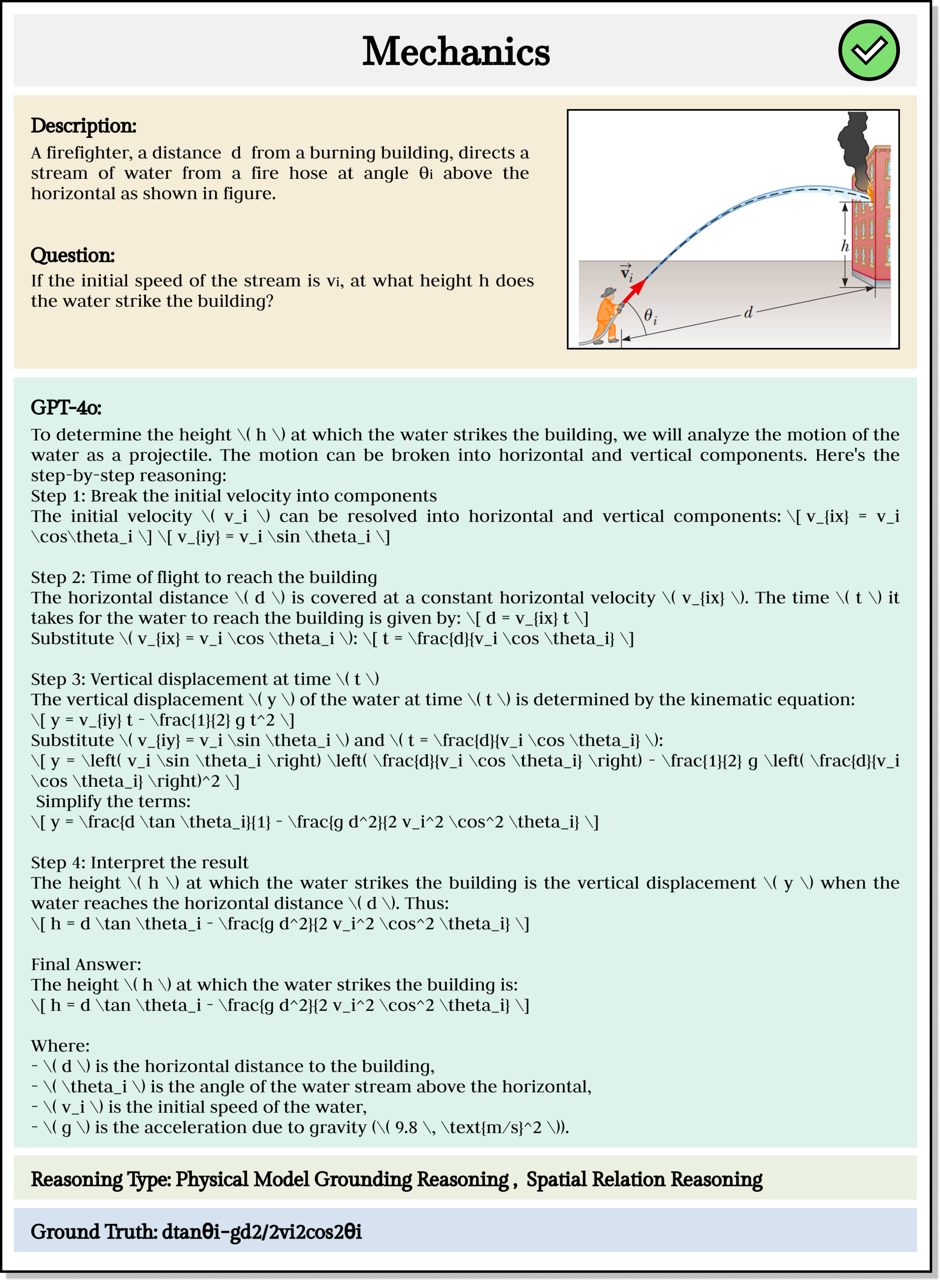}
    \caption{A sample correct case of Mechanics.\\ \hyperref[list:list_of_figures]{Back to List of Figures} \textcolor{red}{$|$} \hyperref[tab:list_of_case_study_figures]{Back to Table Index}}
    \addcontentsline{afg}{appfigures}{\protect\numberline{\thefigure}Mechanics 1: Correct Case}
\label{fig:mechanics_1}
\end{figure*}
\newpage

\begin{figure*}[!htbp]
    \centering
\includegraphics[width=1\linewidth]{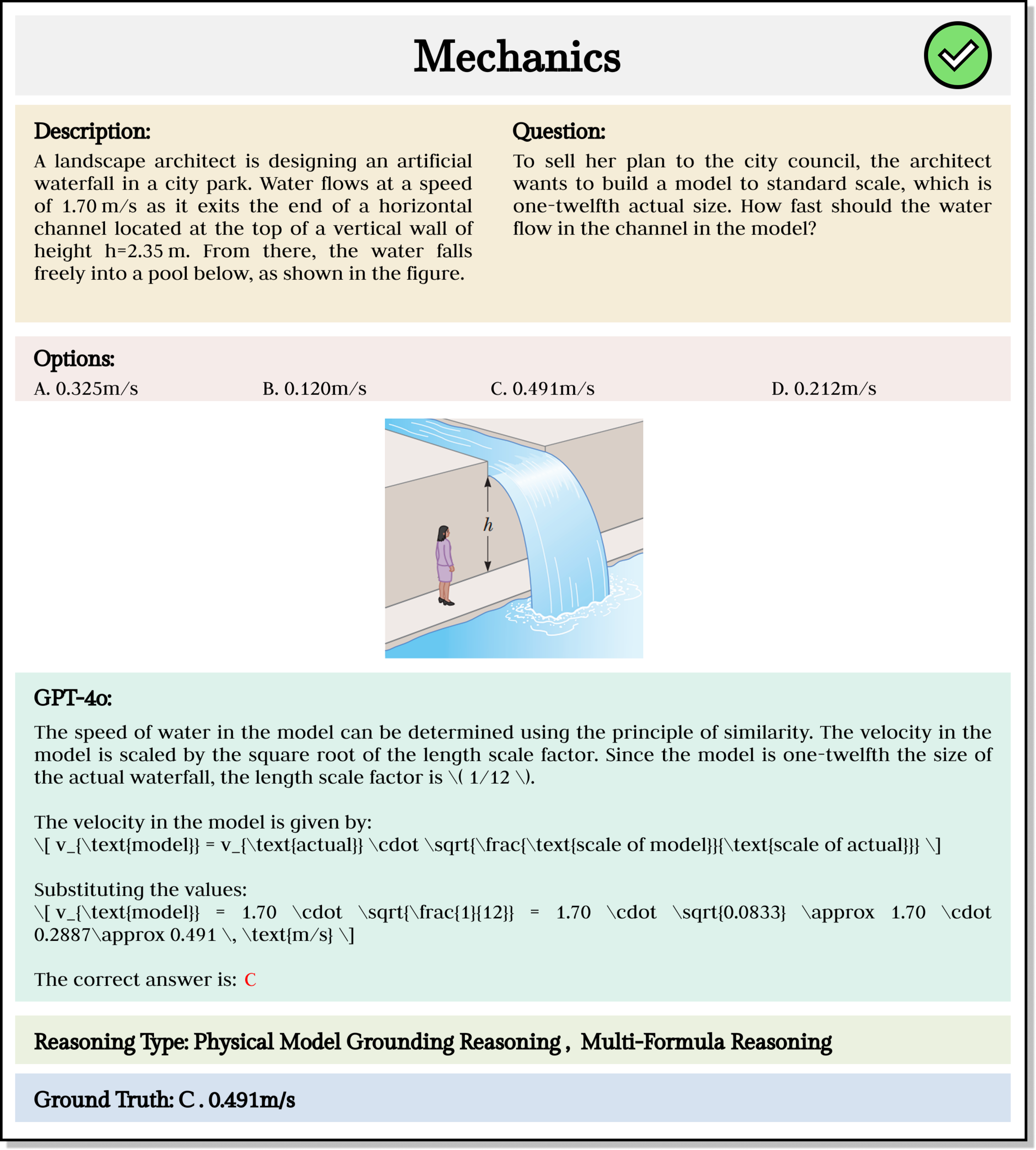}
    \caption{A sample correct case of Mechanics.\\ \hyperref[list:list_of_figures]{Back to List of Figures} \textcolor{red}{$|$} \hyperref[tab:list_of_case_study_figures]{Back to Table Index}}
    \addcontentsline{afg}{appfigures}{\protect\numberline{\thefigure}Mechanics 2: Correct Case}
\label{fig:mechanics_2}
\end{figure*}
\newpage

\begin{figure*}[!htbp]
    \centering
\includegraphics[width=1\linewidth]{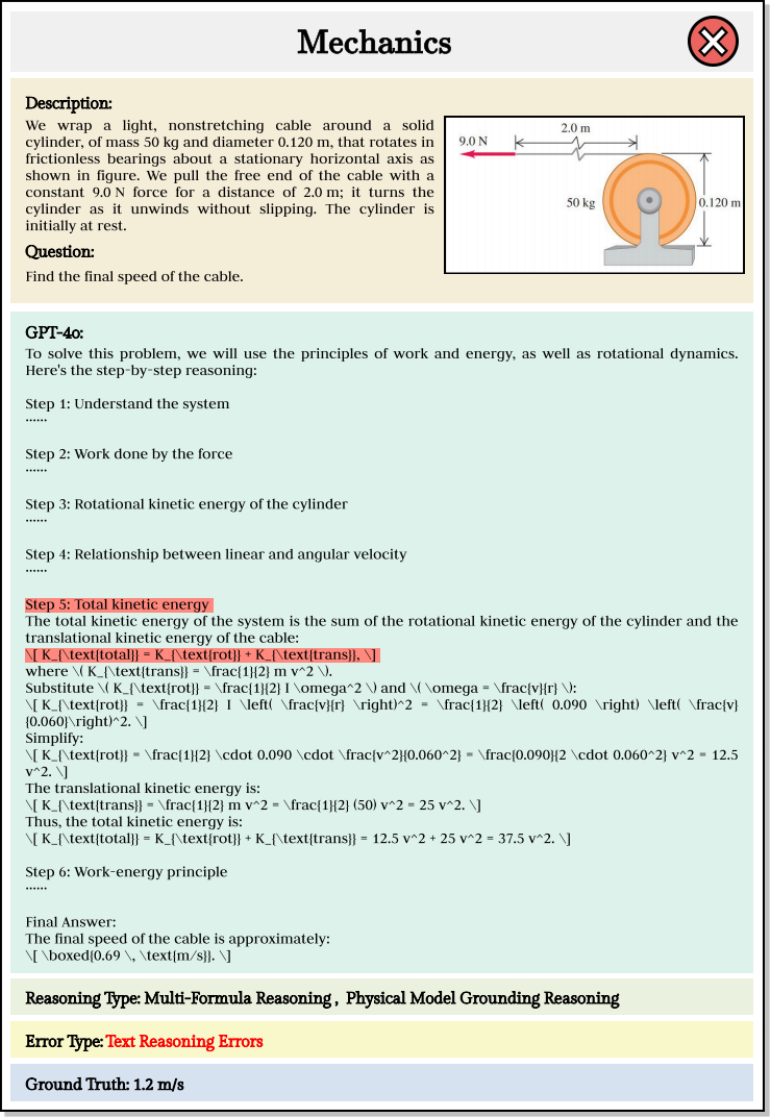}
    \caption{A sample error case of Mechanics. Error category: Visual Reasoning Error \newline \centering \hyperref[list:list_of_figures]{Back to List of Figures} \textcolor{red}{$|$} \hyperref[tab:list_of_case_study_figures]{Back to Table Index}}
    \addcontentsline{afg}{appfigures}{\protect\numberline{\thefigure}Mechanics 3: Visual Reasoning Error}
\label{fig:mechanics_3}
\end{figure*}
\newpage

\begin{figure*}[!htbp]
    \centering
\includegraphics[width=1\linewidth]{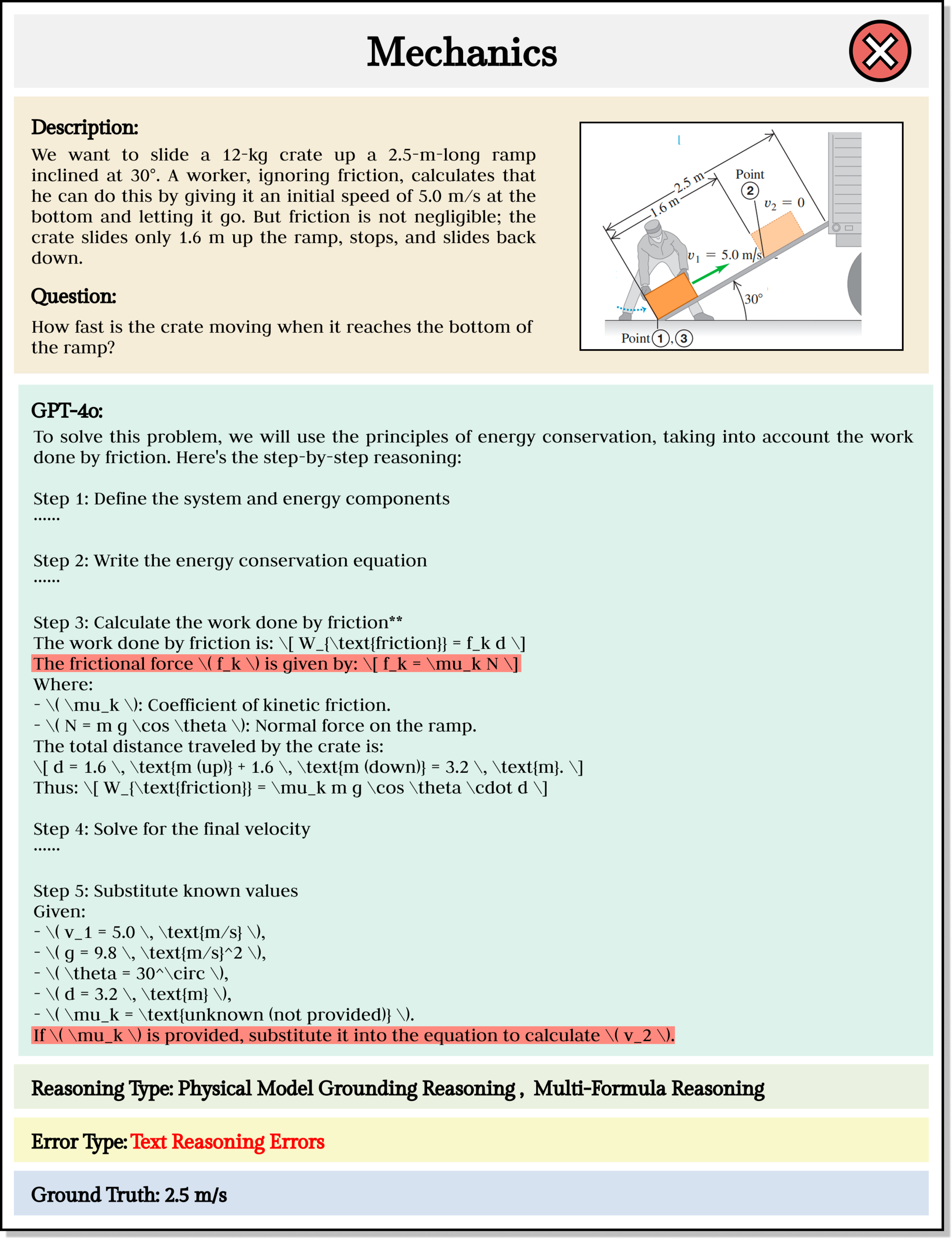}
    \caption{A sample error case of Mechanics. Error category: Text Reasoning Error \newline \centering \hyperref[list:list_of_figures]{Back to List of Figures} \textcolor{red}{$|$} \hyperref[tab:list_of_case_study_figures]{Back to Table Index}}
    \addcontentsline{afg}{appfigures}{\protect\numberline{\thefigure}Mechanics 4: Text Reasoning Error}
\label{fig:mechanics_4}
\end{figure*}
\newpage

\begin{figure*}[!htbp]
    \centering
\includegraphics[width=1\linewidth]{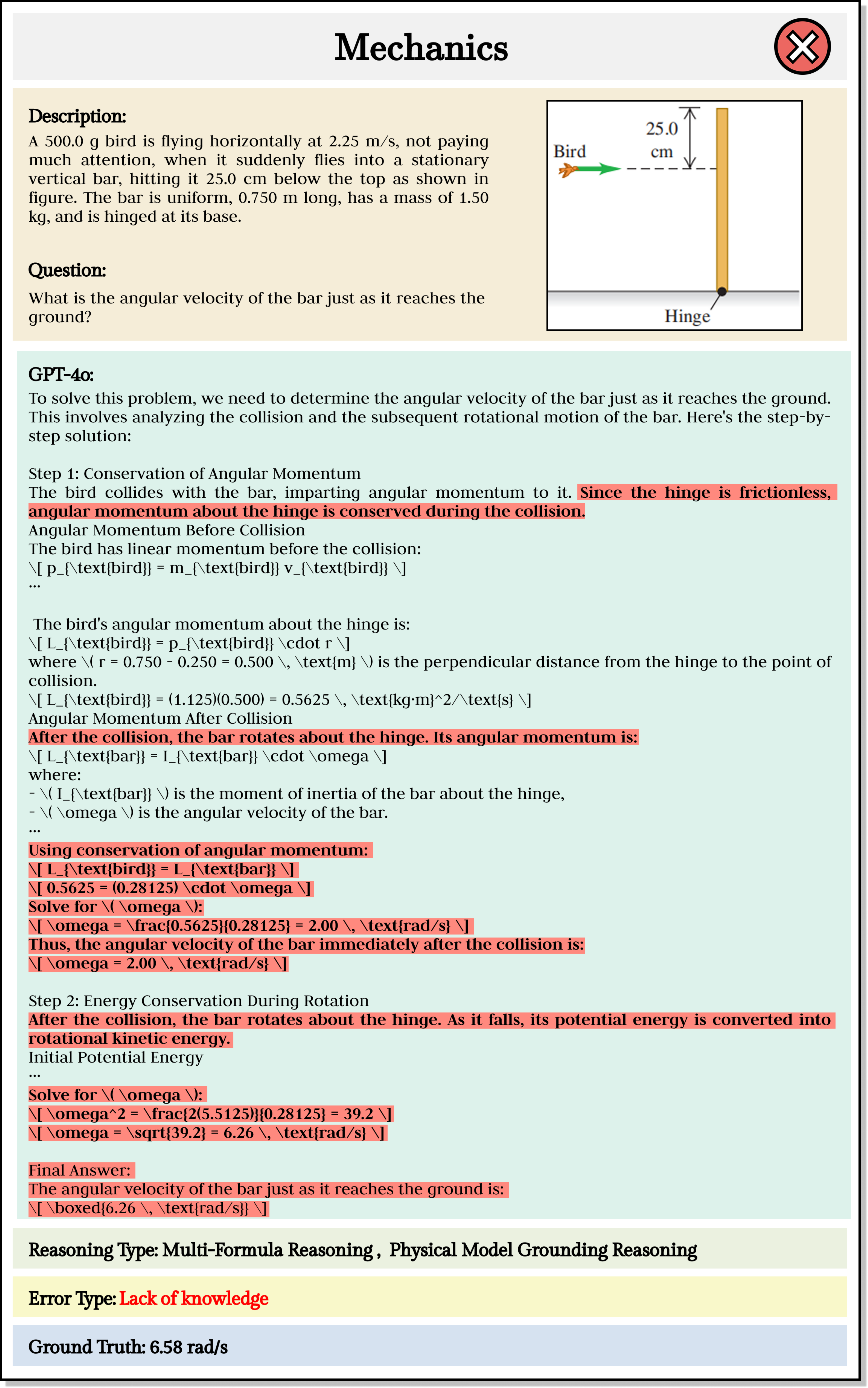}
    \caption{A sample error case of Mechanics. Error category: Lack of Knowledge \newline \centering \hyperref[list:list_of_figures]{Back to List of Figures} \textcolor{red}{$|$} \hyperref[tab:list_of_case_study_figures]{Back to Table Index}}
    \addcontentsline{afg}{appfigures}{\protect\numberline{\thefigure}Mechanics 5: Lack of Knowledge}
\label{fig:mechanics_5}
\end{figure*}
\newpage

\begin{figure*}[!htbp]
    \centering
\includegraphics[width=1\linewidth]{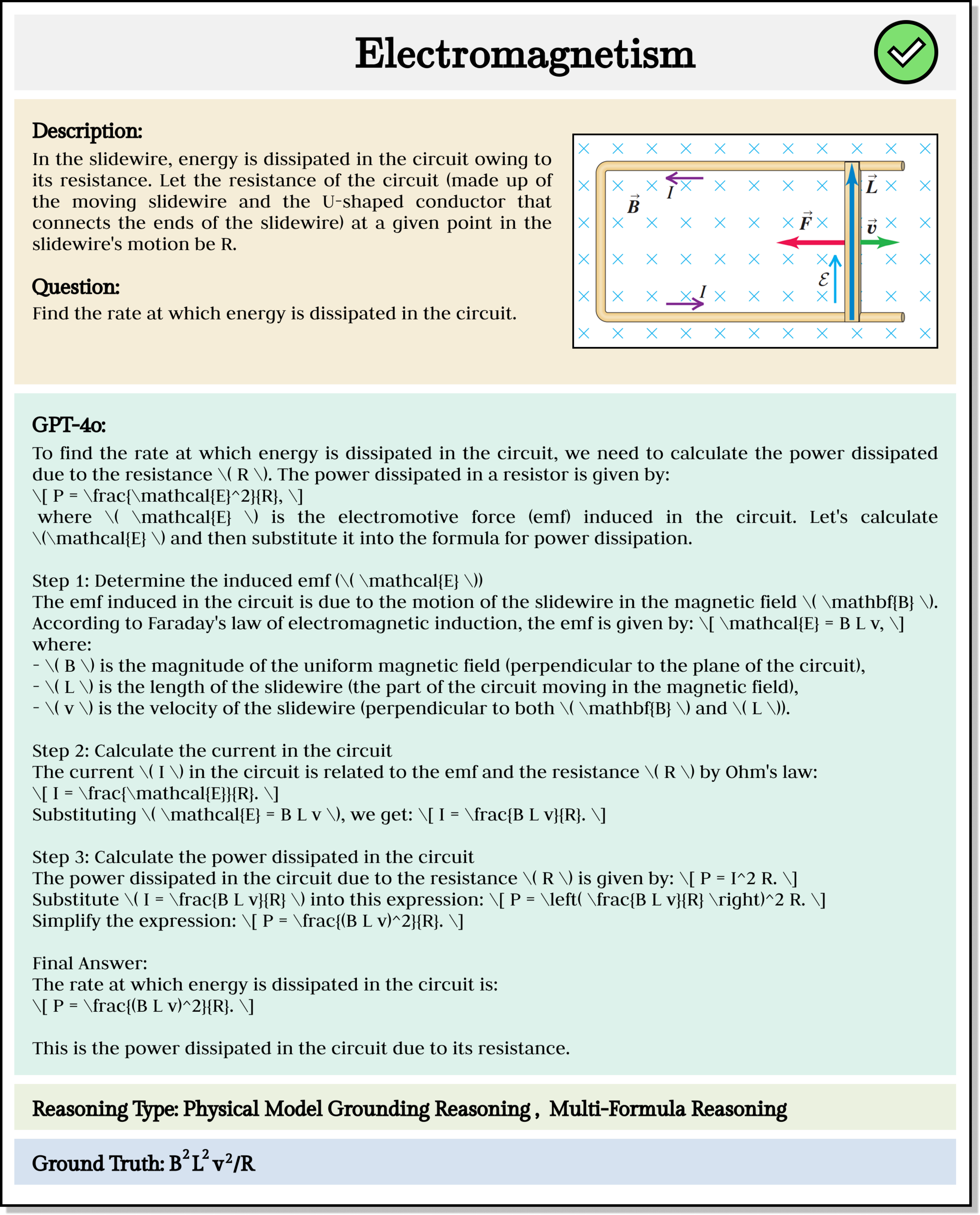}
    \caption{A sample correct case of Electromagnetism.\\ \hyperref[list:list_of_figures]{Back to List of Figures} \textcolor{red}{$|$} \hyperref[tab:list_of_case_study_figures]{Back to Table Index}}
    \addcontentsline{afg}{appfigures}{\protect\numberline{\thefigure}Electromagnetism 1: Correct Case}
\label{fig:electromagnetism_1}
\end{figure*}
\newpage

\begin{figure*}[!htbp]
    \centering
\includegraphics[width=1\linewidth]{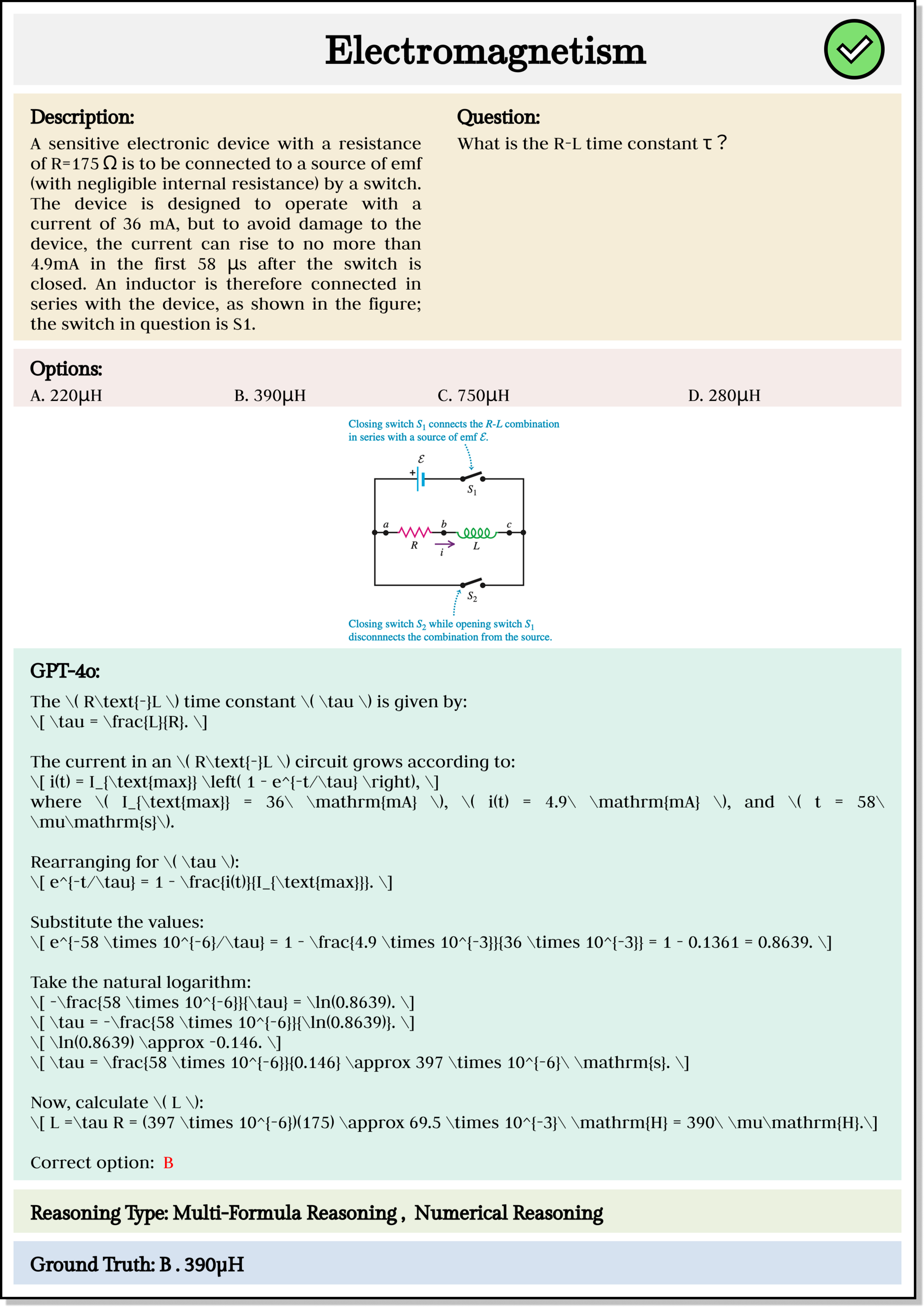}
    \caption{A sample correct case of Electromagnetism.\\ \hyperref[list:list_of_figures]{Back to List of Figures} \textcolor{red}{$|$} \hyperref[tab:list_of_case_study_figures]{Back to Table Index}}
    \addcontentsline{afg}{appfigures}{\protect\numberline{\thefigure}Electromagnetism 2: Correct Case}
\label{fig:electromagnetism_2}
\end{figure*}
\newpage

\begin{figure*}[!htbp]
    \centering
\includegraphics[width=1\linewidth]{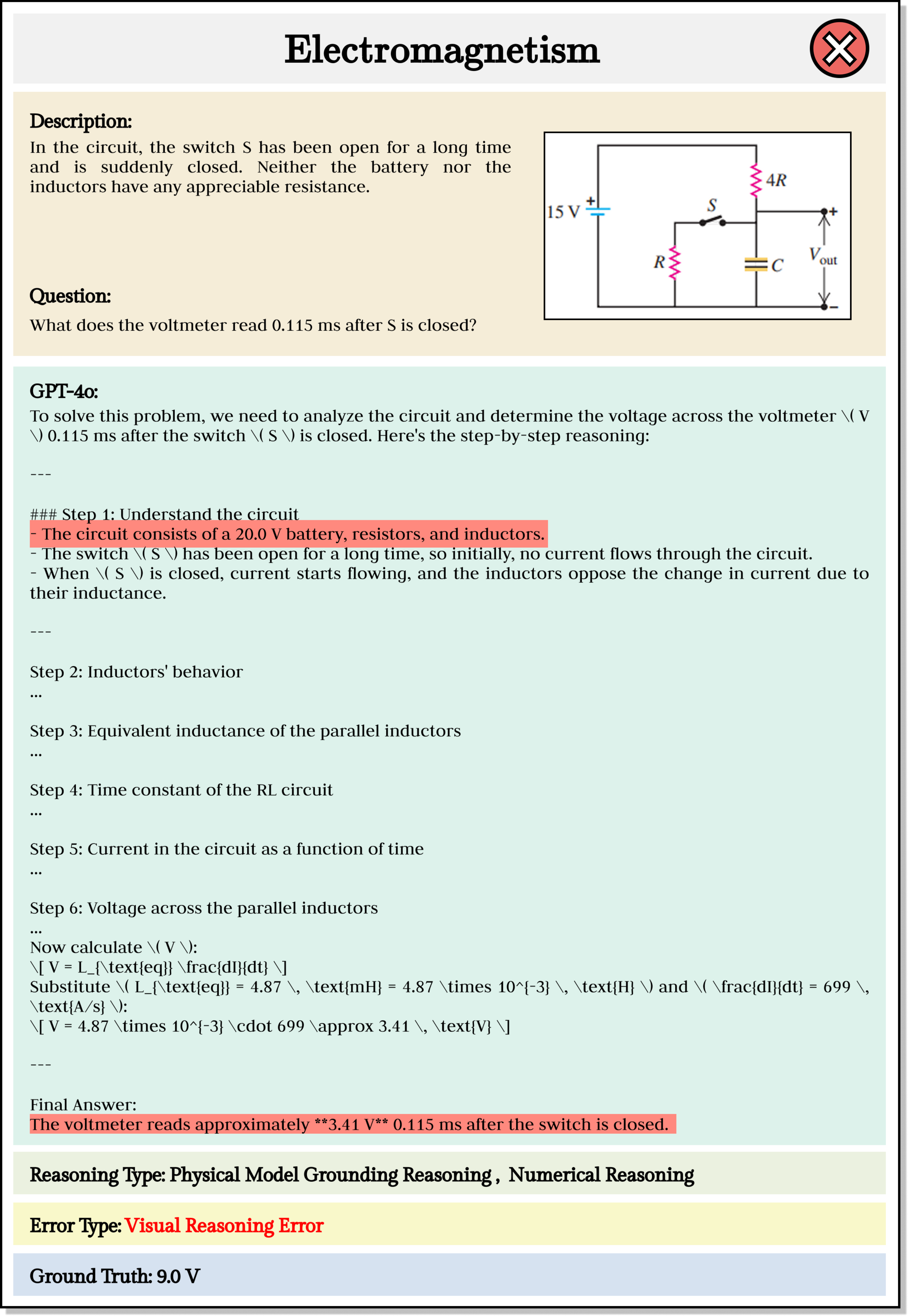}
    \caption{A sample error case of Electromagnetism. Error category: Visual Reasoning Error \newline \centering \hyperref[list:list_of_figures]{Back to List of Figures} \textcolor{red}{$|$} \hyperref[tab:list_of_case_study_figures]{Back to Table Index}}
    \addcontentsline{afg}{appfigures}{\protect\numberline{\thefigure}Electromagnetism 3: Visual Reasoning Error}
\label{fig:electromagnetism_3}
\end{figure*}
\newpage

\begin{figure*}[!htbp]
    \centering
\includegraphics[width=1\linewidth]{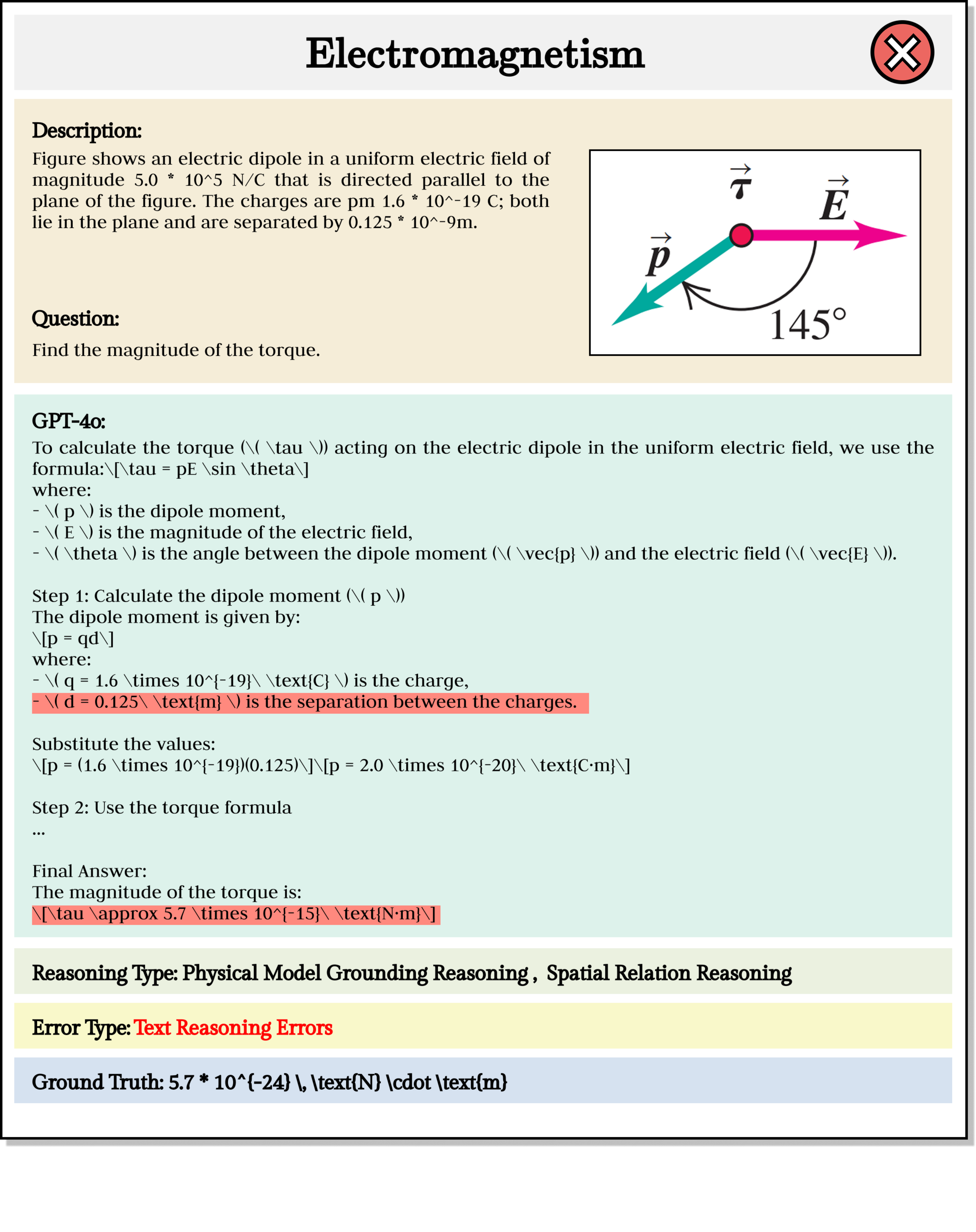}
    \caption{A sample error case of Electromagnetism. Error category: Text Reasoning Error \newline \centering \hyperref[list:list_of_figures]{Back to List of Figures} \textcolor{red}{$|$} \hyperref[tab:list_of_case_study_figures]{Back to Table Index}}
    \addcontentsline{afg}{appfigures}{\protect\numberline{\thefigure}Electromagnetism 4: Text Reasoning Error}
\label{fig:electromagnetism_4}
\end{figure*}
\newpage

\begin{figure*}[!htbp]
    \centering
\includegraphics[width=1\linewidth]{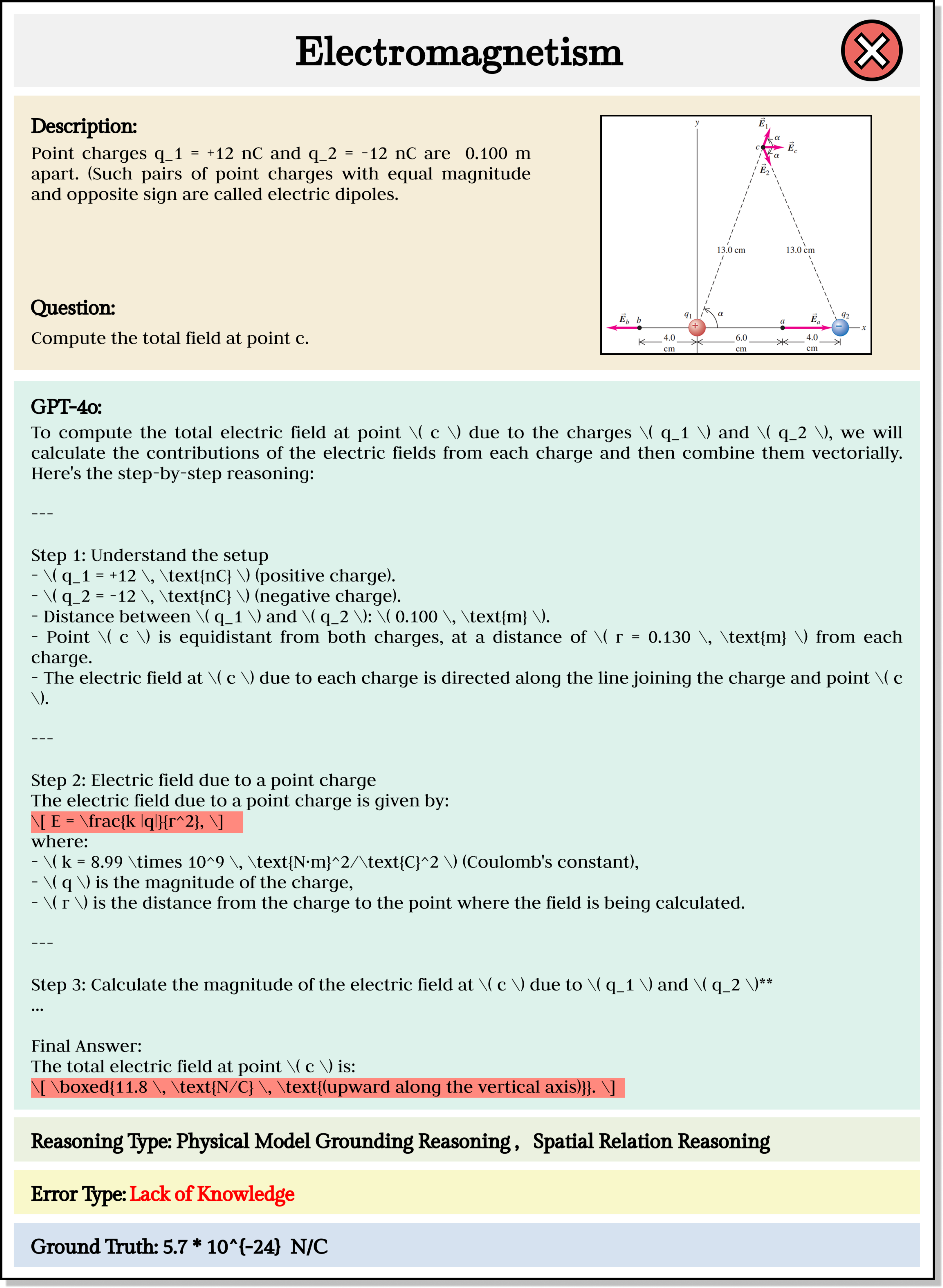}
    \caption{A sample error case of Electromagnetism. Error category: Lack of Knowledge \newline \centering \hyperref[list:list_of_figures]{Back to List of Figures} \textcolor{red}{$|$} \hyperref[tab:list_of_case_study_figures]{Back to Table Index}}
    \addcontentsline{afg}{appfigures}{\protect\numberline{\thefigure}Electromagnetism 5: Lack of Knowledge}
\label{fig:electromagnetism_5}
\end{figure*}
\newpage

\begin{figure*}[!htbp]
    \centering
\includegraphics[width=1\linewidth]{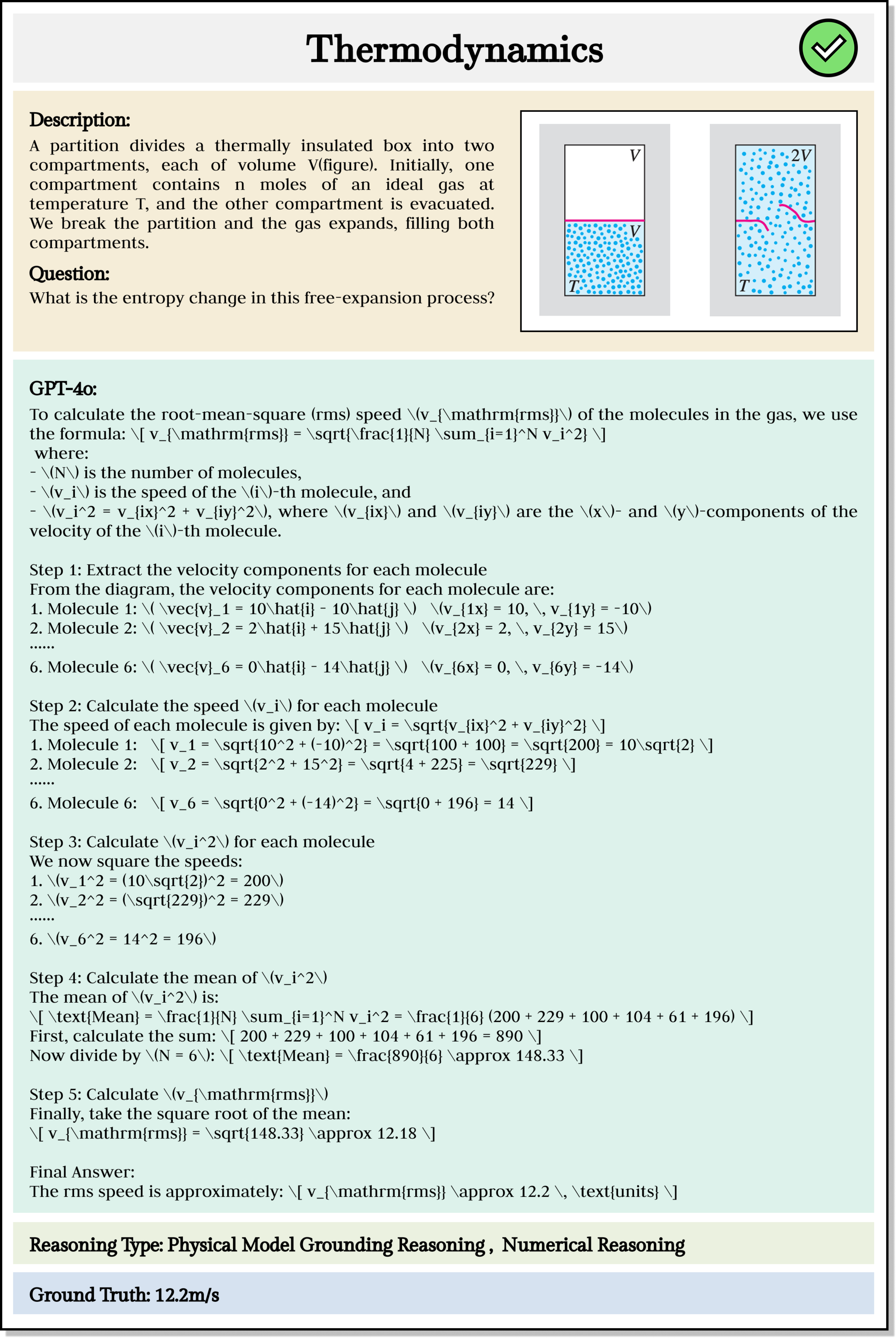}
    \caption{A sample correct case of Thermodynamics.\newline \centering \hyperref[list:list_of_figures]{Back to List of Figures} \textcolor{red}{$|$} \hyperref[tab:list_of_case_study_figures]{Back to Table Index}}
    \addcontentsline{afg}{appfigures}{\protect\numberline{\thefigure}Thermodynamics 1: Correct Case}
\label{fig:thermodynamics_1}
\end{figure*}
\newpage

\begin{figure*}[!htbp]
    \centering
\includegraphics[width=1\linewidth]{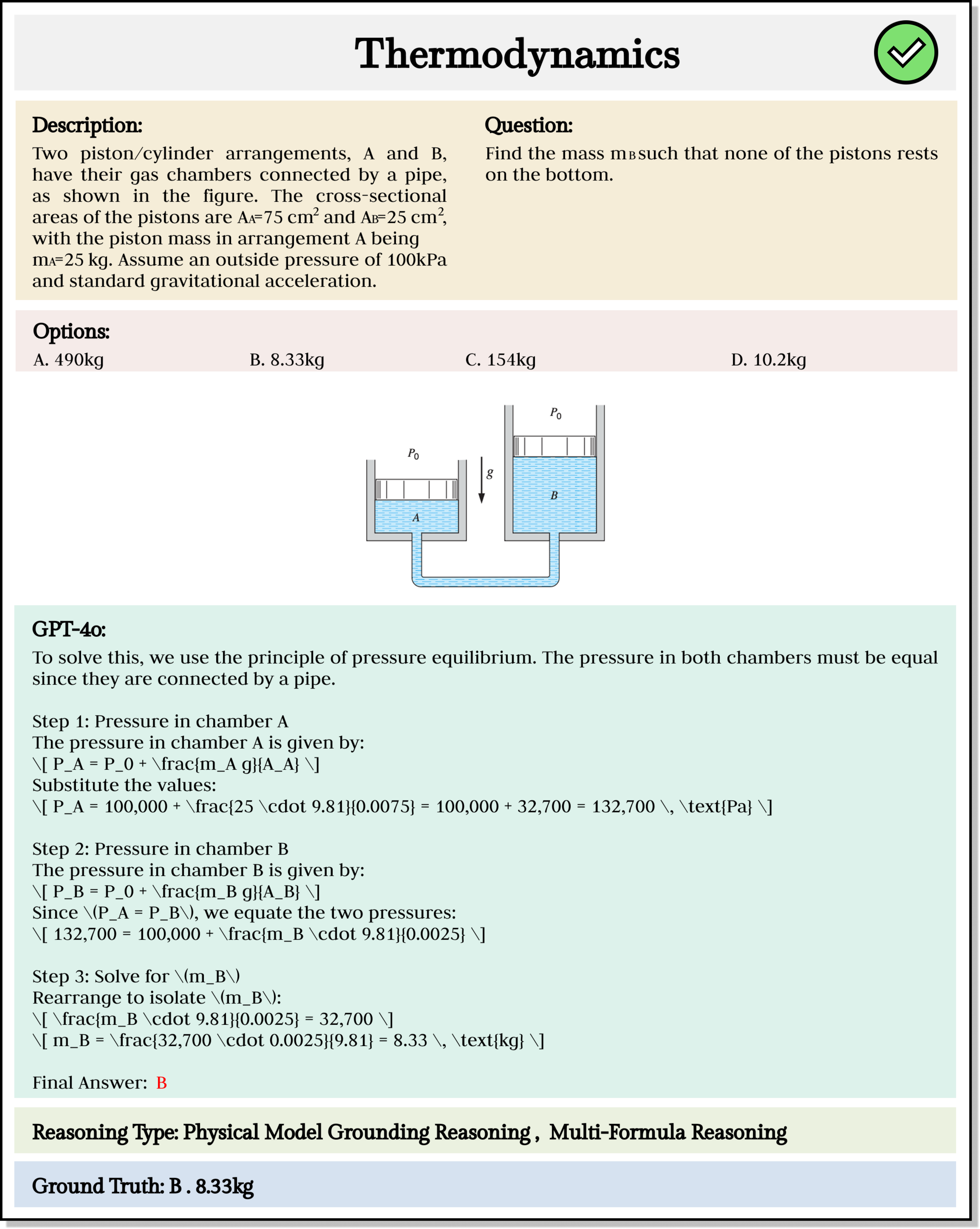}
    \caption{A sample correct case of Thermodynamics.\newline \centering \hyperref[list:list_of_figures]{Back to List of Figures} \textcolor{red}{$|$} \hyperref[tab:list_of_case_study_figures]{Back to Table Index}}
    \addcontentsline{afg}{appfigures}{\protect\numberline{\thefigure}Thermodynamics 2: Correct Case}
\label{fig:thermodynamics_2}
\end{figure*}
\newpage

\begin{figure*}[!htbp]
    \centering
\includegraphics[width=1\linewidth]{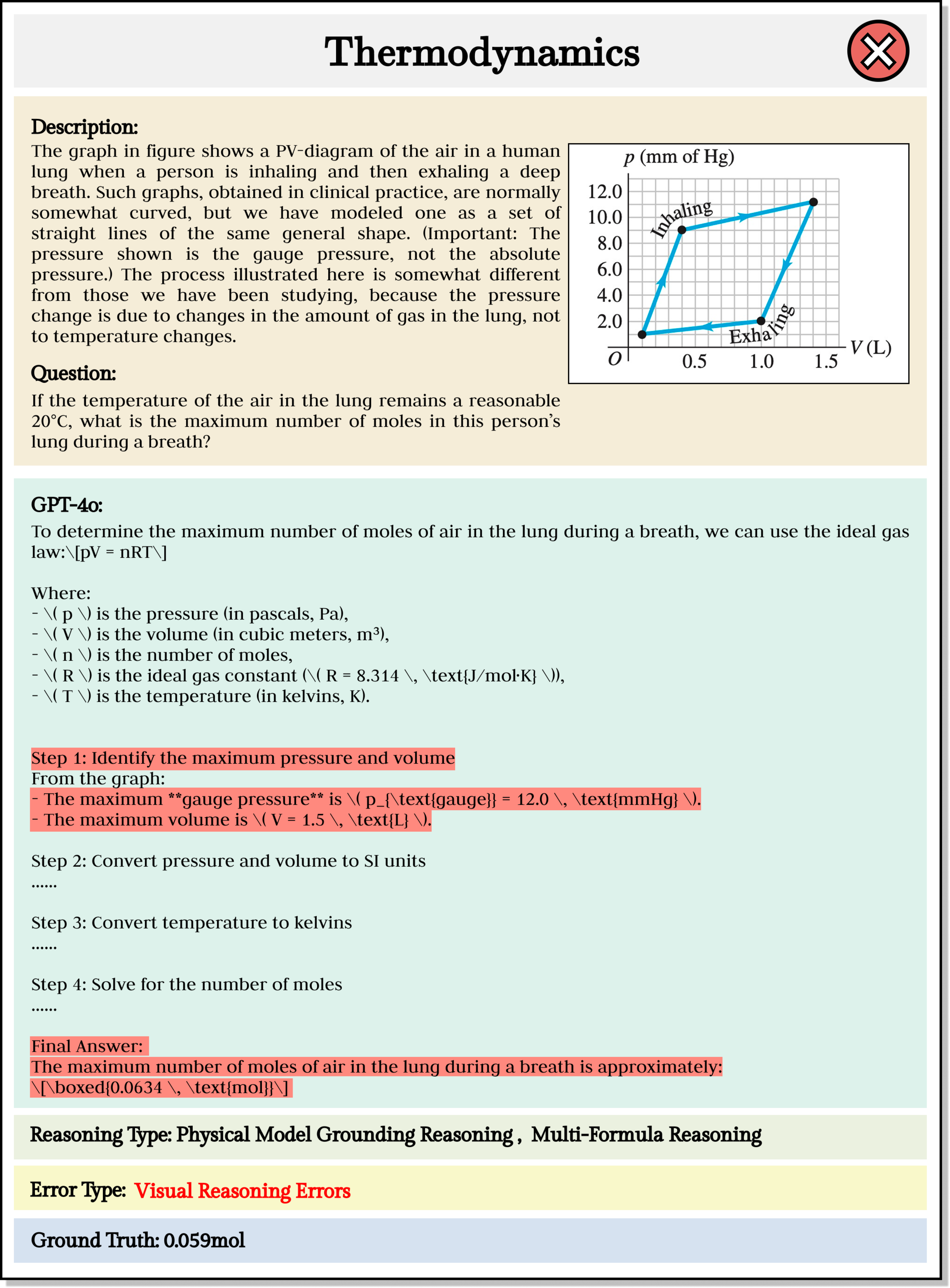}
    \caption{A sample error case of Thermodynamics. Error category: Visual Reasoning Error \newline \centering \hyperref[list:list_of_figures]{Back to List of Figures} \textcolor{red}{$|$} \hyperref[tab:list_of_case_study_figures]{Back to Table Index}}
    \addcontentsline{afg}{appfigures}{\protect\numberline{\thefigure}Thermodynamics 3: Visual Reasoning Error}
\label{fig:thermodynamics_3}
\end{figure*}
\newpage

\begin{figure*}[!htbp]
    \centering
\includegraphics[width=1\linewidth]{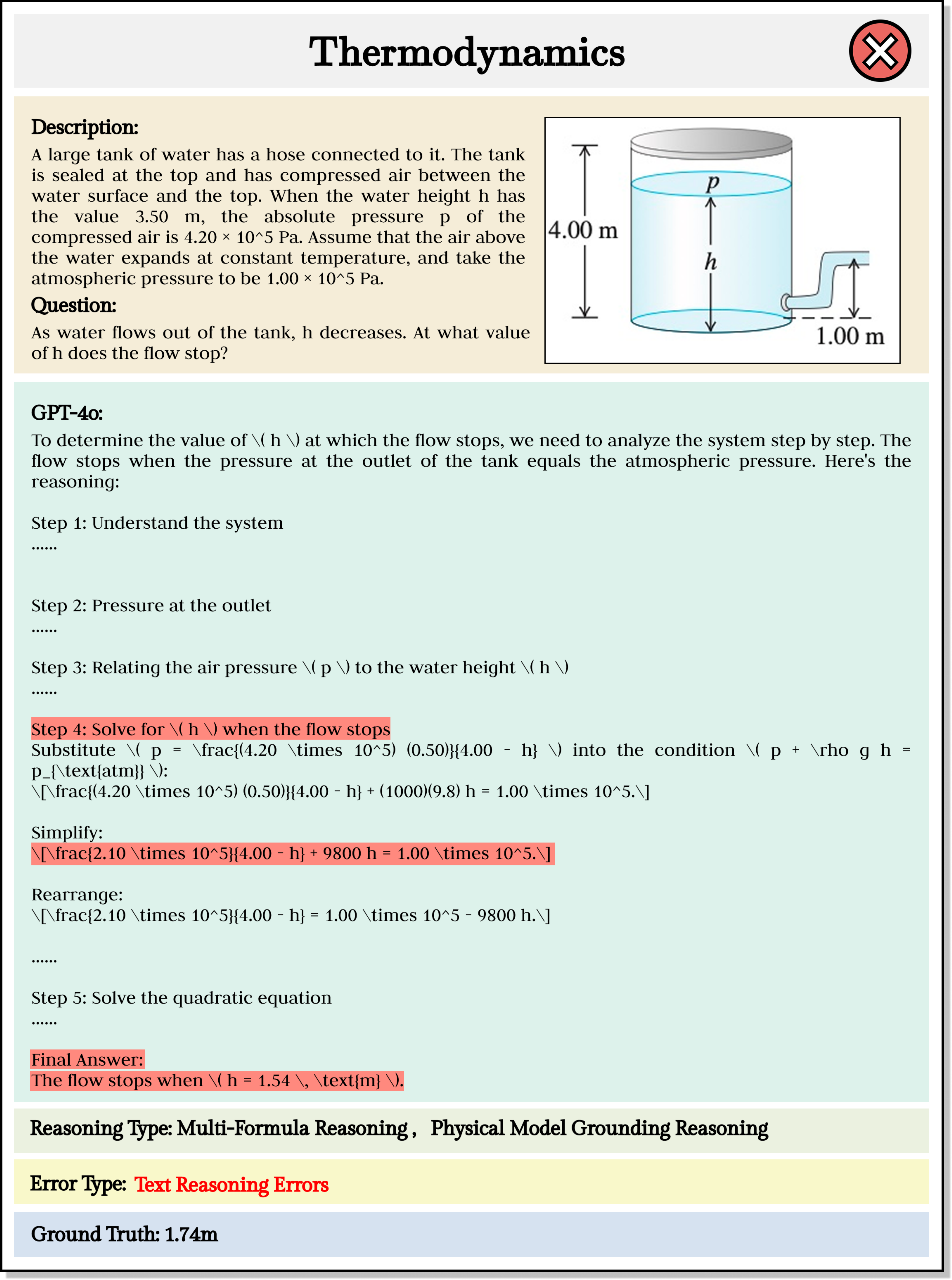}
    \caption{A sample error case of Thermodynamics. Error category: Text Reasoning Error \newline \centering \hyperref[list:list_of_figures]{Back to List of Figures} \textcolor{red}{$|$} \hyperref[tab:list_of_case_study_figures]{Back to Table Index}}
    \addcontentsline{afg}{appfigures}{\protect\numberline{\thefigure}Thermodynamics 4: Text Reasoning Error}
\label{fig:thermodynamics_4}
\end{figure*}
\newpage

\begin{figure*}[!htbp]
    \centering
\includegraphics[width=1\linewidth]{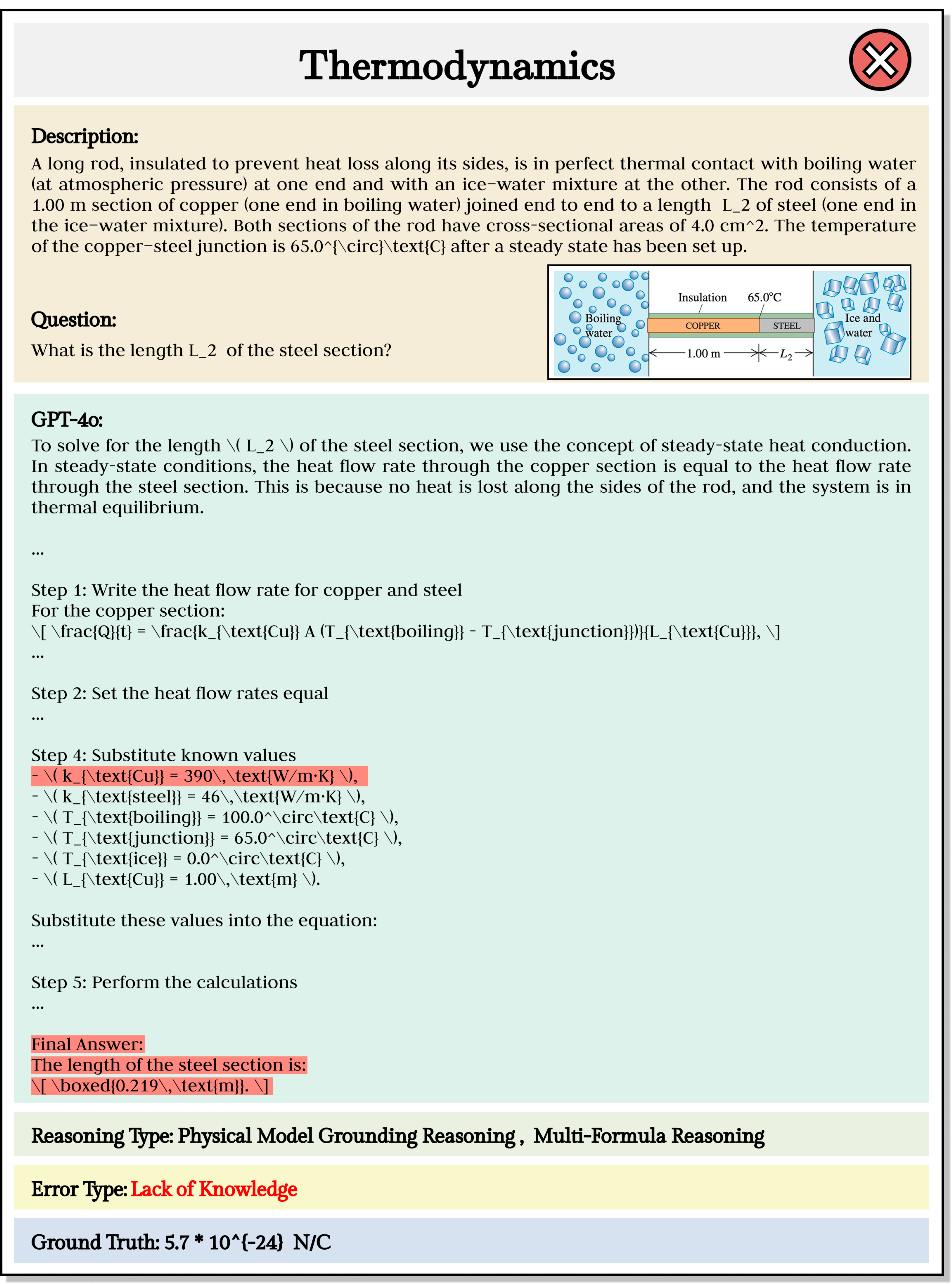}
    \caption{A sample error case of Thermodynamics. Error category: Lack of Knowledge \newline \centering \hyperref[list:list_of_figures]{Back to List of Figures} \textcolor{red}{$|$} \hyperref[tab:list_of_case_study_figures]{Back to Table Index}}
    \addcontentsline{afg}{appfigures}{\protect\numberline{\thefigure}Thermodynamics 5: Lack of Knowledge}
\label{fig:thermodynamics_5}
\end{figure*}
\newpage

\begin{figure*}[!htbp]
    \centering
\includegraphics[width=1\linewidth]{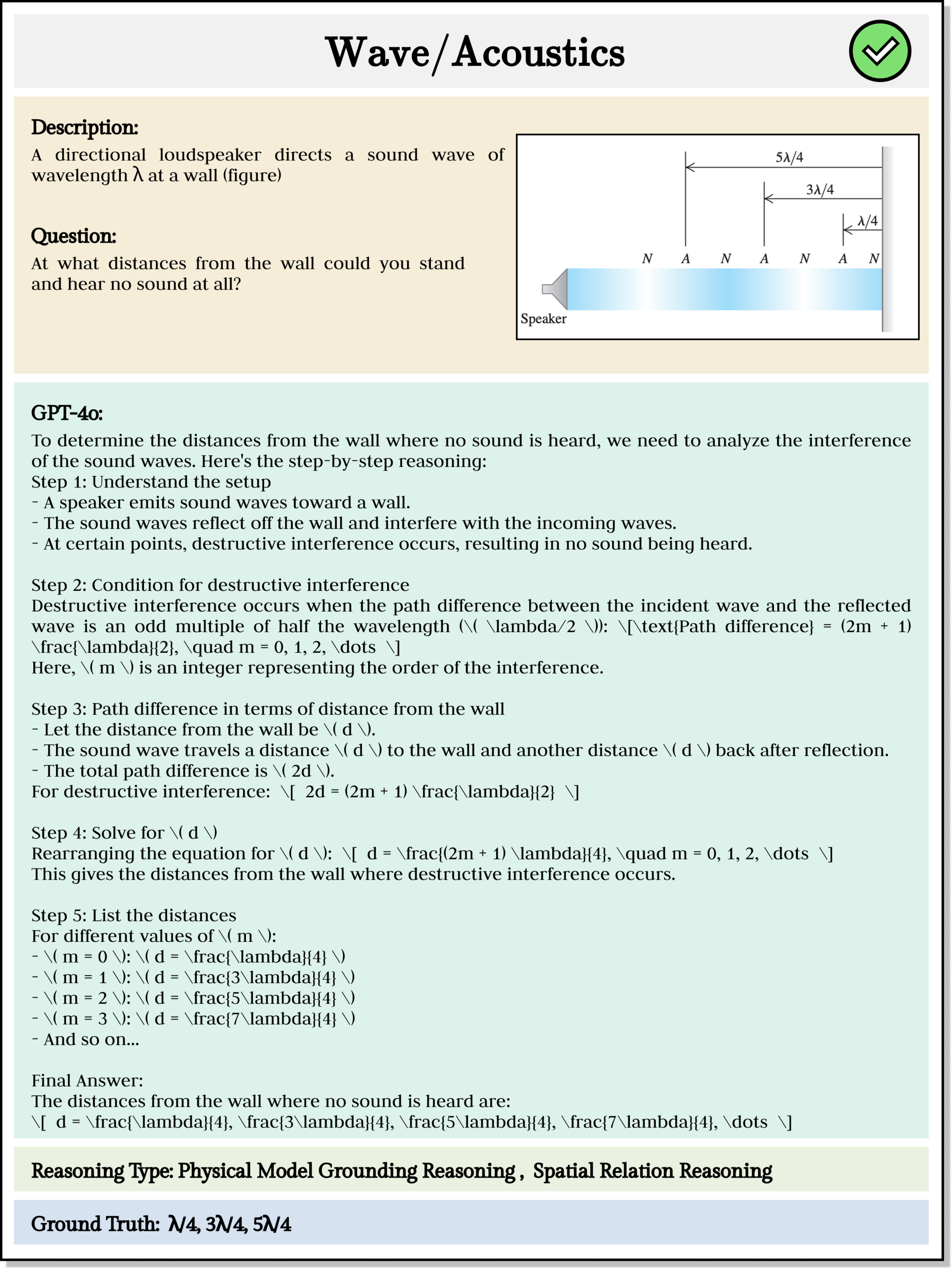}
    \caption{A sample correct case of Wave/Acoustics.\\ \hyperref[list:list_of_figures]{Back to List of Figures} \textcolor{red}{$|$} \hyperref[tab:list_of_case_study_figures]{Back to Table Index}}
    \addcontentsline{afg}{appfigures}{\protect\numberline{\thefigure}Wave/Acoustics 1: Correct Case}
\label{fig:wave_1}
\end{figure*}
\newpage

\begin{figure*}[!htbp]
    \centering
\includegraphics[width=1\linewidth]{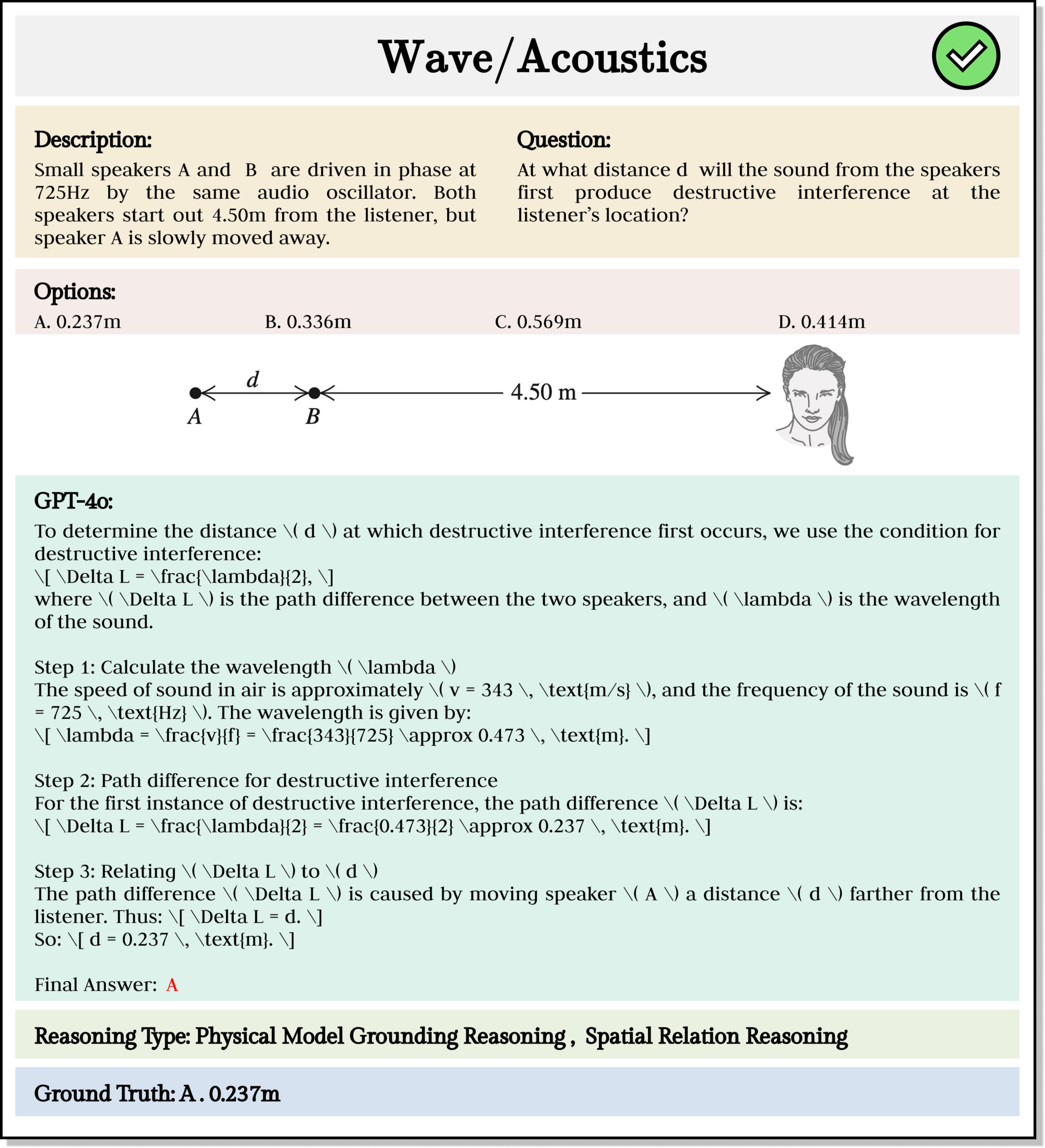}
    \caption{A sample correct case of Wave/Acoustics.\\ \hyperref[list:list_of_figures]{Back to List of Figures} \textcolor{red}{$|$} \hyperref[tab:list_of_case_study_figures]{Back to Table Index}}
    \addcontentsline{afg}{appfigures}{\protect\numberline{\thefigure}Wave/Acoustics 2: Correct Case}
\label{fig:wave_2}
\end{figure*}
\newpage

\begin{figure*}[!htbp]
    \centering
\includegraphics[width=1\linewidth]{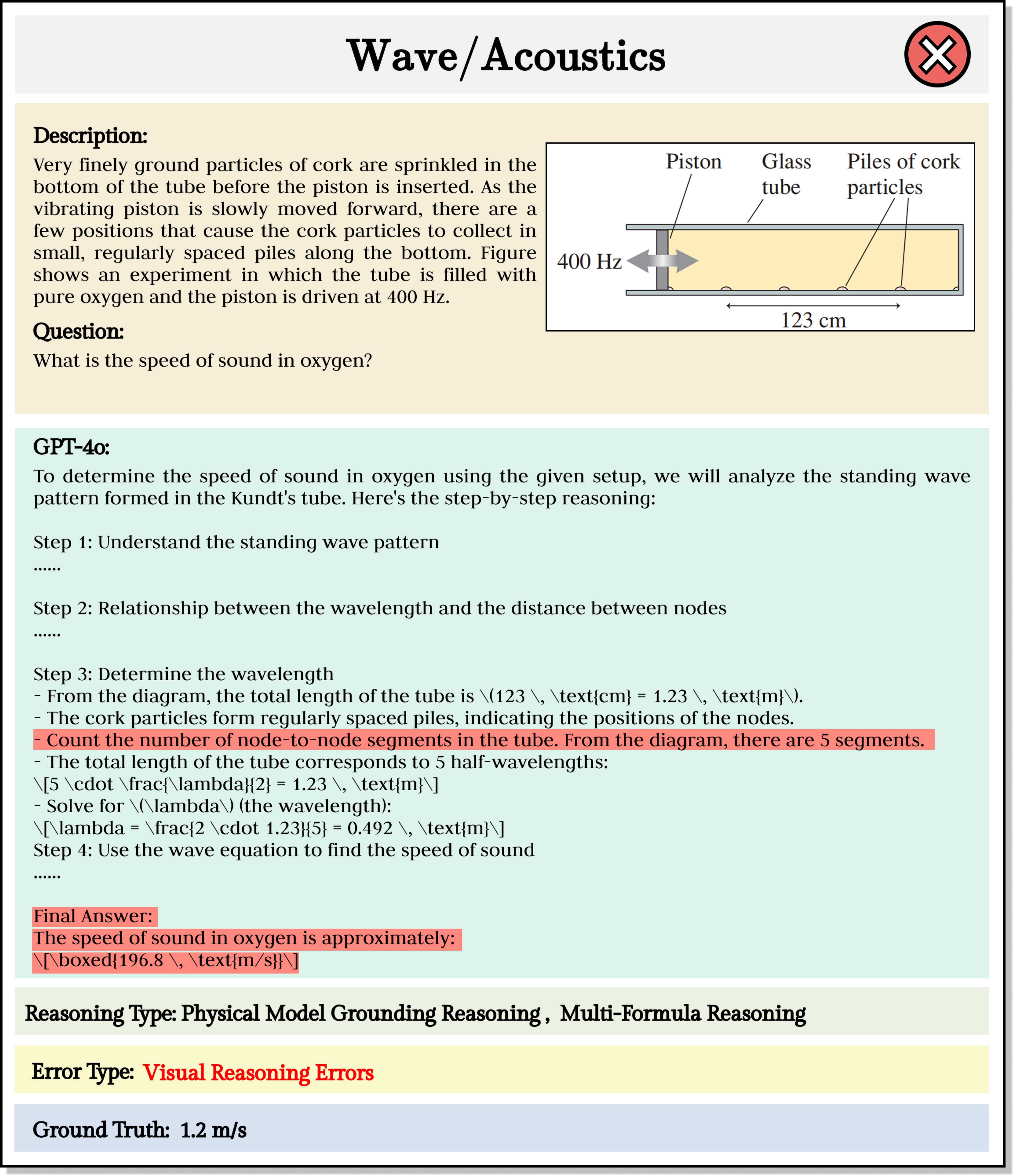}
    \caption{A sample error case of Wave/Acoustics. Error category: Visual Reasoning Error \newline \centering \hyperref[list:list_of_figures]{Back to List of Figures} \textcolor{red}{$|$} \hyperref[tab:list_of_case_study_figures]{Back to Table Index}}
    \addcontentsline{afg}{appfigures}{\protect\numberline{\thefigure}Wave/Acoustics 3: Visual Reasoning Error}
\label{fig:wave_3}
\end{figure*}
\newpage

\begin{figure*}[!htbp]
    \centering
\includegraphics[width=1\linewidth]{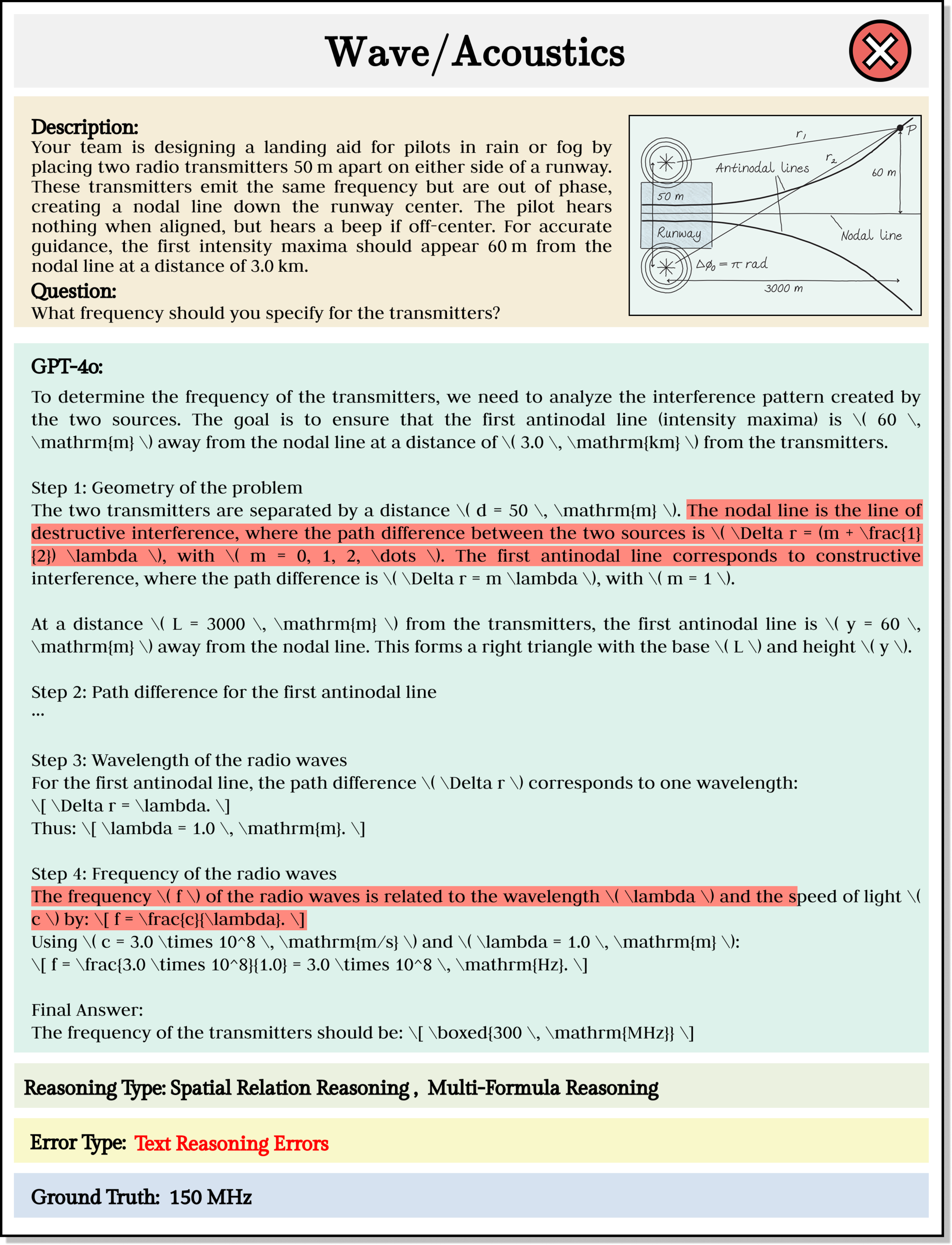}
    \caption{A sample error case of Wave/Acoustics. Error category: Text Reasoning Error \newline \centering \hyperref[list:list_of_figures]{Back to List of Figures} \textcolor{red}{$|$} \hyperref[tab:list_of_case_study_figures]{Back to Table Index}}
    \addcontentsline{afg}{appfigures}{\protect\numberline{\thefigure}Wave/Acoustics 4: Text Reasoning Error}
\label{fig:wave_4}
\end{figure*}
\newpage

\begin{figure*}[!htbp]
    \centering
\includegraphics[width=1\linewidth]{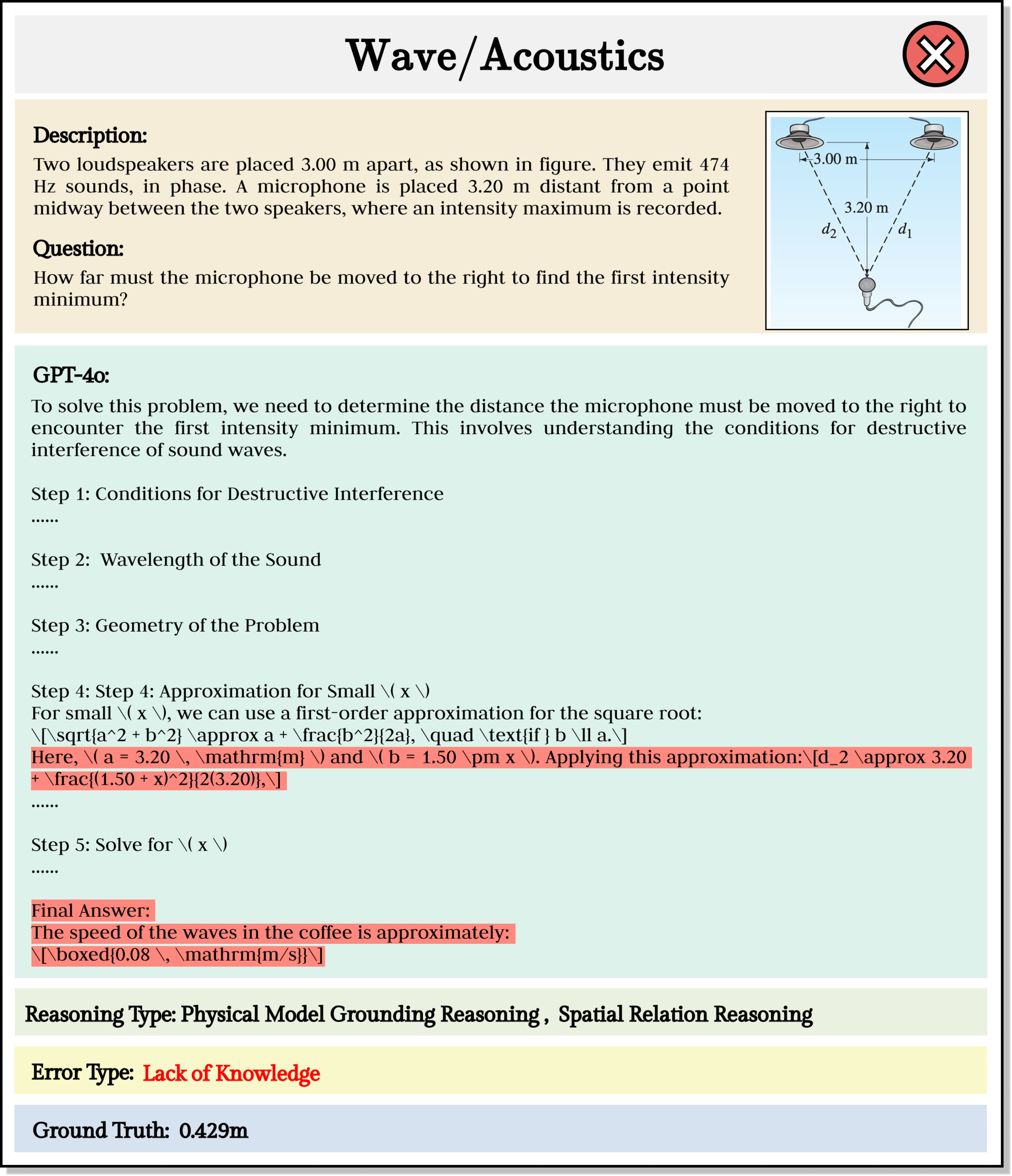}
    \caption{A sample error case of Wave/Acoustics. Error category: Lack of Knowledge \newline \centering \hyperref[list:list_of_figures]{Back to List of Figures} \textcolor{red}{$|$} \hyperref[tab:list_of_case_study_figures]{Back to Table Index}}
    \addcontentsline{afg}{appfigures}{\protect\numberline{\thefigure}Wave/Acoustics 5: Lack of Knowledge}
\label{fig:wave_5}
\end{figure*}
\newpage

\begin{figure*}[!htbp]
    \centering
\includegraphics[width=1\linewidth]{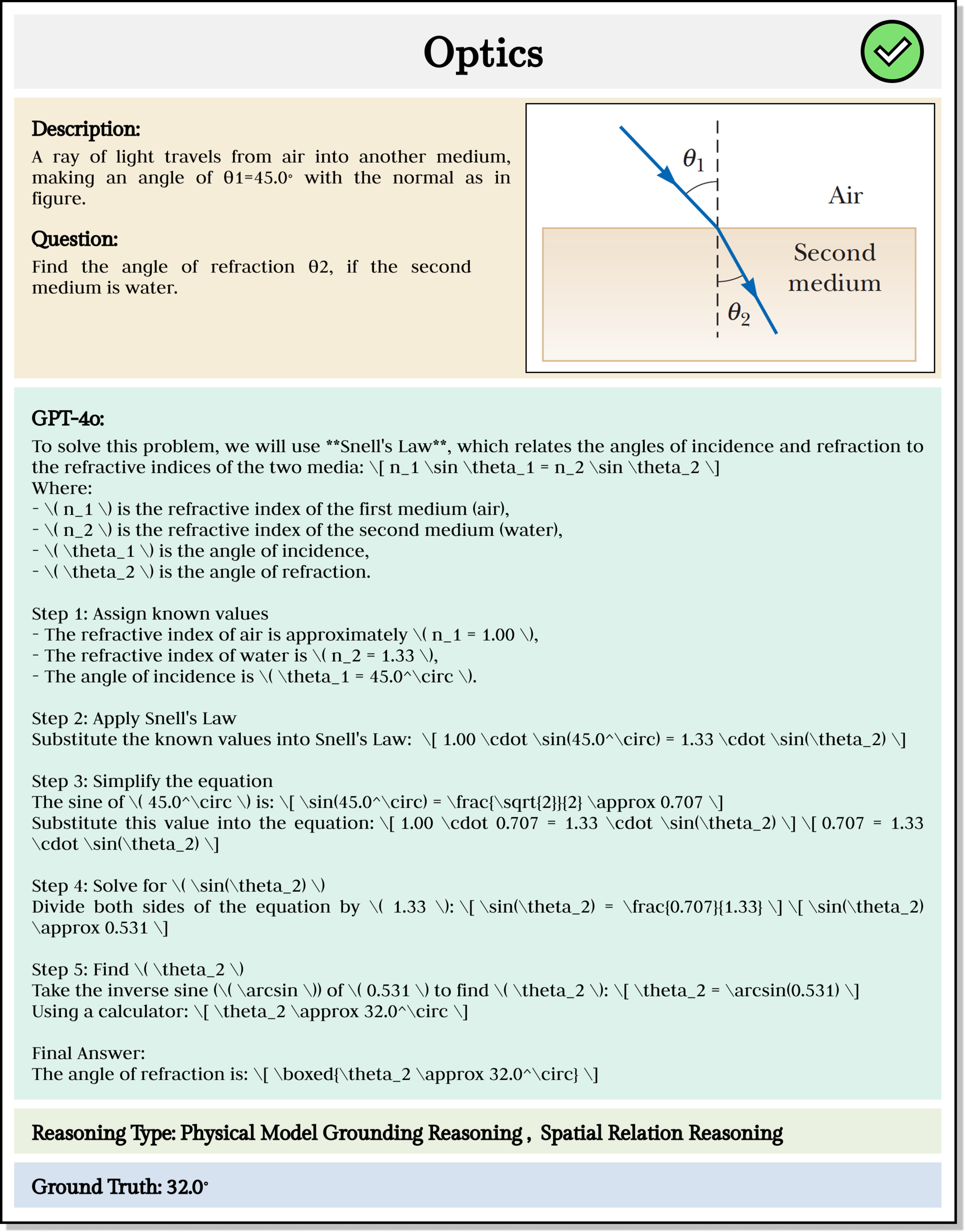}
    \caption{A sample correct case of Optics.\\ \hyperref[list:list_of_figures]{Back to List of Figures} \textcolor{red}{$|$} \hyperref[tab:list_of_case_study_figures]{Back to Table Index}}
    \addcontentsline{afg}{appfigures}{\protect\numberline{\thefigure}Optics 1: Correct Case}
\label{fig:optics_1}
\end{figure*}
\newpage

\begin{figure*}[!htbp]
    \centering
\includegraphics[width=1\linewidth]{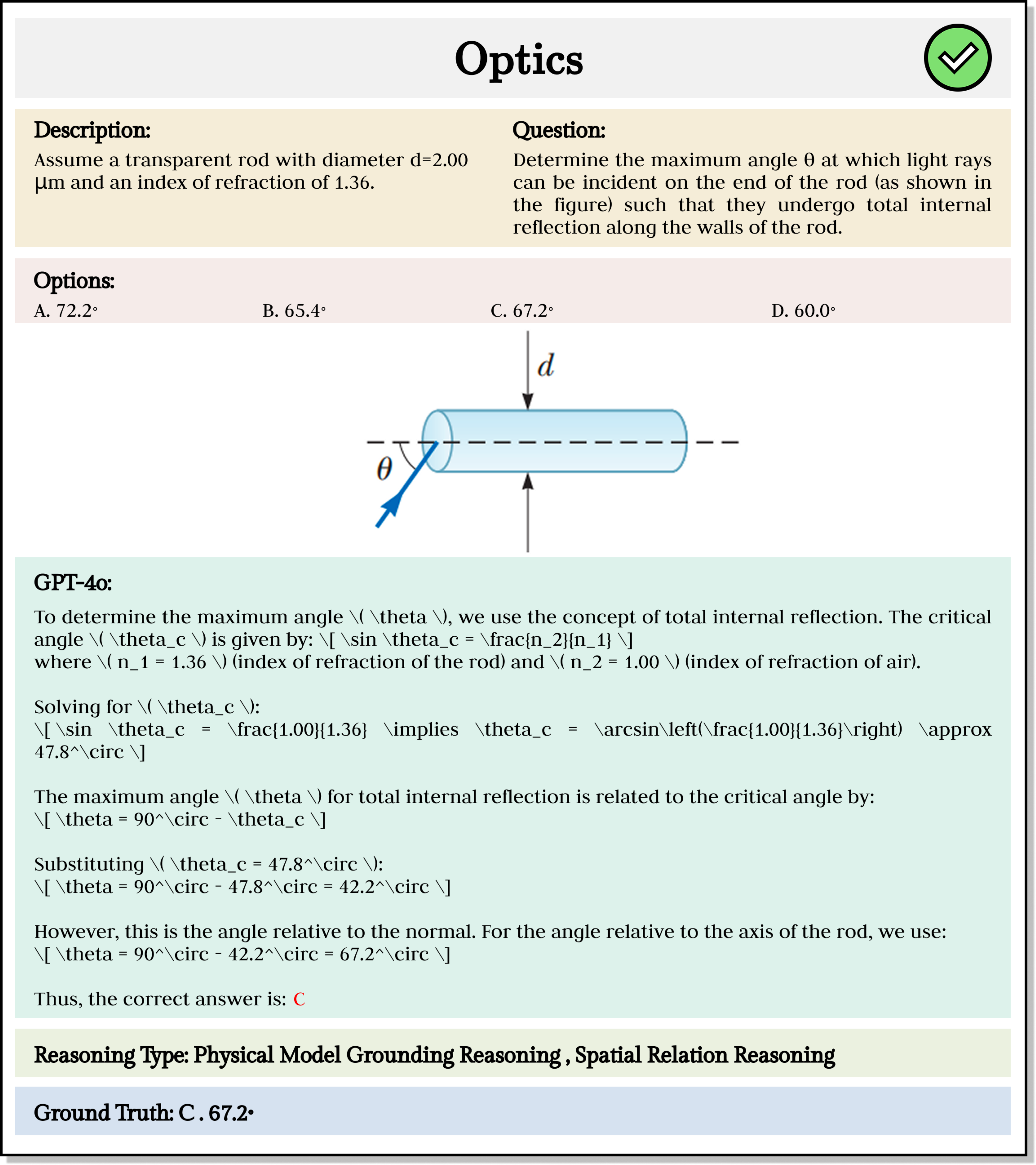}
    \caption{A sample correct case of Optics.\\ \hyperref[list:list_of_figures]{Back to List of Figures} \textcolor{red}{$|$} \hyperref[tab:list_of_case_study_figures]{Back to Table Index}}
    \addcontentsline{afg}{appfigures}{\protect\numberline{\thefigure}Optics 2: Correct Case}
\label{fig:optics_2}
\end{figure*}
\newpage

\begin{figure*}[!htbp]
    \centering
\includegraphics[width=1\linewidth]{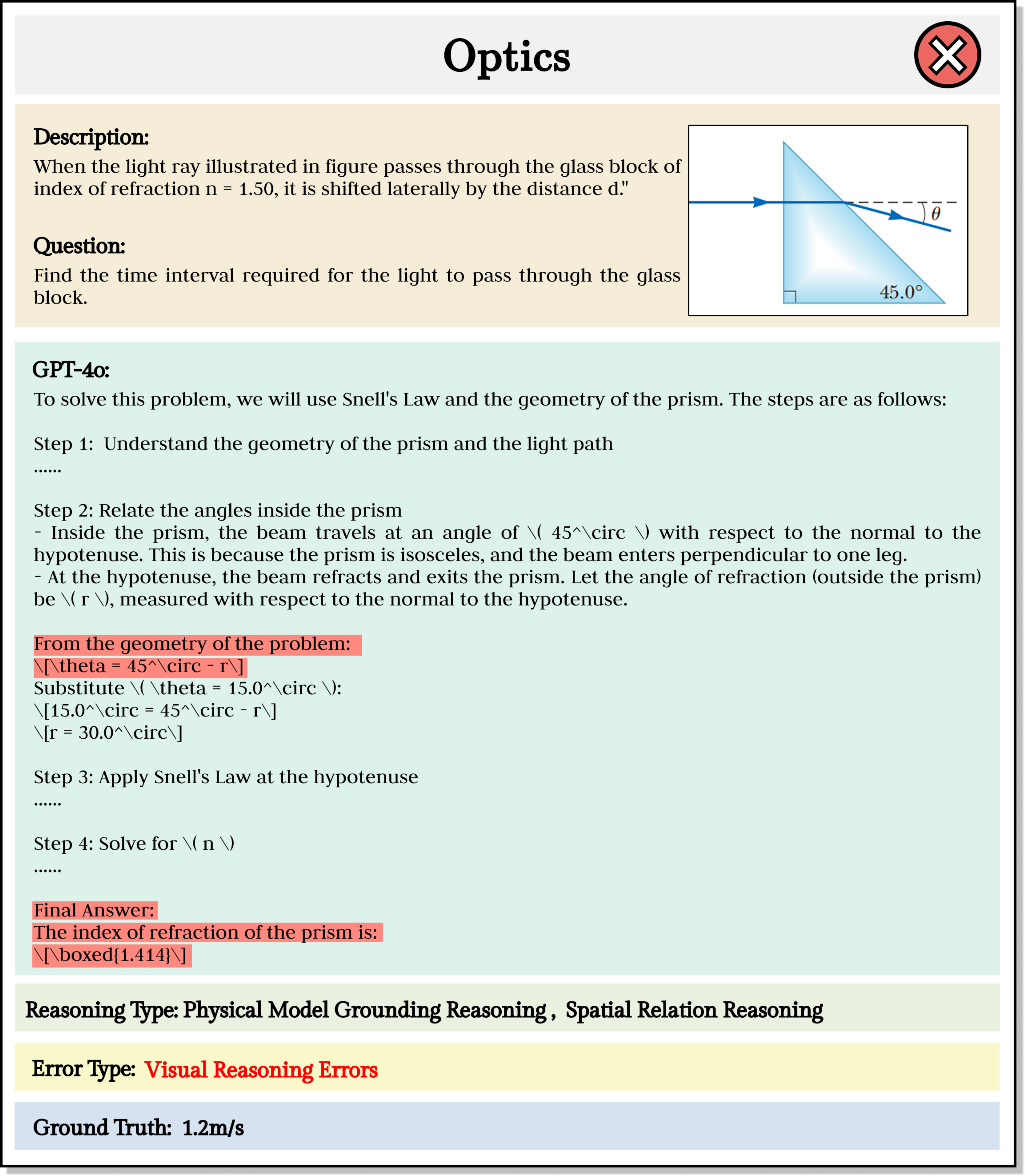}
    \caption{A sample error case of Optics. Error category: Visual Reasoning Error \newline \centering \hyperref[list:list_of_figures]{Back to List of Figures} \textcolor{red}{$|$} \hyperref[tab:list_of_case_study_figures]{Back to Table Index}}
    \addcontentsline{afg}{appfigures}{\protect\numberline{\thefigure}Optics 3: Visual Reasoning Error}
\label{fig:optics_3}
\end{figure*}
\newpage

\begin{figure*}[!htbp]
    \centering
\includegraphics[width=1\linewidth]{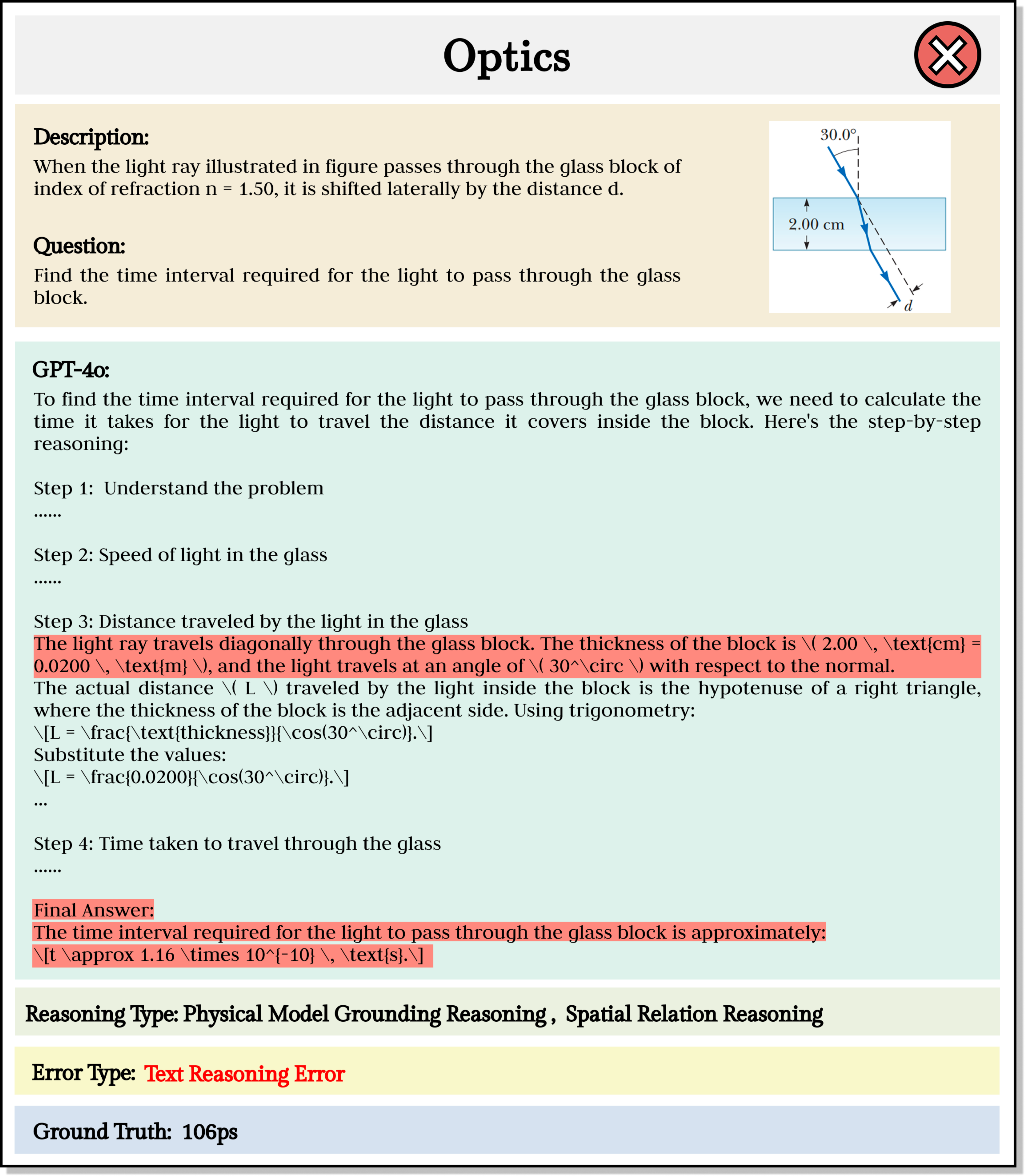}
    \caption{A sample error case of Optics. Error category: Text Reasoning Error \newline \centering \hyperref[list:list_of_figures]{Back to List of Figures} \textcolor{red}{$|$} \hyperref[tab:list_of_case_study_figures]{Back to Table Index}}
    \addcontentsline{afg}{appfigures}{\protect\numberline{\thefigure}Optics 4: Text Reasoning Error}
\label{fig:optics_4}
\end{figure*}
\newpage

\begin{figure*}[!htbp]
    \centering
\includegraphics[width=1\linewidth]{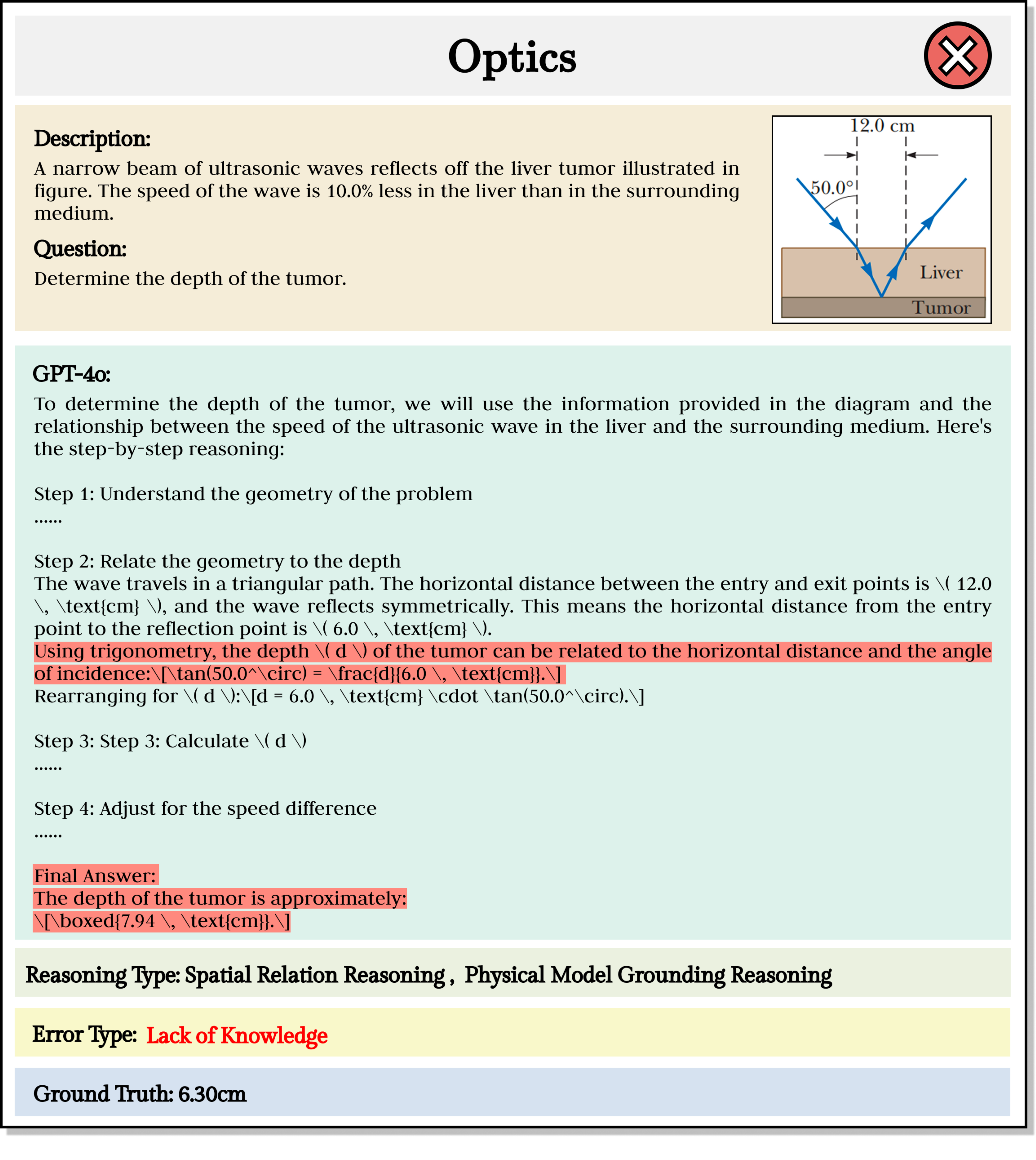}
    \caption{A sample error case of Optics. Error category: Lack of Knowledge \newline \centering \hyperref[list:list_of_figures]{Back to List of Figures} \textcolor{red}{$|$} \hyperref[tab:list_of_case_study_figures]{Back to Table Index}}
    \addcontentsline{afg}{appfigures}{\protect\numberline{\thefigure}Optics 5: Lack of Knowledge}
\label{fig:optics_5}
\end{figure*}
\newpage

\begin{figure*}[!htbp]
    \centering
\includegraphics[width=1\linewidth]{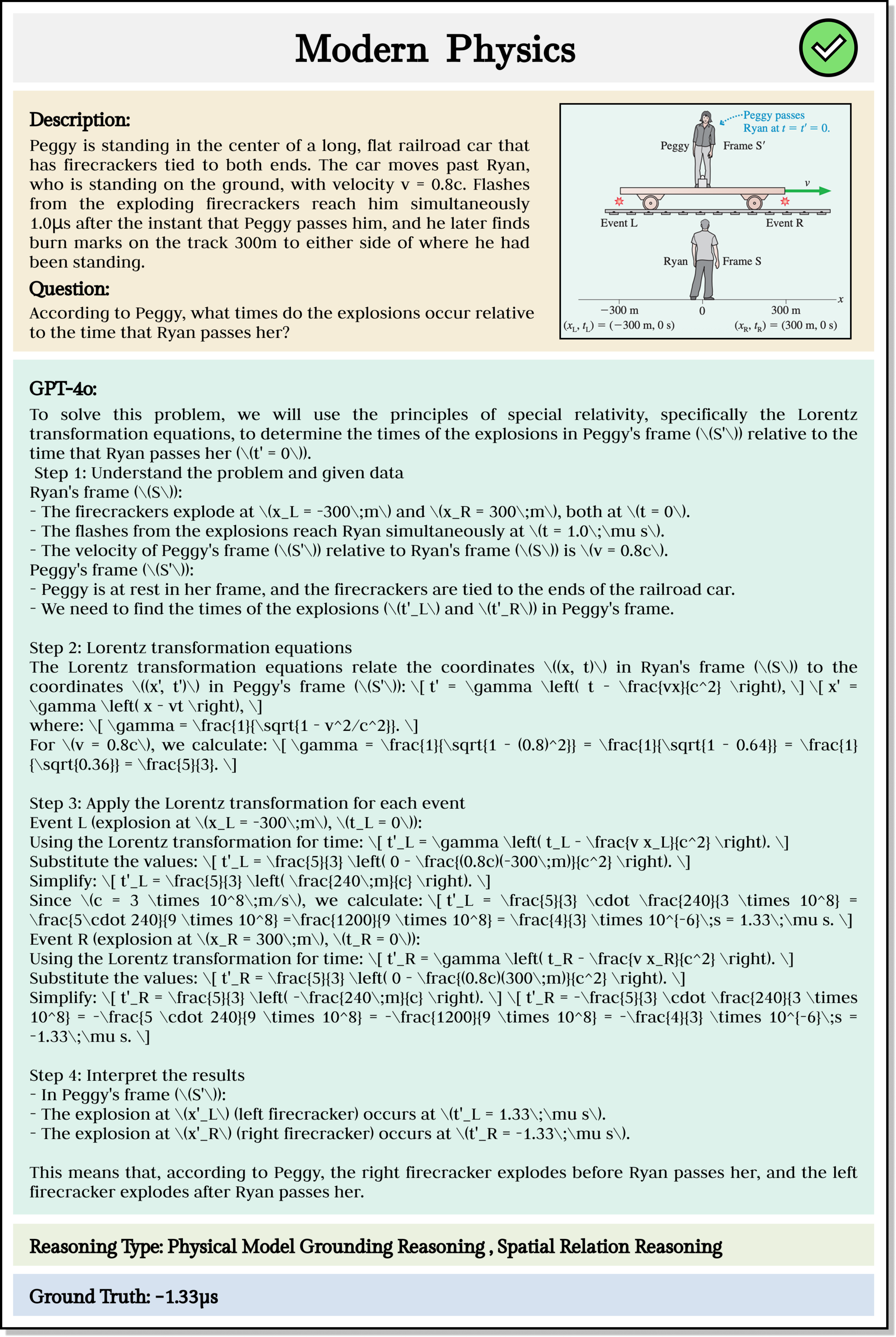}
    \caption{A sample correct case of Modern Physics.\\ \hyperref[list:list_of_figures]{Back to List of Figures} \textcolor{red}{$|$} \hyperref[tab:list_of_case_study_figures]{Back to Table Index}}
    \addcontentsline{afg}{appfigures}{\protect\numberline{\thefigure}Modern Physics 1: Correct Case}
\label{fig:modern_1}
\end{figure*}
\newpage

\begin{figure*}[!htbp]
    \centering
\includegraphics[width=1\linewidth]{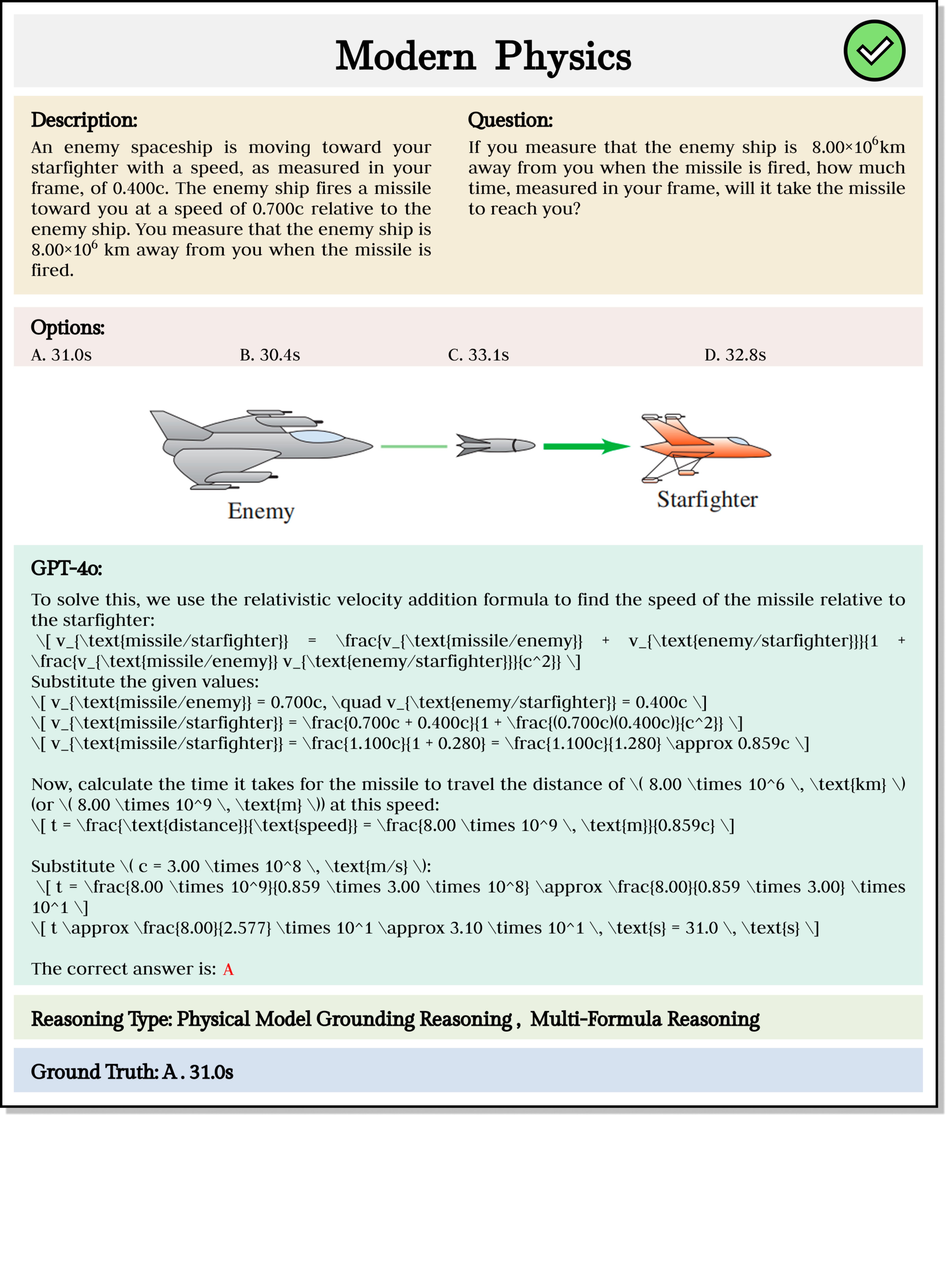}
    \caption{A sample correct case of Modern Physics.\\ \hyperref[list:list_of_figures]{Back to List of Figures} \textcolor{red}{$|$} \hyperref[tab:list_of_case_study_figures]{Back to Table Index}}
    \addcontentsline{afg}{appfigures}{\protect\numberline{\thefigure}Modern Physics 2: Correct Case}
\label{fig:modern_2}
\end{figure*}
\newpage

\begin{figure*}[!htbp]
    \centering
\includegraphics[width=1\linewidth]{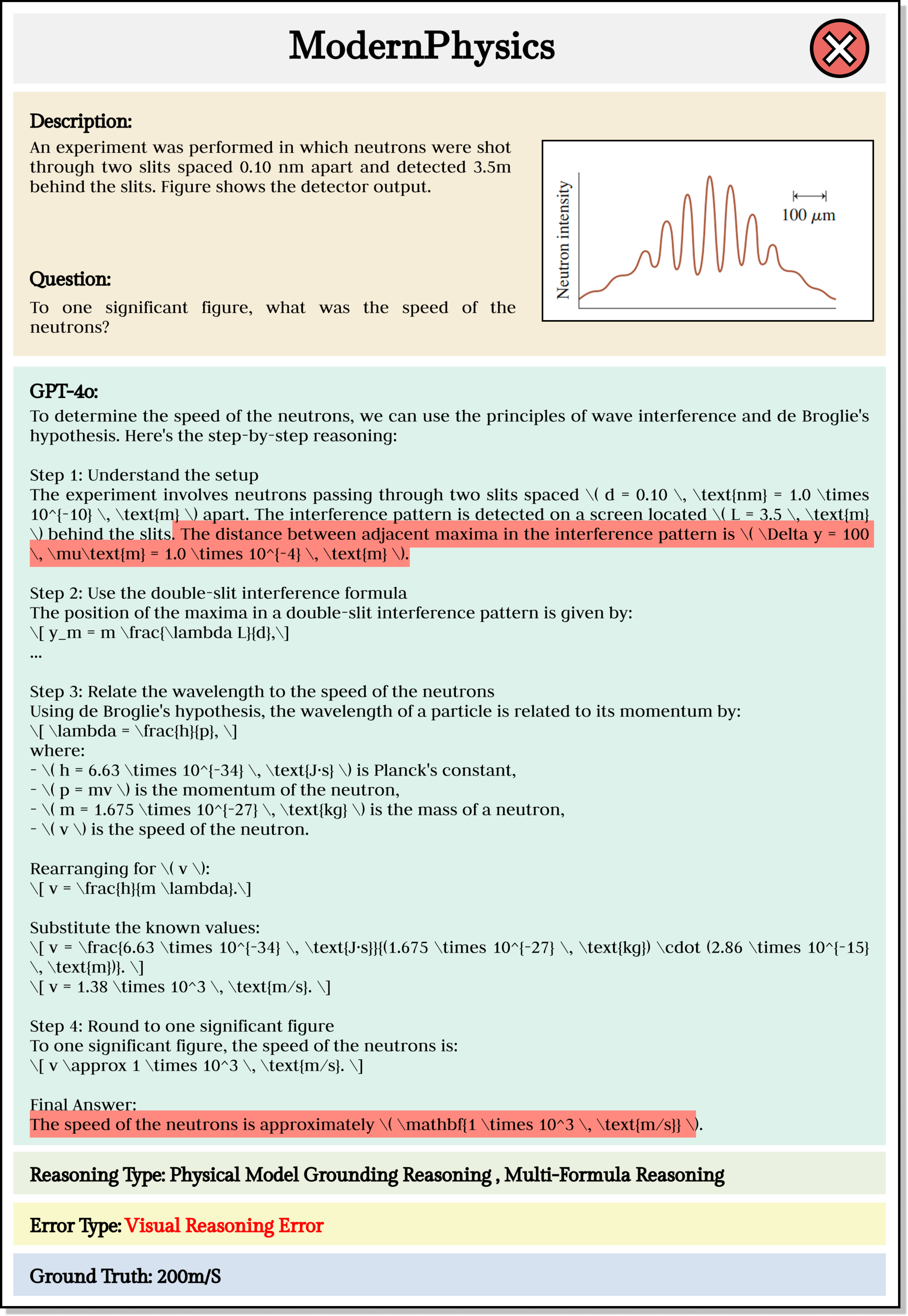}
    \caption{A sample error case of Modern Physics. Error category: Visual Reasoning Error \newline \centering \hyperref[list:list_of_figures]{Back to List of Figures} \textcolor{red}{$|$} \hyperref[tab:list_of_case_study_figures]{Back to Table Index}}
    \addcontentsline{afg}{appfigures}{\protect\numberline{\thefigure}Modern Physics 3: Visual Reasoning Error}
\label{fig:modern_3}
\end{figure*}
\newpage

\begin{figure*}[!htbp]
    \centering
\includegraphics[width=1\linewidth]{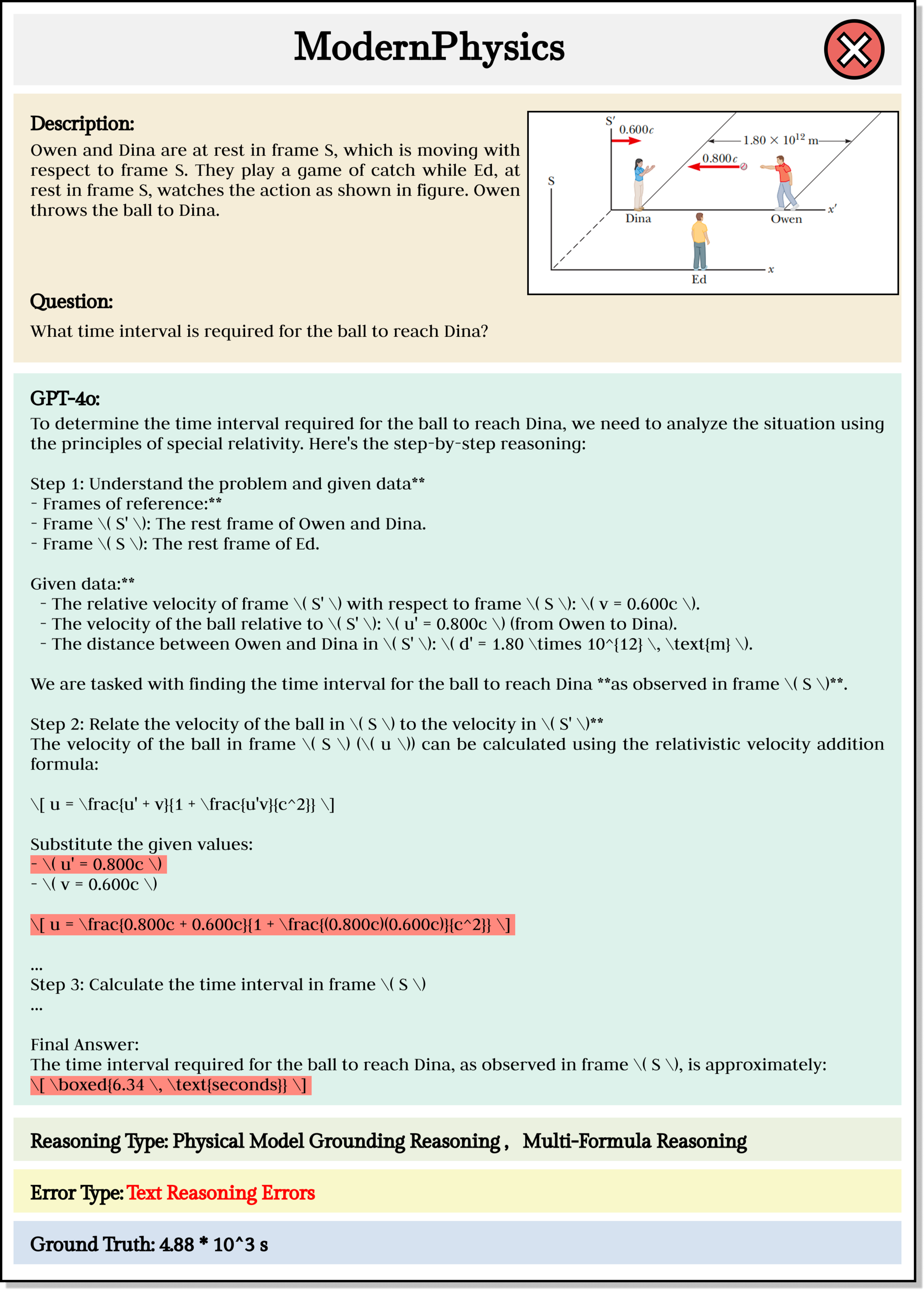}
    \caption{A sample error case of Modern Physics. Error category: Text Reasoning Error \newline \centering \hyperref[list:list_of_figures]{Back to List of Figures} \textcolor{red}{$|$} \hyperref[tab:list_of_case_study_figures]{Back to Table Index}}
    \addcontentsline{afg}{appfigures}{\protect\numberline{\thefigure}Modern Physics 4: Text Reasoning Error}
\label{fig:modern_4}
\end{figure*}
\newpage

\begin{figure*}[!htbp]
    \centering
\includegraphics[width=1\linewidth]{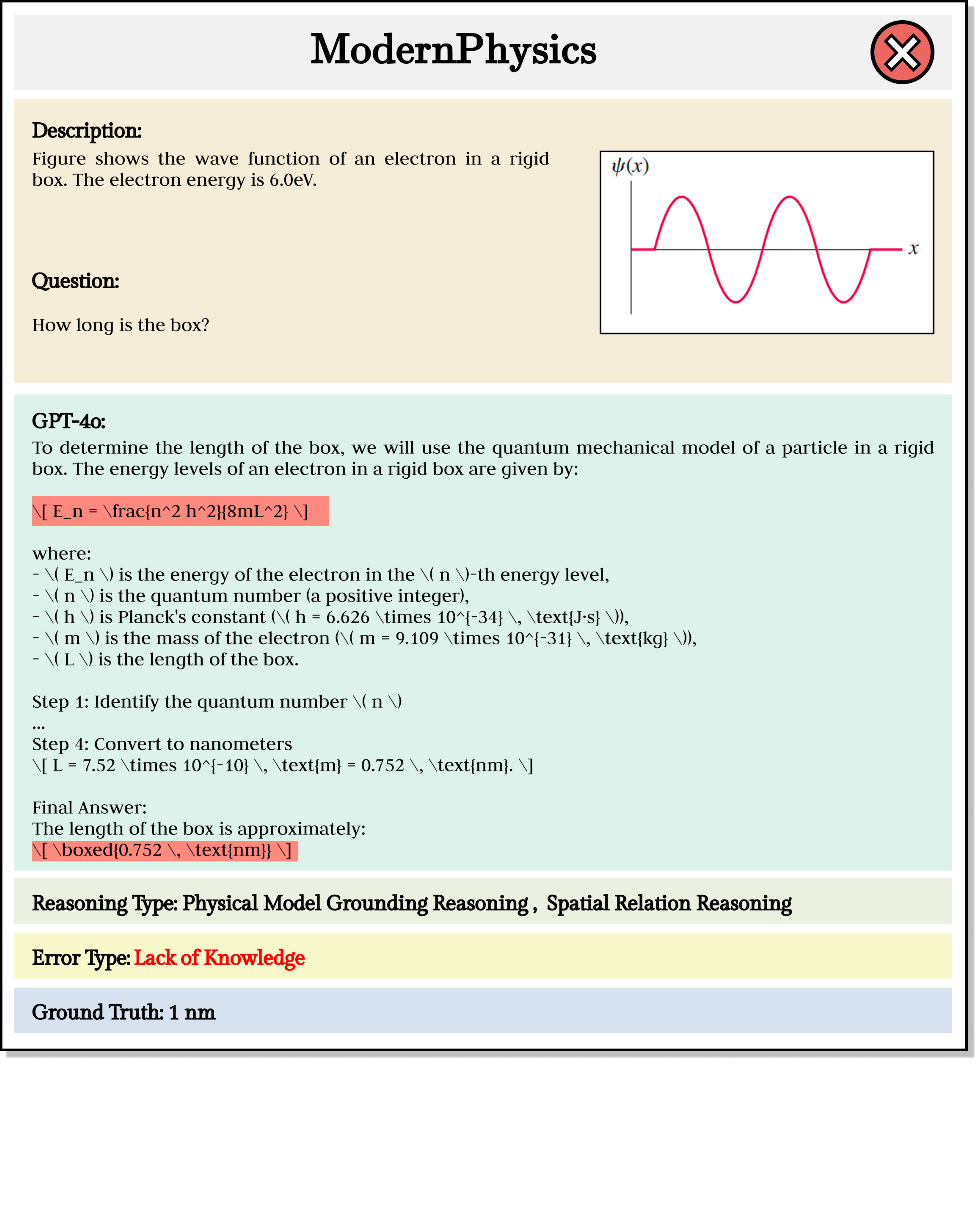}
    \caption{A sample error case of Modern Physics. Error category: Lack of Knowledge \newline \centering \hyperref[list:list_of_figures]{Back to List of Figures} \textcolor{red}{$|$} \hyperref[tab:list_of_case_study_figures]{Back to Table Index}}
    \addcontentsline{afg}{appfigures}{\protect\numberline{\thefigure}Modern Physics 5: Lack of Knowledge}
\label{fig:modern_5}
\end{figure*}
\newpage

\clearpage
\section{Data Annotation Protocol}
\label{appendix:data_annotation}
This document outlines a detailed procedure for annotating a dataset of physics questions that include visual context.

\subsection{Data Collection}

\paragraph{Sources of Data.} Data is collected from freely accessible online resources, textbooks, and other materials. Annotators are instructed to use a wide range of sources rather than relying on just one.

\vspace{3pt}
\noindent\textbf{Types of Questions:}
\vspace{1pt}

\begin{itemize}
    \item \textbf{Multiple-Choice Questions:} These consist of a question accompanied by four answer options, with only one being correct. For each multiple-choice question, annotators are also required to create a corresponding open-ended version of the same problem.
    \item \textbf{Open-Ended Questions:} These include formats such as short-answer and calculation-based problems. Questions with excessively lengthy answers should be avoided. For each open-ended question, a corresponding multiple-choice version should also be constructed.
\end{itemize}
 
\noindent\textbf{Image Types.} The annotators should find images with realistic physical senarios.

\subsection{General Guidelines}

\begin{itemize}
    \item \textbf{General Principles:} Annotations should be accurate, uniform, and maintain a high level of academic quality.
    
    \item \textbf{Specific Instructions:}
    \vspace{1pt}
        \begin{itemize}
            \item All questions should be written in English.
            \item All questions must contain one physical images.
            \item All images in question should be realistic, in specific physical scenarios.
            \item The question should not be ambiguous and can be answered with one of the given options or a short answer. 
            \item Annotate all data fields, including the description, simplified description, question, answer options, the correct answer, image, and domain.
        \end{itemize}
\end{itemize}

\subsection{Data Format and Structure}

\begin{itemize}
    \item \textbf{JSON File Format:} The structured JSON format will include fields for index number, description, simplified description, question, answer options, correct answer, and domain.
    
    \item \textbf{Naming Conventions:}
    \vspace{1pt}
        \begin{itemize}
            \item Each collected sample will be stored in a line into a JSONL file.
            \item Image files following a standard naming rule: \textbf{\{QuesNum\}}.png
        \end{itemize}

    \item \textbf{Interleaving Question with Images:} The images should be inserted as a file path in the question. 
\end{itemize}

\subsection{Quality Control and Validation}

\begin{itemize}
    \item Annotators will cross-check each other’s work to ensure accuracy and compliance with the annotation guidelines.
    \item Periodic reviews of randomly selected samples from the dataset will be carried out to maintain consistent quality over time.
\end{itemize}

\subsection{Handling Ambiguities}
Any ambiguous or unclear data entries should be marked for thorough review. Such questions will be collectively discussed during team meetings to develop a consistent and standardized annotation strategy.

\subsection{Ethical Considerations}

\begin{itemize}
    \item \textbf{Copyright and Licensing:} Annotators must strictly follow all applicable copyright and licensing rules. Content from sources that restrict reproduction or redistribution will be excluded without exception.
    \item \textbf{Data Privacy:} Upholding data privacy and ethical standards is essential. Annotators should refrain from including any questions that involve personal or sensitive information.
\end{itemize}

\subsection{Data Contamination Considerations}
When developing benchmarks for evaluating foundation models, it is crucial to account for the potential risk of data contamination. To mitigate this, annotators should deliberately avoid simple questions with widely available answers. Instead, they should prioritize selecting problems whose solutions are embedded in less conspicuous places—such as in supplementary materials or at the end of lengthy textbooks. This strategy helps ensure that the benchmark effectively challenges models to demonstrate genuine comprehension and reasoning across complex and less accessible content.

\end{document}